\documentclass[final,onefignum,onetabnum]{siamart190516}

\usepackage{braket,amsfonts}

\usepackage{array}

\usepackage[caption=false]{subfig}
\usepackage{pgfplots}

\newsiamthm{claim}{Claim}
\newsiamremark{remark}{Remark}
\newsiamremark{hypothesis}{Hypothesis}
\crefname{hypothesis}{Hypothesis}{Hypotheses}

\usepackage{algorithmic}

\usepackage{graphicx,epstopdf}

\Crefname{ALC@unique}{Line}{Lines}

\usepackage{amsopn}

\usepackage{xspace}
\usepackage{bold-extra}
\usepackage[most]{tcolorbox}

\colorlet{texcscolor}{blue!50!black}
\colorlet{texemcolor}{red!70!black}
\colorlet{texpreamble}{red!70!black}
\colorlet{codebackground}{black!25!white!25}

\usepackage{tikz}
\definecolor{sky}{RGB}{18,102,153}

\newsiamthm{expl}{Example}

\lstdefinestyle{siamlatex}{%
  style=tcblatex,
  texcsstyle=*\color{texcscolor},
  texcsstyle=[2]\color{texemcolor},
  keywordstyle=[2]\color{texemcolor},
  moretexcs={cref,Cref,maketitle,mathcal,text,headers,email,url},
}

\tcbset{%
  colframe=black!75!white!75,
  coltitle=white,
  colback=codebackground, % bottom/left side
  colbacklower=white, % top/right side
  fonttitle=\bfseries,
  arc=0pt,outer arc=0pt,
  top=1pt,bottom=1pt,left=1mm,right=1mm,middle=1mm,boxsep=1mm,
  leftrule=0.3mm,rightrule=0.3mm,toprule=0.3mm,bottomrule=0.3mm,
  listing options={style=siamlatex}
}

\newtcblisting[use counter=example]{example}[2][]{%
  title={Example~\thetcbcounter: #2},#1}

\newtcbinputlisting[use counter=example]{\examplefile}[3][]{%
  title={Example~\thetcbcounter: #2},listing file={#3},#1}

\DeclareTotalTCBox{\code}{ v O{} }
{ %fontupper=\ttfamily\color{texemcolor},
  fontupper=\ttfamily\color{black},
  nobeforeafter,
  tcbox raise base,
  colback=codebackground,colframe=white,
  top=0pt,bottom=0pt,left=0mm,right=0mm,
  leftrule=0pt,rightrule=0pt,toprule=0mm,bottomrule=0mm,
  boxsep=0.5mm,
  #2}{#1}

\patchcmd\newpage{\vfil}{}{}{}
\begin{tcbverbatimwrite}{tmp_\jobname_header.tex}
\title{Uncertainty Quantification for Markov Random Fields}

\author{Panagiota Birmpa\thanks{Department of Mathematics and Statistics, University of Massachusetts, Amherst, U.S.A (\email{birmpa@math.umass.edu}).}
\and Markos A. Katsoulakis\thanks{Department of Mathematics and Statistics, University of Massachusetts, Amherst, U.S.A (\email{markos@math.umass.edu}.}}

\headers{Uncertainty Quantification for Markov Random Fields}{Panagiota Birmpa, Markos A. Katsoulakis}
\end{tcbverbatimwrite}
\input{tmp_\jobname_header.tex}

\ifpdf
\hypersetup{ pdftitle={Uncertainty Quantification for Markov Random Fields} }
\fi

\begin{document}
\maketitle

%% ------------------------------------------------------------------
%% ABSTRACT
%% ------------------------------------------------------------------
\begin{tcbverbatimwrite}{tmp_\jobname_abstract.tex}
\begin{abstract}
We present an information-based uncertainty quantification method  for general  Markov Random Fields, also known as Markov Networks.
Markov Random Fields  (MRFs) are structured, probabilistic graphical models over undirected graphs, and provide  a fundamental unifying modeling tool for
statistical mechanics, probabilistic machine learning, and artificial intelligence.  
Typically MRFs are  complex and  high-dimensional   with nodes and edges (connections) built in  a modular fashion
from simpler, low-dimensional  probabilistic models and their  local connections;  in turn, this modularity  allows to  incorporate available data to MRFs and
efficiently simulate them by leveraging  their graph-theoretic structure.
Learning  graphical models from data and/or constructing them from  physical modeling and constraints necessarily involves uncertainties inherited from
data, modeling choices, or numerical approximations.  These uncertainties in the MRF can be  manifested either  in the graph structure or  the probability distribution functions, and necessarily will propagate in predictions for quantities of interest. Here we quantify such uncertainties using tight, information-based bounds on the predictions of quantities of interest; these bounds take advantage of the graphical structure of MRFs and are capable of  handling the inherent high-dimensionality of such graphical models. We demonstrate our methods in MRFs for medical diagnostics and statistical mechanics models. In the latter,  we develop uncertainty quantification bounds for finite-size effects and phase diagrams, which constitute two of the typical predictions goals  of statistical mechanics modeling.  

\end{abstract}

\begin{keywords}
  Markov Random Fields,  Uncertainty Quantification, Information Theory, Probabilistic Inequalities, Ising model, Long range interactions
\end{keywords}

\begin{AMS}
  62H22, 82B20, 94A17 
\end{AMS}
\end{tcbverbatimwrite}
\input{tmp_\jobname_abstract.tex}
%% ------------------------------------------------------------------
%% END HEADER
%% ------------------------------------------------------------------

\section{Introduction}Probabilistic graphical models (PGM) constitute one of the  fundamental  tools for Probabilistic Machine Learning (ML) and Artificial Intelligence (AI),
allowing for systematic and scalable  modeling  of uncertainty,   causality, domain knowledge, and data assimilation, \cite{GBC, KF, Probabilistic:ML:AI:2015}.
The main idea behind  PGMs is to represent complex models and associated learning processes using random variables and  their interdependence through a graph. We achieve it by constructing  structured, high-dimensional probabilistic models, involving many parameters,   nodes,
and edges,  from simpler ones  with  
 few parameters, nodes, and edges,   thus
     allowing for distributed probability computations, and by incorporating available data, exploiting
     { graph-theoretic} model representations.
PGMs are generally  classified into Markov Random Fields (MRF) defined over undirected graphs,  and Bayesian Networks, defined  over Directed Acyclical Graphs  \cite{KF} that represent conditional independencies between random variables, as well as mixtures of those two classes, \cite{Probabilistic:ML:AI:2015}. 
Furthermore, the modeling flexibility of PGMs also allows  to  combine  dynamics, data, and deep learning in Hidden Markov Models \cite{GBC, Deep_Markov:Krishnan,KingmaWelling}, as well as in recent work brings together multi-scale modeling, physical  constraints, and neural networks, \cite{Katsoulakis:JCP:BNPDE:2020, Katsoulakis:GINN:2020,FLKV}.

%In this paper, we  focus on MRFs which are defined  over undirected graphs. 
Although the term
random field may also refer to continuously indexed processes (e.g. gaussian random fields),
%the term MRF occasionally refers to continuously indexed Markov Random fields (e.g Gaussian MRFs), 
in this paper   MRFs refer to structured probabilistic  models defined on undirected graphs; %see Section~\ref{sec:CIP}, 
such PGMs are also referred to as Markov Networks. MRFs arise in  statistical mechanics where  interactions between  particles  are usually  bi-directional, or when there may be no inherent evidence for causality  (directionality) and thus undirected graphs are the appropriate structure for  such probabilistic models, \cite{GBC,KF,WJ}.
Other applications of MRFs  include  image segmentation, image denoising \cite[Sec.~4.2]{KF}, text processing \cite{TAK,PM}, bioinformatics \cite{SS}, computer vision  \cite{HZP}, Markov logic networks, \cite{DL}, Gaussian Markov networks \cite[Sec.~7.3]{KF}, artificial intelligence \cite{Probabilistic:ML:AI:2015},  and statistical mechanics \cite[Sec.~19.4]{Murphy}. Overall, MRFs  provide  a fundamental unifying modeling tool for statistical mechanics, probabilistic machine learning, and artificial intelligence, \cite{BahriGanguli,GBC}.

Learning  MRFs can be based on  available data,  e.g. for learning the graph we refer to  \cite{KF,FNP,HGJY} for score-based methods,   \cite{Murphy,HTF} for independence tests on the graph, while  maximum likelihood or Bayesian methods can be used for  parameter identification,  \cite{KF}. On the other hand, MRFs in statistical mechanics can be constructed from  physical modeling and  related  constraints, \cite{T, Murphy}.
Therefore, the learning stage of MRFs
necessarily involves uncertainties inherited from
data, modeling choices,  compromises on model  complexity, or numerical approximations.  These uncertainties in the MRF can be  manifested either  in the graph structure or  the probability distribution functions, and necessarily will propagate through the graph structure and the corresponding structured probabilistic model in the  predictions for quantities of interest (QoIs).
To understand and quantify the impact of such uncertainties on model predictions, in this paper  we present an information-based uncertainty quantification (UQ) method  for general  MRFs.
\medskip

\noindent\textbf{Model Uncertainty in Probabilistic Models:}
In general probabilistic models, uncertainties arising just from the fluctuations of the QoIs,   associated with a  given  probabilistic model  $p$, are  referred to as  aleatoric and occur when sampling $p$, \cite{CD}. They are handled by well-known  tools, e.g. central limit theorems, concentration inequalities, Bayesian posteriors, MCMC, generalized Polynomial Chaos, etc. In  contrast to this more standard type of uncertainty quantification, in MRFs, due to the learning process described earlier, we have
{\it model} uncertainties (also known as epistemic),  both in the structure (graph) and the probabilistic model itself--including parametric ones. 

Next, we briefly describe the information-theoretic formulation of model uncertainty for general probabilistic models, without assuming  any graphical model structures, see \cite{GKBW} for more details.
To practically address model uncertainty,  we typically compromise
by constructing a surrogate or approximation or {baseline} model  $p$.
 We  construct   families ${\mathcal Q}$ of  (non-parametric) {\it alternative models} $\tilde{p}$ to compare to $p$, while    the ``true" model $p^*$, which may be intractable or  partly unknown,   should belong to $\mathcal{Q}$; for this reason we can refer to ${\mathcal Q}$ as the ambiguity set, typically defined as a neighborhood of alternative models around the baseline $p$:
\begin{equation}\label{eq:ambiguity:0}
    \mathcal{Q}= \mathcal{Q}^\eta=\big\{ \tilde{p}: d(\tilde{p}, p) \leq \eta\big\}\, ,
 \end{equation}
where  $\eta >0$ corresponds to  the size of the ambiguity set  and  $d=d(\tilde{p}, p)$ denotes a probability metric or divergence.
The next  natural  mathematical goal is to assess the baseline model ``compromise" and understand the resulting biases for QoIs $f$ when we use $p$ for predictions instead of the real model $p^*\in \mathcal{Q}$. We define  the {predictive uncertainty} (or bias)  for the QoI $f$ when we use  the baseline model $p$ instead of any alternative model $\tilde{p} \in \mathcal{Q}$ (including the real one $p^*$) as the  two worst case scenarios:
\begin{equation}\label{eq:Predictive:Uncertainty:0}
        \underset{\tilde{p} \in \mathcal{Q}^\eta}{\textrm{sup/inf}}\ \left\{ E_{\tilde{p}}{f} -E_{p}{f}\right\}   \end{equation}
    where $E_{\tilde{p}}{f}$ denotes the expected value of the QoI $f$.  Therefore, \eqref{eq:Predictive:Uncertainty:0} provides {a robust performance guarantee} for the predictions of the baseline model $p$  for  $f$ within
    the ambiguity set  $\mathcal{Q}^\eta$. This robust perspective for general probabilistic models $p$ is known in Operations Research as \textit{Distributionally Robust Optimization} (DRO), e.g. \cite{GOO,GX}. 
    While the definition \eqref{eq:Predictive:Uncertainty:0}
is rather natural and  intuitive,  it is not obvious that it is  practically computable since the neighborhood   $\mathcal{Q}^\eta$ is infinite-dimensional.
However it becomes  tractable  if we use for metric $d$ in \eqref{eq:ambiguity:0}  the Kullback-Leibler (KL) divergence  $R(\tilde{p}||p)$. Accordingly, $\eta$ is a  measure of the confidence in KL we put in the baseline model $p$.
In recent work \cite{CD, DKPP, GKBW}, it has been demonstrated that \eqref{eq:Predictive:Uncertainty:0} (an infinite dimensional optimization problem) is directly computable using the variational formula (follows directly from the Donsker Varadhan variational principle, \cite{DKPP}):
\begin{equation}
\label{eq:UQ variational form}
   \underset{\tilde{p} \in \mathcal{Q}^\eta}{\textrm{sup/inf}}\ \left\{ E_{\tilde{p}}{f} -E_{p}{f}\right\}    = \pm \inf_{c>0}\Big[\frac{1}{c} \log \int e^{\pm c(f - E_{p}{f})}p(dx) + \frac{\eta}{c}\Big] \, .
\end{equation}
In this formula we recognize two main ingredients: $\eta$ is  model uncertainty from \eqref{eq:ambiguity:0} while the Moment Generating Function (MGF) $\int e^{\pm cf }p(dx)$ encodes the QoI $f$ at the baseline model $p$. In \cite{DKPP, GKBW} the authors have developed techniques to compute (exactly or approximately via asymptotics \cite{DKPP}) as well as to provide explicitly upper and lower bounds on \eqref{eq:Predictive:Uncertainty:0} in terms of concentration inequalities \cite{GKBW}. Tightness, i.e  when the $\sup$ and $\inf$ in \eqref{eq:Predictive:Uncertainty:0} are attained by an appropriate measure $\tilde{p}$   have also been studied in \cite{GKBW}. Finally, related  UQ bounds have been derived for   Markov processes using variational principles and functional inequalities \cite{BB1}, and in rare events \cite{ACD, DKPB}.

\medskip
\noindent
\textbf{Main results:}
The main  thrust of our results here is to build on  the aforementioned perspective for information-based UQ, in order to develop   UQ methods for  MRFs, and to address their specific UQ challenges.
In particular, here we address both structure (graph) and probabilistic uncertainties--including parametric ones--using tight, information-based bounds on the predictions of QoIs; although these new UQ bounds rely on \eqref{eq:Predictive:Uncertainty:0}, they  specifically, (a) take advantage of the graphical structure of MRFs,  and (b) are capable of  handling the inherent high-dimensionality of such graphical models, i.e. there is a necessity for scalable UQ in the size of the system, namely the number of nodes in MRFs  such as in the thermodynamic limit of statistical mechanics models.

Regarding the scalability issue, in  \cite{KBW} the authors tested various model uncertainty metrics in defining $d(\tilde{p}, p)$ in \eqref{eq:ambiguity:0} such as the  Hellinger distance and  $\chi^2$ divergence  and  inequalities, such as Csiszar-Kullback–Pinsker  and the Hammersley-Chapman-Robbins inequalities, \cite{T1}, in order to bound the model bias with respect of  a QoI in the spirit of \eqref{eq:UQ variational form}. It was shown that among these bounds the  only one  that scales with the  dimension of the model $p$  is \eqref{eq:UQ variational form}  and $d(\tilde{p}, p)$ should be the KL divergence.

Once we have settled to the use of the KL divergence for the aforementioned scalability reasons, we turn our attention to the baseline MRF $p$, the ambiguity set \eqref{eq:ambiguity:0} and the corresponding alternative MRFs $\tilde{p}$. Based on the earlier discussion on model uncertainty for MRFs arising from statistical learning of graph models or physical modeling, we introduce a unifying perspective of three general types of alternative models $\tilde{p}$, based on their relative structure to the baseline $p$:  Type I MRFs  where the graph structures (nodes and edges) are identical to the baseline $p$  and the parameters of probability distributions  are different, Type II where the nodes are the same, but the edges and parameters  are different. Finally, Type III where  the nodes, structure, and parameters are all different. 

 In general, MRFs satisfy the  specific conditional independence properties discussed in subsection~ \ref{subsec:CIP}. Contrary to Bayesian Networks, their distributions cannot always be factorized by a product of local conditional distributions or local functions over the graph. The celebrated {\it Hammersley-Clifford Theorem}, also known as {\it Fundamental Theorem of Random Fields}  \cite{KF, HC, Murphy}, guarantees such a factorization along maximal cliques of the graph under the assumption that $p>0$. Here, we make such an assumption for both baseline and  Type I-III MRFs. Consequently, the KL divergence is finite without requiring absolute continuity with respect to $p$.

%In Type III there can be  loss of absolute continuity between different models and KL divergence or any other f-divergences are not the right tool since they are typically infinite. We will not discuss Type III perturbations here, although the Wasserstein metric or the  $\Gamma$-Divergence, \cite{DM}, could potentially be good alternatives for the KL divergence in defining \eqref{eq:ambiguity:0}.

We take advantage of all the above and we study UQ problems by developing a unified strategy for Type I and II MRFs while Type III  is not covered here as explained in Section~\ref{subsec:MF}. We focus on the two primary ingredients of \eqref{eq:Predictive:Uncertainty:0}, namely the KL divergence and the MGF, and how they manifest themselves on MRFs.  In KL divergence, the factorization discussed earlier is a crucial tool for its simplification and numerical calculation. It allows us to compare local discrepancies in parameters and structure between the baseline $p$ and alternative models. We call these discrepancies  {\it excess factors of Type I-II} given $p$.  We develop a unifying method for computing the excess factors by interrelating the maximal cliques of alternative MRFs and the baseline MRF $p$.  As for the MGF,  the choice of QoIs is determinant. We focus on two different QoIs; those that are involved in the models (e.g. sufficient statistics) as well as characteristic functions defined on events of interest.

Regarding tightness  of UQ bounds discussed earlier, we find specific distributions that the derived UQ bounds for MRFs are attainable. In addition, we go beyond that, and pose the question: {\it Given a QoI and a baseline MRF $p$, what are the possible associated undirected graphs such that the conditional independence properties implied by the graphs are satisfied by the distributions?} Such a question introduces the concept of {\it tightness at a graph level}. There are cases where we can explicitly determine the associated graphs and others (when the structure is different than the baseline) that depend on the QoI. In the latter case, we give an example that points out a unifying method to construct the right graph or at least, a set of possible graphs.

\medskip
\noindent
\textbf{Demonstration of UQ for MRFs:}
We first  demonstrate  all the above concepts and UQ methods in a fairly simple and low dimensional MRF example from medical diagnostics. Subsequently, we implement our approach on several high-dimensional  statistical mechanics models  as they are fundamental in ML %such as Boltzmann machines
\cite{BahriGanguli,GBC}. We develop UQ  bounds for finite size effects and phase diagrams, which constitute two of the typical predictions goals  of statistical mechanics modeling and both require scalable UQ methods.

Specifically, we consider as a baseline model $p$ an  Ising-spin system with  Kac-type interactions, see \cite{errico}.  Such a model combines sufficient complexity--since  it is not a mean field model--but it is still analytically fairly  tractable to serve as a good benchmark problem for  high-dimensional MRF. Alternative models $\tilde{p}$ considered here are  1) Ising models with   perturbed  interaction potentials with respect to the baseline,   2)  models with   truncated interactions to facilitate computational implementations,  \cite{T},  and 3)   perturbations by a long-range interaction (even longer than a Kac interaction). As we discuss in Section~\ref{sec:Ex2}, these systems are typically defined in bounded domains with    boundary conditions being a given configuration outside of the domain. To have a graph description of these systems, MRFs need to be modified to account for conditioning a Gibbs distribution on an eliminated set of nodes identified as a configuration defined outside of the domain by using reduced Markov Random Fields $\mathrm{(rMRFs)}$ (see \cite{KF}). Typical questions we address in these examples include the following:  (i)  How to capture the phase diagram of a perturbed model through its comparison with the baseline phase diagram by bounding the model bias. (ii) How to truncate an interaction so as the phase diagram of the baseline model and the truncated one are close  within a prescribed tolerance.  Note that an extensive analysis on the intersection between other concepts and methods from statistical mechanics--also including non-equilibrium statistical mechanics--and deep learning have been reviewed in \cite{BahriGanguli}.

\medskip
\noindent
\textbf{Related methods:}
We  note that  existing general-purpose UQ \& sensitivity analysis methods, e.g., gradient and ANOVA-based methods, \cite{Smith, Saltelli:global, Dakota}   cannot handle UQ with model   uncertainties, due to their inherently parametric nature, while  it is not clear how they can take advantage of the graphical, causal  structure in MRFs. Furthermore, there is earlier work on model uncertainty that represents missing physics
with a stochastic noise but without the detailed structure of a graphical model, \cite{LL, TAT}.
In our work, there is a natural structure embedded in the model uncertainty,  arising  through  the graph structure of the  MRFs.

Sensitivity analysis has also a long history in statistical mechanics, known as linear and nonlinear response theory, \cite{R,BKLM}, addressing the impact of small and larger parametric  perturbations respectively. These types of methods are covered by our approach, as models with perturbed weights are clearly of Type I.

Furthermore, in contrast to these results, a  key point in our work here, also  immediately clear from  \eqref{eq:UQ variational form}, is that the model perturbations we can consider are not necessarily  small. For instance,  the parameter $\eta$  in \eqref{eq:UQ variational form} does  not  need to be  small, allowing for global and  non-parametric sensitivity analysis; the latter since the KL divergence allows us to consider models outside a specific parametric family, e.g. comparing statistical mechanics models with different potentials. Similarly, we
explicitly compute the UQ bounds for large perturbations in a medical diagnostics example. 

Sensitivity analysis in MRFs has been also studied in \cite{CD'}. The authors tackle fundamental questions such as bounding belief change between  Markov networks with the same structure but different parameter values. They propose a distance measure and  bound the relative change in probability queries by the relative change in parameters (Type I). 
Global sensitivity in parameters has been studied in \cite{BAC}. In particular, the authors developed an algorithm that checks the robustness of a MAP configuration i.e. the most likely configuration, in discrete probabilistic graphical models under global perturbations. The present work goes beyond local or global parametric sensitivity analysis that  allows us to consider perturbations  in both parameters and  edges of the graph of the MRF and examines their impact  on the prediction of specific QoIs. Special cases of our results for mean field and nearest neighbor Ising models were considered earlier in \cite{KBW}.
Finally, we note that parametric sensitivity analysis for the other class of (directed) probabilistic graphical models, namely Bayesian Networks, was developed in \cite{Darviche} using similar tools to \cite{CD'}. Parametric sensitivity analysis based on  mutual information for multi-scale partial differential equations and neural networks informed by Bayesian Network priors  was developed in \cite{UHKT} and  \cite{THKT}. Model uncertainty quantification based on information theory inequalities in the spirit of \eqref{eq:UQ variational form} were recently introduced  for Bayesian Networks arising in chemical sciences, \cite{FLKV}.

This article is organized as follows: We start with some concepts from graph theory to fix notation and then we  give a brief background of MRFs/rMRFs (Section~\ref{sec:prel1}).   Supplementary background behind rMRFs is provided in the Appendix~\ref{sec:PII}. We formally introduce the idea of graph interconnections, the impact on distributions and alternative models in Section~\ref{subsec:MF}. The main results are presented in Section~\ref{FMR} and provide UQ bounds for rMRFs, preparing the ground for applications to statistical mechanics models. In Section~\ref{sec:Applications}, we present a simple example from medical diagnostics. Section~\ref{sec:Ex2} is devoted to UQ for finite size effects, scalability, and finally UQ for phase diagrams for generic interactions and the Ising-Kac model. In the remaining sections of the Appendix, we further discuss the Ising-Kac model, we provide the technical background required for the UQ analysis of Section~\ref{sec:Ex2} (e.g Lebowitz-Penrose (LP) limit), we include the proofs of the main results, and explicit calculations of the UQ for medical diagnostics example and statistical mechanics.

\section{Preliminaries}\label{sec:prel1}\label{sec:CIP}
\subsection{Definitions from Graph Theory} We start with some  notation and terminology from graph theory.
A {\rm\bf{graph}} is a data structure $\mathcal{G}$ consisting of a set of $\mathbf{nodes}$, $\mathcal{V}=\{1,2,\dots, N\}$ and a set of {\rm\bf{edges}} $\mathcal{E}$, i.e. all pairs of nodes $i,j\in \mathcal{V}$ which are connected by an edge, denoted by $(i,j)$. An edge can be directed, denoted by $i\to j$ or undirected, denoted by $i- j$. A graph is {\rm\bf{directed}} {\rm[resp. \bf{undirected}]} if all the edges are directed [resp. undirected]. The nodes $i,j\in\mathcal{V}$ are {\rm\bf{adjacent}} if and only if $(i,j)\in\mathcal{E}$. The {\rm\bf{neighborhood}} of node $i$, denoted by $\mathcal{N}_i$ is the set of nodes which $i$ is adjacent. For sets of nodes $A, B$ and $C$,  $C$ {\rm\bf{separates}} $A$ {\rm\bf{from}} $B$, denoted by $\{i\in A\}\perp_{\mathcal{G}}\{j\in B\} \mid \{k:k\in C\}$, if and only if when we remove all the nodes in $C$ there is no path connecting any node in $A$ to any
node in $B$. Lastly, if $\mathcal{M}\subset\mathcal{V}$, the {\rm{\bf induced subgraph}}  of $\mathcal{G}$ is defined as $\mathcal{G}[\mathcal{M}]=(\mathcal{M}, \mathcal{E}')$ where $\mathcal{E}'$ includes all the edges $(i,j)\in\mathcal{E}$ such that $i,j\in\mathcal{M}$. 

\subsection{Conditional Independence Properties and MRFs}\label{subsec:CIP}
In this subsection, we define three conditional independence properties that are necessary for MRFs.
\medskip

Let $\mathcal{G}=(\mathcal{V}, \mathcal{E})$ and let $\mathbf{Y}=\{Y_i\}_{i=1}^{|\mathcal{V}|}$ be a set of random variables that each one is attached to a node and $|\mathcal{V}|$ denotes the cardinality of $\mathcal{V}$.
\medskip

\noindent$\bullet$ {\bf Pairwise Markov property (P)}: Any two non adjacent variables are conditionally independent {$\mathbf{(CI)}$} given the rest, i.e. a conditional joint can be written as
a product of conditional marginals; CI is  denoted by  $
Y_i\perp    Y_j\mid \{Y_k:k\neq i,j\}
$,\\
$\bullet$ {\bf Local Markov property (L)}: Any variable $Y_i$ is conditionally independent of all the others given its neighbors, that is $
Y_i\perp\{Y_k:k\notin\mathcal{N}_i\}    \mid \{Y_k:k\in\mathcal{N}_i\}$,\\
$\bullet$ {\bf Global Markov property (G)}: If $A, B, C$ are sets of nodes then any two sets of variables, $\mathbf{Y}_A=\{Y_i:i\in A\}$ and  $\mathbf{Y}_B=\{Y_i:i\in B\}$ are conditionally independent given a separating set of variables $\mathbf{Y}_C=\{Y_i:i\in C\}$, that is $
\mathbf{Y}_A\perp\mathbf{Y}_B\mid \mathbf{Y}_C$.

It is obvious that {\bf (G)} implies {\bf (L)} which implies {\bf (P)}.  
\begin{definition} \label{defMRF}
Let $\mathcal{G}=(\mathcal{V}, \mathcal{E})$ be an undirected graph where $\mathcal{V}=\{1,2,\dots, N\}$ is the set of nodes and $\mathcal{E}$ is the set of edges. Let also consider a set of random variables ${\mathbf{Y}}=(Y_i)_{i\in\mathcal{V}}$ indexed by $\mathcal{V}$ where each $Y_i$ takes values on a finite set $\mathcal{S}$. Their joint probability distribution is denoted by $p$. We say that 
$({\mathbf{Y}}, p)$ is a {\rm\bf{Markov Random Field (MRF)}}  iff {\bf (G)} is satisfied.
\end{definition}
As MRFs are defined on an undirected graph, it does not allow to use  chain rule of conditional probabilities and further describe the probability distribution $p(\mathbf{y})$.  A factorization rule for MRFs (i.e. for undirected graphs and the conditional independencies) is important and is provided by Hammersley and Clifford  in their unpublished work \cite{HC,G1}. To state their result, we need a few more definitions.  Let $\mathcal{G}=(\mathcal{V}, \mathcal{E})$ be a graph and let $c\subset \mathcal{V}$.
\begin{itemize}
\item[$(i)$] $c$ is called {\rm\bf{clique}} if any pair of nodes in $c$ is connected by some edge.
\item[$(ii)$] $c$ is called {\rm\bf{maximal clique}} if any superset $c'$ of $c$ (i.e $c'\supset c$) is not a clique any more. The set of all maximal cliques of graph $\mathcal{G}$ is denoted by $\mathcal{C}_{\mathcal{G}}$.
\end{itemize}
\smallskip

\noindent{\bf Hammersley-Clifford Theorem} {\it A positive distribution $p(\mathbf{y})> 0$ satisfies one of} {\bf (P), (L)} {\it and} {\bf (G)} {\it of an undirected graph G iff $p$ parametrized by some parameters $\mathbf{w}=\{\mathbf{w}_c\}_{c\in\mathcal{C}_{\mathcal{G}}}$ can be represented as a product of clique potentials, i.e}
\begin{equation}\label{distributioncliques}
P_{\Psi}^{\mathbf{w}}(\mathbf{y})\equiv p(\mathbf{y}\mid \mathbf{w})=\frac{1}{Z(\mathbf{w})}\prod_{c\in\mathcal{C}_{\mathcal{G}}}\Psi_{c}(\mathbf{y}_{c}\mid\mathbf{w}_c)
\end{equation}
{\it where $\Psi_{c}(\mathbf{y}_{c}\mid\mathbf{w}_c)$ is a positive function defined on the random variables in clique $c$ and parametrized by some parameters $\mathbf{w}_c$, and is called {\rm \bf{clique potential}}. Also $Z(\mathbf{w})$ is the partition function given by}
\begin{equation}\label{partitioncliques}
Z(\mathbf{w})=\sum_{\mathbf{y}}\prod_{c\in\mathcal{C}_{\mathcal{G}}}\Psi_{c}(\mathbf{y}_{c}\mid\mathbf{w}_c)
\end{equation}

The theorem states that the set of all joint distributions on an undirected graph $\mathcal{G}$ that can be factorized as in \eqref{distributioncliques} is identical to the set of joint distributions that satisfy the conditional independence properties, under the restriction of strictly positive distributions.
%The one direction of the theorem (the simplest) states that if $p$ is a probability distribution over $\mathbf{Y}$ with its components being attached to the nodes of an undirected graph $\mathcal{G}$ and $p$ factorizes over the maximal cliques of $\mathcal{G}$ (such $p$ is called {\it Gibbs distribution}), then $p$ satisfies at least one of {\bf (P), (L)} and {\bf (G)} as implied by the strictly positive probability. Hence $(\mathbf{Y},p)$ is an MRF. 
%
%In the other direction, if $p>0$ (i.e all events are possible) and $(\mathbf{Y},p)$ is an MRF, the Hammersley-Clifford Theorem guarantees a factorization over maximal cliques.  
\begin{remark}
Without the assumption of strict positiveness of the joint distribution $p$, the theorem is not valid. A counterexample has been obtained in \cite{moussouris}.
\end{remark}

\begin{remark}
The KL divergence or any other f-divergences between a baseline MRF that is assumed nonnegative and alternative MRFs of Type II-III ( different structure, see Introduction) could be infinite due to the loss of absolute continuity. In that case,  the Wasserstein metric or the  $\Gamma$-Divergence, \cite{DM}, could potentially be good alternatives for the KL divergence in defining \eqref{eq:ambiguity:0}. The implementation of the Wasserstein metric or the  $\Gamma$-Divergence is still unexplored in the context of such MRFs. For this purpose, the development of new methods constitutes an important step towards comparing MRFs with different structures and nonnegative distributions. In this article, we restrict our attention to the Hammersley-Clifford Theorem and we assume strictly positive probability distributions.
\end{remark}

Given a MRF $({\mathbf{Y}}, p)$, a reduced Markov Random Field (rMRFs) is obtained by conditioning  $p$ on some observation ${\mathbf{U}}={\mathbf{u}}$ with ${\mathbf{U}}\subset {\mathbf{Y}}$. Hence, the distribution of the resulting rMRF has a reduced number of clique potentials.  As we discuss in Section~\ref{sec:Ex2}, rMRFs are appropriate for formulating statistical mechanics models defined on bounded domains with a given configuration outside of the domain in a graph language. Next, we formally introduce rMRFs.

\subsubsection{\bf Reduced Markov Random Fields (rMRFs)}\label{MRFformulation}Let $\mathbf{Y}=\{Y_i\}_{i\in \mathcal{V}}$  be a collection of random variables indexed by a set of nodes $\mathcal{V}$ of a graph $\mathcal{G}=( \mathcal{V},  \mathcal{E})$, taking values in some space $\mathbf{\mathcal{Y}}^{\mathcal{V}}=\otimes_{i=1}^{\mathcal{V}}\mathcal{Y}_i$. Let $p\equiv p(\cdot |\mathbf{w})$ be a strictly positive joint probability distribution of $\mathbf{Y}$ parametrized by $\mathbf{w}$ such that $(\mathbf{Y}, p(\cdot |\mathbf{w}))$ is a MRF.  

Let $\mathbf{u}$ be a context  and $\mathcal{M}\subset\mathcal{V}$. If $\mathbf{U}=\{Y_i\}_{i\in\mathcal{M}}$ with $\mathbf{U}=\mathbf{u}$, we construct the corresponding rMRF as follows: let $\mathbf{Z}=\{Y_i\}_{i\in\mathcal{V}\setminus\mathcal{M}}$  and $q(\mathbf{z} |\mathbf{w})$ be the probability distribution factorized according to Proposition~\ref{thmrMRF}  (the analogue of the Hammersley-Clifford Theorem for rMRFs): $q(\mathbf{z})\equiv q(\mathbf{z} |\mathbf{w})=\frac{1}{Z_{\mathbf{u}}(\mathbf{w})}\prod_{c\in\mathcal{C}_{\mathcal{G}}}\Psi_{c}[\mathbf{u}](\mathbf{z}_{c}\mid\mathbf{w}_c)$.
More details on rMRFs are given in Appendix~\ref{sec:PII}.

\medskip

\noindent The next two sections are presented for rMRFs as we can then recall formulas and the main results directly in the UQ analysis of statistical mechanics models in Section~\ref{sec:Ex2}. Their formulation and analysis hold for MRFs and when required, we will be providing more details for their implementation to MRFs.

%In general, a probability distribution may factorize over MRFs, i.e. the clique potentials are defined on complete subgraphs. This can be done in many different ways. However, for strictly positive probability distributions, the Hammersley-Clifford theorem guarantees a factorization over maximal cliques that  takes advantage of the MRF graph structure. This works as a benchmark in the approach we develop. In fact, the factorized distribution itself includes all the information coming from the graph independencies. Later, we use this idea to develop a method of interrelating MRFs, which in turn makes KL divergence well-controlled, see Section~\ref{FMR}.

\section{Mathematical Formulation of UQ on MRFs/rMRFs}\label{subsec:MF} Let $q$ be a rMRF constructed by learning from available data or from physical modeling and related constraints. Constructing such a model involves uncertainties either in the graph structure or the probability distribution
functions, and necessarily will propagate through the graph structure and the
corresponding structured probabilistic model in the predictions for QoIs. We quantify the impact of such uncertainties on model predictions by constructing ambiguity sets such as \eqref{eq:ambiguity:0} consisting of alternative rMRFs given by \begin{equation}\label{eq:ambiguity:*}
    \mathcal{Q}^\eta=\big\{ {\textrm{rMRFs}} \;\;\tilde{q}: R(\tilde{q}\|q) \leq \eta\big\}\, ,
 \end{equation}
where  $\eta >0$ corresponds to  the size of the ambiguity set. The alternative models $\tilde{q}$ in \eqref{eq:ambiguity:*} can be classified into: Type I MRFs  where the graph structures (nodes and edges) are identical to the baseline $q$  and the parameters of probability distributions  are different, Type II where the nodes are the same, but the edges and parameters  are different. Finally, Type III where  the nodes, structure, and parameters are all different. Next, we mathematically formulate the alternative models. 

\subsection{ Alternative models}\label{subsec:altm}Let
$
(\mathcal{G},\mathbf{w},p)$ and $(\tilde{\mathcal{G}},\tilde{\mathbf{w}},\tilde{p})
$
be two MRFs with $ \mathcal{G}=(\mathcal{V},\mathcal{E})$ and $ \tilde{\mathcal{G}}=(\tilde{\mathcal{V}},\tilde{\mathcal{E}})$ being the associated graphs, where $\mathcal{V}$ and $\tilde{\mathcal{V}}$ are the sets of nodes and $\mathcal{E}$ and $ \tilde{\mathcal{E}}$ are the sets of edges. 
\begin{definition}
    \label{def:types}
$(\tilde {\mathcal{G}},\tilde{\mathbf{w}},\tilde{p})$ and $(\mathcal{G},\mathbf{w},p)$ can have one of the following interconnections:
\begin{itemize}
\item[] {\rm{Type I:}} $\qquad\tilde{\mathcal{V}}={\mathcal{V}}$,  $\tilde{\mathcal{E}}={\mathcal{E}}$ and  $\tilde{\mathbf{w}}\neq\mathbf{w}$, or

\medskip

\item[] {\rm Type II:} $\qquad\tilde{\mathcal{V}}={\mathcal{V}}$,  $\mathcal{E}\subset\tilde{\mathcal{E}}$ and  $\tilde{\mathbf{w}}\neq\mathbf{w}$ or

\medskip

\item[] {\rm Type III:}$\qquad\tilde{\mathcal{V}}\neq{\mathcal{V}}$,  $\mathcal{E}\neq\tilde{\mathcal{E}}$ and  $\tilde{\mathbf{w}}\neq\mathbf{w}$.
\end{itemize}
\end{definition}
From now on, we refer to the baseline model when we use the notation $(\mathcal{G},\mathbf{w},p)$ and without loss of generality we assume $\mathcal{E}\subset\tilde{\mathcal{E}}$.  This assumption simplifies the presentation of our approach but  intuitively speaking, the fewer edges a rMRF has, the more information it provides, since in a sparser graph, there are more conditional independencies specified. 

Based on that, we interrelate the maximal cliques of Type I-II models with those of $p$. In particular, for  Type I  there is one to one correspondence between maximal cliques. Changes on the set of edges of a Type II model lead to different sets of maximal cliques and one needs to examine the nature of the new edges and their impact on the maximal cliques of $p$. Finally, the new set of nodes of a Type III model leads to a drastically new structure that makes such interrelation of maximal cliques hard to achieve. Therefore, this case is not examined here.
\medskip

 Let $\mathbf{u}$ be a context  and $\mathcal{M}\subset\mathcal{V}\cap \tilde{\mathcal{V}}$. For $\mathbf{U}=\{Y_i\}_{i\in\mathcal{M}}$ with $\mathbf{U}=\mathbf{u}$, we construct the corresponding rMRFs $(\mathbf{Z}, q(\cdot |{\mathbf{w}}))
$ and $(\tilde{\mathbf{Z}}, \tilde{q}(\cdot |\tilde{\mathbf{w}}))$ parametrized by ${\mathbf{w}}$ and $\tilde{\mathbf{w}}$ respectively. Based on the structural classification Type I-III, the probability distributions of $\tilde{q}$ are treated as follows:

\medskip

% Under the same collection of random variables, we consider alternative with $\tilde{p}(\cdot |\tilde{\mathbf{w}})>0$,  parametrized by $\tilde{\mathbf{w}}$ so as $(\mathbf{Y}, \tilde{p}(\cdot |\tilde{\mathbf{w}}))$ is a MRF. Given $\mathbf{u}$  and $\mathcal{M}\subset\mathcal{V}$ as before, we construct the corresponding rMRF  $\left(\mathbf{Z}, \tilde{q}(\cdot |\tilde{\mathbf{w}})\right)
%$ with the probability joint distribution, $\tilde{q}(\cdot)\equiv\tilde{q}(\cdot |\tilde{\mathbf{w}})$. The corresponding partition function and clique potentials are denoted by $\tilde{Z}_{\mathbf{u}}(\tilde{\mathbf{w}})$ and $\tilde{\Psi}_{\tilde{c}}[\mathbf{u}](\mathbf{z}_{\tilde{c}}\mid\tilde{\mathbf{w}}_{\tilde{c}})$.

\noindent {\bf  Type I.} Let $\mathcal{B}\subset\mathcal{C}_{\mathcal{G}}$ be the set of maximal cliques whose weights differ, i.e for each $
c\in\mathcal{B},\;\;\tilde{\mathbf{w}}_c\neq\mathbf{w}_c.
$
The clique potentials of $\tilde{q}(\cdot |\tilde{\mathbf{w}})$ can be rewritten as
\begin{equation}\label{AsA}
 \tilde{\Psi}_{c}[\mathbf{u}](\mathbf{z}_{c}\mid\tilde{\mathbf{w}}_c )=
\left\{
    \begin{array}{ll}
        \Psi_{c}[\mathbf{u}](\mathbf{z}_{c}\mid\mathbf{w}_{c})\Phi_c[\mathbf{u}] (\mathbf{z}_{c}\mid\mathbf{w}_{c},\tilde{\mathbf{w}}_{c})& ,\mbox{if } c\in \mathcal{B} \\
        \Psi_{c}[\mathbf{u}](\mathbf{z}_{c}\mid\mathbf{w}_c) & ,\mbox{otherwise}.
    \end{array}
\right.
\end{equation}
We call $\Phi_c[\mathbf{u}] (\cdot\mid\tilde{\mathbf{w}}_{c},\mathbf{w}_{c})>0$   {\it $\tilde{q}$-excess factor of type I relative to $q$ on $c$} and is defined on variables $\mathbf{z}_{c}$ in clique $c\in\mathcal{B}$. Cliques where no change on  weights has occurred, remain the same.
\medskip

\begin{figure}[h]  
\begin{tikzpicture}[ node distance=1.4cm,
                   main node/.style={circle,draw,font=\sffamily\tiny\bfseries}]
  \node[main node,scale=0.4, white] (1) [top color =white , bottom color = black]{1};
  \node[main node,scale=0.4,white] (2) [top color =white , bottom color = black,above right of=1] {2};
  \node[main node,scale=0.4,white] (3) [top color =white , bottom color = black,below right of=1] {3};
  \node[main node,scale=0.4,white] (4) [top color =white , bottom color = black,right of=3] {4};
  \node[main node,scale=0.4,white] (5) [top color =white , bottom color = black,above right of=4] {5};
  \node[main node,scale=0.4,white] (6) [top color =white , bottom color = black,below right of =4] {6};
  \node[main node,scale=0.4,white] (7) [top color =white , bottom color = black,below left of=6] {7};
  \node[main node,scale=0.4,white] (8) [top color =white , bottom color = black,above right of=5] {8};
  \node[main node,scale=0.4,white] (9) [top color =white , bottom color = black,above right of=8] {9};
  \node[main node,scale=0.4,white] (10) [top color =white , bottom color = black,below right of=8] {10};

  \path[every node/.style={font=\sffamily\small}]
    (1) edge[ultra thick] node [yellow,] {} (2)
    (2) edge[ultra thick] node [] {} (3)
    %(3) edge[gray!30!white] node [] {} (4)
    (3) edge[ultra thick] node [] {} (4)
    (3) edge[ultra thick] node [] {} (7)
    (3) edge[ultra thick] node [] {} (6)
    %(9) edge[gray!30!white] node [] {} (8)
    (9) edge[ultra thick] node [] {} (8)
    (5) edge[ultra thick]  node [] {} (8)
    (5) edge[ultra thick]  node [] {} (6)
    (9) edge[ultra thick] node [] {} (10)
    %(9) edge[gray!30!white] node [] {} (10)
    %(4) edge[gray!30!white] node [] {} (6)
    %(4) edge[gray!30!white] node [] {} (5)
    (4) edge[ultra thick] node [] {} (6)
    (4) edge[ultra thick] node [] {} (5)
    (6) edge[ultra thick] node [] {} (7)
    (8) edge[ultra thick]  node [] {} (10)
    (5) edge[ultra thick]  node [] {} (8);    
\end{tikzpicture}
\hspace{1cm}
\begin{tikzpicture}[node distance=1.4cm,
                   main node/.style={circle,draw,font=\sffamily\tiny\bfseries}]
 \node[main node,scale=0.4, white] (1) [top color =white , bottom color = black]{1};
  \node[main node,scale=0.4,white] (2) [top color =white , bottom color = black,above right of=1] {2};
  \node[main node,scale=0.4,white] (3) [top color =white , bottom color = black,below right of=1] {3};
  \node[main node,scale=0.4,white] (4) [top color =white , bottom color = black,right of=3] {4};
  \node[main node,scale=0.4,white] (5) [top color =white , bottom color = black,above right of=4] {5};
  \node[main node,scale=0.4,white] (6) [top color =white , bottom color = black,below right of =4] {6};
  \node[main node,scale=0.4,white] (7) [top color =white , bottom color = black,below left of=6] {7};
  \node[main node,scale=0.4,white] (8) [top color =white , bottom color = black,above right of=5] {8};
  \node[main node,scale=0.4,white] (9) [top color =white , bottom color = black,above right of=8] {9};
  \node[main node,scale=0.4,white] (10) [top color =white , bottom color = black,below right of=8] {10};

  \path[every node/.style={font=\sffamily\small}]
    (1) edge[ultra thick] node [yellow,] {} (2)
    (2) edge[ultra thick] node [] {} (3)
       (3) edge[ultra thick] node [] {} (4)
    (3) edge[ultra thick] node [] {} (7)
    (3) edge[ultra thick] node [] {} (6)
  
    (9) edge[ultra thick] node [] {} (8)
    (5) edge[ultra thick]  node [] {} (8)
    (5) edge[ultra thick]  node [] {} (6)
    (9) edge[ultra thick] node [] {} (10)
        (4) edge[ultra thick] node [] {} (6)
    (4) edge[ultra thick] node [] {} (5)
    (6) edge[ultra thick] node [] {} (7)
    (8) edge[ultra thick]  node [] {} (10)
    (5) edge[ultra thick]  node [] {} (8)   
    (4) edge[ultra thick,yellow!90!black]  node [] {} (7);   
\end{tikzpicture}
\hspace{1cm}
\begin{tikzpicture}[node distance=1.4cm,
                   main node/.style={circle,draw,font=\sffamily\tiny\bfseries}]
 \node[main node,scale=0.4, white] (1) [top color =white , bottom color = black]{1};
  \node[main node,scale=0.4,white] (2) [top color =white , bottom color = black,above right of=1] {2};
  \node[main node,scale=0.4,white] (3) [top color =white , bottom color = black,below right of=1] {3};
  \node[main node,scale=0.4,white] (4) [top color =white , bottom color = black,right of=3] {4};
  \node[main node,scale=0.4,white] (5) [top color =white , bottom color = black,above right of=4] {5};
  \node[main node,scale=0.4,white] (6) [top color =white , bottom color = black,below right of =4] {6};
  \node[main node,scale=0.4,white] (7) [top color =white , bottom color = black,below left of=6] {7};
  \node[main node,scale=0.4,white] (8) [top color =white , bottom color = black,above right of=5] {8};
  \node[main node,scale=0.4,white] (9) [top color =white , bottom color = black,above right of=8] {9};
  \node[main node,scale=0.4,white] (10) [top color =white , bottom color = black,below right of=8] {10};

  \path[every node/.style={font=\sffamily\small}]
    (1) edge[ultra thick] node [yellow,] {} (2)
    (2) edge[ultra thick] node [] {} (3)
        (3) edge[ultra thick] node [] {} (4)
    (3) edge[ultra thick] node [] {} (7)
    (3) edge[ultra thick] node [] {} (6)
        (9) edge[ultra thick] node [] {} (8)
    (5) edge[ultra thick]  node [] {} (8)
    (5) edge[ultra thick]  node [] {} (6)
    (9) edge[ultra thick] node [] {} (10)
    
    (4) edge[ultra thick] node [] {} (6)
    (4) edge[ultra thick] node [] {} (5)
    (6) edge[ultra thick] node [] {} (7)
    (8) edge[ultra thick]  node [] {} (10)
    (5) edge[ultra thick]  node [] {} (8)   
    (6) edge[ultra thick, red]  node [] {} (10);   
\end{tikzpicture}
\hspace{1cm}
\begin{tikzpicture}[node distance=1.4cm,
                   main node/.style={circle,draw,font=\sffamily\tiny\bfseries}]
  \node[main node,scale=0.4, white] (1) [top color =white , bottom color = black]{1};
  \node[main node,scale=0.4,white] (2) [top color =white , bottom color = black,above right of=1] {2};
  \node[main node,scale=0.4,white] (3) [top color =white , bottom color = black,below right of=1] {3};
  \node[main node,scale=0.4,white] (4) [top color =white , bottom color = black,right of=3] {4};
  \node[main node,scale=0.4,white] (5) [top color =white , bottom color = black,above right of=4] {5};
  \node[main node,scale=0.4,white] (6) [top color =white , bottom color = black,below right of =4] {6};
  \node[main node,scale=0.4,white] (7) [top color =white , bottom color = black,below left of=6] {7};
  \node[main node,scale=0.4,white] (8) [top color =white , bottom color = black,above right of=5] {8};
  \node[main node,scale=0.4,white] (9) [top color =white , bottom color = black,above right of=8] {9};
  \node[main node,scale=0.4,white] (10) [top color =white , bottom color = black,below right of=8] {10};

  \path[every node/.style={font=\sffamily\small}]
    (1) edge[ultra thick] node [yellow,] {} (2)
    (2) edge[ultra thick] node [] {} (3)
        (3) edge[ultra thick] node [] {} (4)
    (3) edge[ultra thick] node [] {} (7)
    (3) edge[ultra thick] node [] {} (6)
       (9) edge[ultra thick] node [] {} (8)
    (5) edge[ultra thick]  node [] {} (8)
    (5) edge[ultra thick]  node [] {} (6)
    (9) edge[ultra thick] node [] {} (10)
        (4) edge[ultra thick] node [] {} (6)
    (4) edge[ultra thick] node [] {} (5)
    (6) edge[ultra thick] node [] {} (7)
    (8) edge[ultra thick]  node [] {} (10)
    (5) edge[ultra thick]  node [] {} (8)   
    (5) edge[ultra thick, blue]  node [] {} (10);   
\end{tikzpicture}
\caption{\small{ (First) Baseline MRF model $p$ demonstrated by graph $\mathcal{G}$. (Second) Alternative model $\tilde{p}$ with the associated graph obtained by adding the yellow edge $(4-7)$ and connecting two maximal cliques of $p$ model, $\{3,4,6\}$ and $\{3,6,7\}$, thus $\tilde{p}$ has a new maximal clique $\{3,4,6,7\}$. (Third) Alternative model $\tilde{p}$ with the associated graph obtained by adding the red edge $(6-10)$ and thus $\tilde{p}$ has a totally new maximal clique $\{6,10\}$. (Forth) Alternative model $\tilde{p}$ with the associated graph obtained by adding the blue edge $(5-10)$ and enlarging the already existing clique, $\{5,8\}$, to $\{5,8,10\}$.}}\label{fig:K1}  
\end{figure}
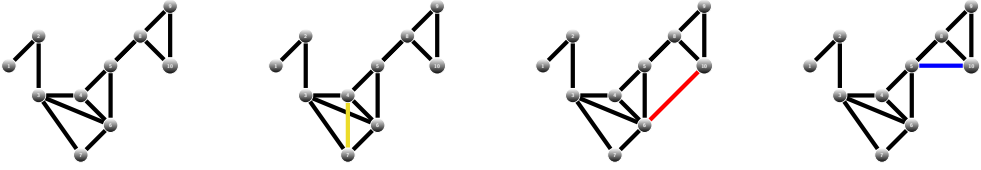

\noindent{\bf Type II.} In this type, the class of maximal cliques $\mathcal{C}_{\tilde{\mathcal{G}}}$ is different. The analysis becomes more complicated and clique potentials need to be carefully  considered. We look into the nature of one or more new edges by categorizing it as one of the following types: a new edge (i) can create a totally new maximal clique, see Figure~\ref{fig:K1}, third graph, (ii) can connect two or more already existing maximal cliques, see Figure~\ref{fig:K1}, second graph, and (iii) can enlarge an already existing maximal clique, see Figure~\ref{fig:K1}, forth graph.
By adding more than one new edges, the new maximal cliques of $\tilde{\mathcal{G}}$ can be obtained by a combination of $(i)$, $(ii)$, and $(iii)$. We introduce the following sets:
 \begin{align}
 \mathcal{B}_{\cup}&=\{\tilde{c}\in\mathcal{C}_{\tilde{\mathcal{G}}}\setminus\mathcal{C}_{\mathcal{G}}: \tilde{c}=\cup_{i}c_i,\mbox{ for }c_i\in\mathcal{C}_{\mathcal{G}}\}\label{bset}\\
 \mathcal{B}_{\subseteq}&=\{\tilde{c}\in\mathcal{C}_{\tilde{\mathcal{G}}}\setminus\mathcal{C}_{\mathcal{G}}: \mbox{there exists } c\in\mathcal{C}_{\mathcal{G}}\mbox{ s.t. } c\subseteq\tilde{c}\}\\
 \mathcal{B}_{\rm{new}}&=(\mathcal{C}_{\mathcal{G}}\cup\mathcal{B}_{\cup}\cup\mathcal{B}_{\subseteq})^{c}
 \end{align}
 Then the clique potentials of  $\tilde{q}$ can be rewritten as:
\begin{equation}\label{AsB}
 \tilde{\Psi}_{\tilde c}[\mathbf{u}](\mathbf{z}_{\tilde{c}}\mid\tilde{\mathbf{w}}_{\tilde{c}} )=
\left\{
    \begin{array}{ll}
        \prod_{c_i}\Psi_{c_i}[\mathbf{u}](\mathbf{z}_{c_i}\mid\mathbf{w}_{c_i})\Phi_{\tilde{c}}^{(ii)}[\mathbf{u}] (\mathbf{z}_{\tilde{c}}\mid\mathbf{w}_{c},\tilde{\mathbf{w}}_{c})  & ,\mbox{if } \tilde{c}\in \mathcal{B}_{\cup},\\
        \Psi_{c}[\mathbf{u}](\mathbf{z}_{c}\mid\mathbf{w}_c)\Phi_{\tilde{c}}^{(iii)}[\mathbf{u}] ((\mathbf{z}_{\tilde{c}}\mid\mathbf{w}_{c},\tilde{\mathbf{w}}_{c}) &  ,\mbox{if } \tilde{c}\in \mathcal{B}_{\subset},\\
            \tilde{\Psi}_{\tilde c}[\mathbf{u}](\mathbf{z}_{\tilde{c}}\mid\tilde{\mathbf{w}}_{\tilde{c}} )&  ,\mbox{if } \tilde{c}\in\mathcal{B}_{\rm{new}}\\
                \Psi_{c}[\mathbf{u}](\mathbf{z}_{c}\mid\mathbf{w}_c) &  ,\mbox{if } \tilde{c}\in\mathcal{C}_{\tilde{\mathcal{G}}}
            \end{array}
\right.
\end{equation}
We call  $\Phi_{\tilde{c}}^{(ii)},\Phi_{\tilde{c}}^{(iii)}>0$  {\it $\tilde{q}$-excess factors of type II relative to $q$ on $\tilde{c}$} defined on the variables of $\tilde{c}$. In fact, the two functions play the role of the discrepancy  at a distribution level when new maximal clique $\tilde{c}$ has been created by connecting existing maximal cliques $c_i$ and by enlarging an existing maximal clique. When $\tilde{c}\in\mathcal{B}_{\rm{new}}$, there is no need to express the clique potential through the potentials of $q(\cdot\mid \mathbf{w})$. For simplicity, we assume that clique potentials on common maximal cliques between $\mathcal{G}$ and $\tilde{\mathcal{G}}$ do not change. However, one can consider  different potentials and in that case, a term $\Phi$ should be introduced similar to (ii) and (iii). For convenience, we establish one last unifying  terminology. We call
\begin{align}
    \label{eq:psi}
\Phi_{\mathbf{u}}^{\mathrm{I}}(\mathbf{Z})&:={\prod_{c\in\mathcal{B}}}\Phi_c[\mathbf{u}] (\mathbf{Z}_{c})\\\label{eq:phi}
\Phi_{\mathbf{u}}^{\mathrm{II}}(\mathbf{Z})&:=\prod_{\tilde{c}\in\mathcal{B}_{\mathrm{new}}} \tilde{\Psi}_{\tilde c}[\mathbf{u}](\mathbf{Z}_{\tilde{c}}\mid\tilde{\mathbf{w}}_{\tilde{c}} )\prod_{\tilde{c}\in\mathcal{B}_{\cup}}\Phi_{\tilde{c}}^{(ii)}[\mathbf{u}] ({\mathbf{Z}}_{\tilde{c}})\prod_{\tilde{c}\in\mathcal{B}_{\subseteq}}\Phi_{\tilde{c}}^{(iii)}[\mathbf{u}] ({(\mathbf{Z}_{\tilde{c}})}\mid\tilde{\mathbf{w}}_{\tilde{c}})
\end{align}
{\it total $\tilde{q}$-excess factor of type I and II relative to $q$} respectively. The total $\tilde{q}$-excess factor of type I relative to $q$ captures all the parameters changes while the total $\tilde{q}$-excess factor of type II relative to $q$ captures all the structural discrepancies.  In the case of MRF, we drop the context $\mathbf{u}$ from \eqref{eq:psi} and \eqref{eq:phi} and $\mathbf{Z}$ is replaced by $\mathbf{Y}$.   Equations \eqref{bset}-\eqref{eq:phi} are explicitly specified in medical diagnostics application in Section~\ref{sec:Applications} and its detailed analysis in Appendix~\ref{App:MedicalDiagnostics}, and in statistical mechanics, Section~\ref{sec:Ex2}. In Type III, there exists the total $\tilde{q}$-excess factor of type III relative to $q$. However, due to the high degree of discrepancies, we cannot  interrelate maximal cliques of Type III model with $q$, and by extension each $\tilde{q}$-excess factor cannot be determined. The next results are straightforward but essential in our calculations.  To avoid heavy notation, we remind that $q(\cdot)=q(\cdot\mid\mathbf{w})$ and $\tilde{q}(\cdot)=\tilde{q}(\cdot\mid\tilde{\mathbf{w}})$.
\medskip

\noindent {\bf Partition function of alternative models.} Based on the above description of alternative models,  the partition function of $\tilde{q}$ is given in the next lemma.
\begin{lemma}\label{lem:rationPart}
Let $(\mathbf{Z},q)$ be a rMRF. Then for any alternative rMRF $(\mathbf{Z},\tilde{q})$ of Type $\mathrm{i}$  with $\mathrm{i}=\mathrm{I},\mathrm{II}$ its partition function is expressed as:
\begin{equation}
\tilde{Z}_{\mathrm{u}}(\tilde{\mathbf{w}})=E_{q}[\Phi_{\mathbf{u}}^{\mathrm{i}}]Z_{\mathrm{u}}(\mathbf{w})
\end{equation}
where $\Phi_{\mathbf{u}}^{\mathrm{i}}$ are given by \eqref{eq:psi} and \eqref{eq:phi}.
\end{lemma}
\begin{proof}The proof is based on the method of interrelating the distribution $q$ and $\tilde{q}$, utilizing the total $\tilde{q}$-excess factors relative to $q$ given by \eqref{eq:psi} and \eqref{eq:phi}. The explicit computation is provided in Appendix~\ref{prooflem:rationPart}.
\end{proof}
\medskip

\noindent {\bf Likelihood ratio.} The following lemma provides the likelihood ratio between $\tilde{q}$ and $q$ and constitutes the key ingredient for the simplification of \eqref{eq:mgf,RL} and the UQ bounds provided in \eqref{probUQ2}.
\begin{lemma}\label{thm:fracdistrib}
Let $(\mathbf{Z},q)$ be a rMRF. Then for any alternative rMRF $(\mathbf{Z},\tilde{q})$ of Type $\mathrm{i}$  with $\mathrm{i}=\mathrm{I},\mathrm{II}$,  the corresponding likelihood ratio satisfies:
\begin{equation}\label{eq:rationmeas}
    \frac{d\tilde{q}}{d q}=\frac{\Phi_{\mathbf{u}}^{\mathrm{i}}}{E_{q}[\Phi_{\mathbf{u}}^{\mathrm{i}}]}
\end{equation}
 where $\Phi_{\mathbf{u}}^{\mathrm{i}}$ is  given by \eqref{eq:psi} and \eqref{eq:phi}.
\end{lemma}
The proof is omitted as the lemma is a direct consequence of the method of interrelating two distributions discussed above and Lemma~\ref{lem:rationPart}. Note that both results hold for MRFs denoted by $(\mathbf{Y},p)$ and $(\mathbf{Y},\tilde{p})$, dropping the context $\mathbf{u}$ from $\Phi_{\mathbf{u}}^{\mathrm{i}}$ in \eqref{eq:rationmeas}.
%\begin{expl}
%In the medical diagnostics example, $\Phi^{\mathrm{i}}(\mathbf{Y})$ are given by \eqref{eq:phiEx1} and \eqref{eq:phiEx12}, representing the total excess factor after reducing the weight on $\{1,2\}$ by 20\% and after adding a new link between smoking and cough $1-4$ respectively. We note that in the latter case, $\mathcal{B}_{\rm{new}}=\mathcal{B}_{\cup}=\emptyset$ and $\mathcal{B}_{\subseteq}=\{\{1,2,4\}\}$. 
%\end{expl}

 \subsection{KL divergence}\label{subsec:mrKL}
As we see in Section~\ref{FMR}, our UQ  methods  rely  on the  KL divergence as means  to measure  ``distance" between baseline and alternative MRFs.  The fact that it scales  correctly with the dimension of the baseline model \cite{DKPP} as well as the commonalities in parameters and structure between baseline and alternative models combined with the Hammersley-Clifford Theorem allows the KL divergence to be expressed in a simplified and informative form. In particular, we show that KL divergence (which is finite due to the positive probabilities $q$ and $\tilde{q}$) depends only on the total $\tilde{q}$-excess factor relative to $q$ given by \eqref{eq:psi} and \eqref{eq:phi}. To simplify the notation, we omit the dependence of $\mathbf{Z}$ from $\kappa_i, f$ and  $\Phi_{\mathbf{u}}^{\mathrm{i}}$.
\begin{lemma}\label{lemma:RE}
 Let $(\mathbf{Y}, p^{\mathbf{w}})$, $ (\mathbf{Y}, \tilde{p}^{\,\tilde{\mathbf{w}}})$ be two MRFs defined over graphs $\mathcal{G}=( \mathcal{V},  \mathcal{E})$ and $\tilde{\mathcal{G}}=( \mathcal{V},  \tilde{\mathcal{E}})$ respectively. Let $\mathbf{u}$ be a context and $\mathcal{M}\subset\mathcal{V}$.  We consider the corresponding rMRFs $(\mathbf{Z}, q), (\mathbf{Z}, \tilde{q})$.

a. If $\tilde{q}$ is Type $\mathrm{i}$, with $\mathrm{i}=\mathrm{I}$ or $\mathrm{II}$, then the KL divergence is given \begin{eqnarray}\label{eq:KLnew}
R(\tilde{q}\|q)&=&E_{\tilde{q}}\left[\log\frac{\tilde{q}}{q}\right]=E_{q}\left[\frac{\tilde{q}}{q}\log\frac{\tilde{q}}{q}\right]\nonumber\\
&=&E_{\tilde{q}}[\log\Phi_{\mathbf{u}}^{\mathrm{i}}]-\log E_{q}[\Phi_{\mathbf{u}}^{\mathrm{i}}]=\frac{1}{E_{q}[\Phi_{\mathbf{u}}^{\mathrm{i}}]}E_{q}\left[\Phi_{\mathbf{u}}^{\mathrm{i}}\log\Phi_{\mathbf{u}}^{\mathrm{i}}\right]-\log E_{q}[\Phi_{\mathbf{u}}^{\mathrm{i}}]
\end{eqnarray}
where $\Phi_{\mathbf{u}}^{\mathrm{i}}$ is defined in \eqref{eq:psi} and \eqref{eq:phi} accordingly.

b. If $\tilde{q}$ is Type $\mathrm{i}$, with $\mathrm{i}=\mathrm{I}$ or $\mathrm{II}$, then for any $f$ satisfying \eqref{qoitypeI}, the KL divergence is given by
\begin{equation}\label{eq:RE}
\mathcal{R}(\tilde{q}\|q)=C_{\mathrm{i}}E_{\tilde{q}}[f]+\frac{E_{q}\left[\kappa_{\mathrm{i}}\Phi_{\mathbf{u}}^{\mathrm{i}}\right]}{E_{q}\left[\Phi_{\mathbf{u}}^{\mathrm{i}}\right]}-\log E_{q}\left[\Phi_{\mathbf{u}}^{\mathrm{i}}\right],\;\; \Phi_{\mathbf{u}}^{\mathrm{i}}(\mathbf{Z})=e^{C_{\mathrm{i}} f(\mathbf{Z})+\kappa_{\mathrm{i}}(\mathbf{Z})}
\end{equation}

\end{lemma}
{\it Proof.} $a.$ We express the KL divergence as follows
\[
R(\tilde{q}\|q)=E_{\tilde{q}}\left[\log\frac{\tilde{q}}{q}\right]=E_{q}\left[\frac{\tilde{q}}{q}\log\frac{\tilde{q}}{q}\right]
\]
Then, we use Theorem~\ref{thm:fracdistrib} and we obtain \eqref{eq:KLnew}. For $b.$, we additionally recall \eqref{qoitypeI}.

\begin{remark}
As mentioned in Theorem~\ref{thm:fracdistrib}, the result holds for MRFs denoted by $(\mathbf{Y},p)$ and $(\mathbf{Y},\tilde{p})$, dropping the context $\mathbf{u}$ from $\Phi_{\mathbf{u}}^{\mathrm{i}}$. 
\end{remark}

\section{Main Results}\label{FMR}

In this section, we present a information-based UQ  method on the predictions for QoIs for general MRFs/rMRFs by  quantifying the model uncertainty for MRFs/rMRFs arising from statistical learning of graph models or from physical modeling. Our starting point is the Donsker-Varadhan variational principle \cite{DE}, which in turn implies  the {\it Gibbs Variational principle} for the KL divergence (see \cite{CD,DKPP}):
\begin{equation}\label{probUQ2}
\sup_{\lambda>0}\left\{\frac{-\Lambda_{p}^{f}(-\lambda)-R(\tilde{q}\|q)]}{\lambda} \right\}\leq E_{\tilde{q}}[f] \leq \inf_{\lambda>0}\left\{\frac{\Lambda_{q}^{f}(\lambda)+R(\tilde{q}\|q)]}{\lambda} \right\}
\end{equation}
As mentioned earlier, we focus on KL divergence as it scales correctly with the dimension of the baseline model \cite{DKPP}. In the above inequality, $q$ is the baseline rMRF and $\tilde{q}$ is an alternative model in the ambiguity set defined in \eqref{eq:ambiguity:*}. We note that at a MRF point of view, \eqref{probUQ2} holds  as well. Moreover, $\Lambda_{q}^{f}(\lambda)$ is the cumulant generating function (CGF) computed with respect to $p$ given by
\begin{equation}\label{eq:mgf,RL}
\Lambda_{q}^{f}(\lambda):=\log E_{q}[e^{\lambda f}] 
\end{equation}
while  $f$ is a QoI.  The class of QoI that we examine here as discussed in the next subsection. 

We take advantage of the total $\tilde{q}$-excess factors relative to $q$, likelihood ratio and an explicit formula for KL divergence on MRFs/rMRFs (see Lemma~\ref{lemma:RE}) in Section~\ref{subsec:MF} as well as in handling of the inherent high-dimensionality of such graphical models and we obtain tight and scalable, information-based bounds on the predictions for QoIs. Finally, we prove tightness of the UQ bounds, i.e. we prove that the bounds are attainable by MRFs/rMRFs, we compute their probability distribution  and we develop a strategy to determine their associated graph structures.

\subsection{Quantities of Interest}\label{qoigeneral}
We primarily consider two classes of QoIs $f(\mathbf{Z})$.  The first  has QoIs that are expressed as a characteristic function on events of interest such as \eqref{QoItoymodel}  in the medical diagnostics example presented in Section~\ref{sec:Applications}. The second class consists of QoIs that are sufficient statistics for the models $q$ and $\tilde{q}$ and  are also present in the  total $\tilde{q}$-excess factor of type I and II relative to $q$, i.e.  we consider $f(\mathbf{Z})$ that satisfies
\begin{equation}\label{qoitypeI}
 f(\mathbf{Z})=\frac{1}{C_{\mathrm{i}}}\left(\log\Phi_{\mathbf{u}}^{\mathrm{i}}(\mathbf{Z})+\kappa_{\mathrm{i}}(\mathbf{Z})\right),\;\;
\mathrm{i}=\mathrm{I},\mathrm{II}.
\end{equation}
for some non-zero constant $C_{\mathrm{i}}\equiv C_{\mathrm{i}}({\mathbf{w}, \tilde{\mathbf{w}},\mathbf{u}})<1$ and a function $\kappa_{\mathrm{i}}(\cdot)\equiv \kappa_{\mathrm{i}}(\cdot\mid\mathbf{w},\tilde{\mathbf{w}},\mathbf{u})$ that may depend on $\mathbf{w},\tilde{\mathbf{w}}, \mathbf{u}$,  see also \eqref{eq:RE}. Such a class covers observables involved in finite size effects and phase diagrams for statistical mechanics models examined later (e.g. averages of spins given by \eqref{AverExample}). The CGF given by \eqref{eq:mgf,RL} is computable for QoIs in both classes.

\subsection{UQ bounds}\label{subsec:mrUQ} The next theorem is an UQ  result on rMRFs that is obtained by consolidating the total $\tilde{q}$-excess factors relative to $q$, likelihood ratio, KL divergence and QoIs.   Part (a) provides the UQ bounds for a general QoI and hence we use such bounds for QoIs examined in the medical diagnostics application in Section~\ref{sec:Applications}. Part (b) is particularly applicable for QoIs satisfy \eqref{qoitypeI}, so they are exploited by the statistical mechanics section.  
\begin{theorem}\label{Mainthm1}
Let $(\mathbf{Y}, p)$, $ (\mathbf{Y}, \tilde{p})$ be two MRFs defined over graphs $\mathcal{G}=( \mathcal{V},  \mathcal{E})$ and $\tilde{\mathcal{G}}=( \mathcal{V},  \tilde{\mathcal{E}})$ respectively. Let $\mathbf{u}$ be a context and $\mathcal{M}\subset\mathcal{V}$.  We consider the corresponding rMRFs $(\mathbf{Z}, q), (\mathbf{Z}, \tilde{q})$.  If $\tilde{q}$ is of Type $\mathrm{i}$, with $\mathrm{i}=\mathrm{I}$ or $\mathrm{II}$, then 

\noindent (a) for any QoI $f(\mathbf{Z})$, the following bounds hold
\begin{eqnarray}\label{UQmainine*}
 \pm E_{\tilde{q}}[f]\leq\inf_{\lambda>0} \frac{1}{\lambda}\Big\{ \log E_{q}[e^{\pm\lambda f}]+\frac{1}{E_{q}[\Phi_{\mathbf{u}}^{\mathrm{i}}]}E_{q}\left[\Phi_{\mathbf{u}}^{\mathrm{i}}\log\Phi_{\mathbf{u}}^{\mathrm{i}}\right]-\log E_{q}[\Phi_{\mathbf{u}}^{\mathrm{i}}]\Big\}
\end{eqnarray}
(b) for any QoI $f(\mathbf{Z})$ that satisfies \eqref{qoitypeI}, the following bounds hold:
\begin{eqnarray}\label{UQmainine}
 \pm E_{\tilde{q}}[f]\leq\frac{1}{1-C_{\mathrm{i}}}\inf_{\lambda>0} \frac{1}{\lambda}\Big\{ \log E_{q}[e^{\pm\lambda f}]-\log E_{q}\left[\Phi_{\mathbf{u}}^{\mathrm{i}}\right]+\frac{E_{q}\left[\kappa_{i}\Phi_{\mathbf{u}}^{\mathrm{i}}\right]}{E_{q}\left[\Phi_{\mathbf{u}}^{\mathrm{i}}\right]}\Big\}
 \end{eqnarray}
where $\Phi_{\mathbf{u}}^{\mathrm{i}}$ is the total $\tilde{q}$-excess factor relative to $q$ given by \eqref{eq:psi} and \eqref{eq:phi}, $\kappa_{\mathrm{i}}$ and $C_{\mathrm{i}}$ are defined in \eqref{qoitypeI}. Note that when $\tilde{q}$ is of Type $\mathrm{I}$, $\tilde{Z}_{\mathbf{u}}(\tilde{\mathbf{w}})=Z_{\mathbf{u}}(\tilde{\mathbf{w}})$.
\end{theorem}
The proof given in Appendix~\ref{app:Proofmain} is based on Lemma~\ref{lemma:RE} and the characterization of the exponential integrals. An application to a single parameter exponential family is given in Appendix~\ref{app:Proofmain}.

\subsection{Tightness of UQ bounds for MRFs/rMRFs} Here, we prove that the inequalities \eqref{UQmainine*} and \eqref{UQmainine} are tight i.e. they become an equality for a suitable model $\tilde{q}\in \mathcal{Q}^{\eta}$ given by \eqref{eq:ambiguity:*} standing for the worst case scenarios. The practical interpretation of the tightness of UQ bounds is that these distributions are reasonable as they belong to the ambiguity set in \eqref{eq:ambiguity:*}.

\begin{theorem}\label{thm:Tightness}Let $(\mathbf{Z},q)$ be a rMRF defined in subsection~\ref{MRFformulation} and $f(\mathbf{Z})$ be a QoI with finite MGF $E_{q}[e^{\lambda f(\mathbf{Z})}]$ in a neighborhood of the origin. Then there exist $0<\eta_{\pm}\leq\infty$ such that for any $\eta\leq\eta_{\pm}$
there exist probability measures $q^{\pm}=q^{\pm}(\eta)\in\mathcal{Q}_{\eta}$, where $\mathcal{Q}_{\eta}$ is given in \eqref{eq:ambiguity:*}, such that \eqref{UQmainine*} and \eqref{UQmainine} become an equality. Furthermore,  $q^{\pm}=q^{\lambda_{\pm}}$ with
\begin{equation}\label{eq:plusminusmeas}
dq^{\lambda_{\pm}}=\frac{e^{\lambda_{\pm}f}}{ E_{q}[e^{\lambda_{\pm} f}]} dq
\end{equation}
and $\lambda_{\pm}$ being the unique solutions of $R(q^{\lambda_{\pm}}\|q)=\eta$. In particular, the total $q^{\pm}$-excess factor relative to $q$ denoted by $\Phi_{\mathbf{u}}^{\pm}$, satisfies
\[
\Phi_{\mathbf{u}}^{\pm}=e^{\lambda_{\pm}f}\;\;\mbox{and}\;\; C_{\mathrm{i}}=\lambda_{\pm},\; \kappa_{\mathrm{i}}=0 \;\mbox{respectively}.
\]
\end{theorem}
\begin{proof}See Appendix~\ref{proofofthm:Tightness}.\end{proof}
\noindent The result holds also for MRFs.  The corresponding quantities involved in the theorem are denoted by $p,p^{\lambda_{\pm}}$ and $\Phi^{\pm}$.

\begin{remark} For convenience  we use its MRF version. Given a baseline MRF $(\mathbf{Y},p)$, its associated graph $\mathcal{G}$ and  a QoI $f$, Theorem~\ref{thm:Tightness}  guarantees the existence of probability distributions $p^{\lambda_{\pm}}$ such that  \eqref{UQmainine*} and \eqref{UQmainine} become an equality (this is not an unlikely extreme case) and also specifies the distributions explicitly.  However, it does not imply how different the associated graphs of $p^{\pm}$ are, compared to the graph associated to $p$ or grossly speaking, if they are Type I or II. Depending on $f$, there are cases where this can be determined. In fact, by recalling the Hammersley-Clifford Theorem, we express
\begin{equation}\label{eq:plusminusmeas1}
dp^{\lambda_{\pm}}=\frac{e^{\lambda_{\pm}f}}{ E_{p}[e^{\lambda_{\pm} f}]} \frac{1}{Z(\mathbf{w})}\prod_{c\in\mathcal{C}_{\mathcal{G}}}\Psi_{c}(\mathbf{y}_{c}\mid\mathbf{w}_c)=\frac{1}{ Z^{\pm}(\lambda^{\pm},\mathbf{w})} \prod_{c\in\mathcal{C}_{\mathcal{G}}}e^{\lambda_{\pm}f}\Psi_{c}(\mathbf{y}_{c}\mid\mathbf{w}_c)
\end{equation}
where  $Z^{\pm}(\lambda^{\pm},\mathbf{w})= E_{p}[e^{\lambda_{\pm} f}]Z(\mathbf{w})$ is the partition function of $p^{\lambda_{\pm}}$.

We turn our attention to the product in \eqref{eq:plusminusmeas1}. Each factor is defined on a maximal clique of $\mathcal{G}$ apart from $e^{\lambda_{\pm} f}$. We focus on $f$; Suppose that $f$ is a QoI with domain $\mathrm{Dom}(f)$ and cannot be written as a sum of more than two functions e.g. sample average. If there is a maximal clique $c_0$ such that $\mathrm{Dom}(f)\subseteq c_0$, then it turns out that all clique potentials of $p^{\lambda_{\pm}}$ and $p$ are equal except $\tilde{\Psi}_{c_0}=e^{\lambda_{\pm} f}\Psi_{c_{0}}$, and hence 
\begin{equation}\label{eq:plusminusmeas2}
dp^{\lambda_{\pm}}=\frac{1}{ E_{p}[e^{\lambda_{\pm} f}]Z(\mathbf{w})} \underbrace{e^{\lambda_{\pm} f}\Psi_{c_{0}}}_\text{$\tilde{\Psi}_{c_0}$} \prod_{c\neq c_0}\Psi_{c}(\mathbf{y}_{c}\mid\mathbf{w}_c)
\end{equation}
The associated graphs of $p^{\lambda_{\pm}}$ are apparently of Type I as no change on maximal cliques occurs. If  $\mathrm{Dom}(f)\cap c\neq\emptyset$ for more than two maximal cliques $c$, then the associated graphs to $p^{\lambda_{\pm}}$ have been changed and thus are Type II.  An example is discussed in subsection~\ref{subsec:tightnessMedDiagn}. On the other hand, if $f$ can be expressed as a sum of some functions $f=\sum_i f_i$, then we may have more than one candidate graphs associated to $p^{\lambda_{\pm}}$ either Type I or II. In fact,  the exponential can be factorized further (e.g. $e^{\lambda_{\pm} f}=\prod_{i}e^{\lambda_{\pm} f_i}$), giving rise to more than one ways of matching the clique potentials in the sense of \eqref{eq:plusminusmeas2}. 
\end{remark}

\begin{remark}
 The parameter $\eta$ in Theorem~\ref{thm:Tightness} is also called {\it misspecification parameter}, and can be thought of as a {\it  non-parametric ``stress test''} for the rMRF, and can be tuned by hand so one can explore how the level of uncertainty affects QoIs. Alternatively, $\eta$ can be computed as the KL divergence from the available data (e.g data used to construct the baseline model in Medical Diagnostics, Section~\ref{sec:Applications}) in the form of a histogram or a KDE and thus subs for the distance of the baseline model from the unknown true model, \cite{FLKV}.
\end{remark}

\section{UQ for Medical Diagnostics}\label{sec:Applications} 
% \section{A motivating Example}\label{sec:fundamentalexamples}
Let us  introduce a simple example from medical diagnostics. We exploit its simplicity and low dimensionality to demonstrate  MRF  modeling with parameters and structure learned from data as well as the types of uncertainties that arise naturally in MRF modeling.
\smallskip

\noindent {\bf Setup.} Consider the problem of investigating  interdependence  (structure) and its strength (parameters) between Smoking (S), Asthma (A), Lung cancer (L), and Cough (C), \cite{D1}. It is assumed there are   prior expert  knowledge  and data  encoded  by a probabilistic model (distribution) $p^*$ defined on $\{S,C,L,A\}$. Due to limitations in expert knowledge and data, the true distribution $p^*$ itself may be altogether unknown. This, in turn, forces us to build a surrogate baseline model $p$, which therefore is uncertain in ways we will specify next.
\smallskip

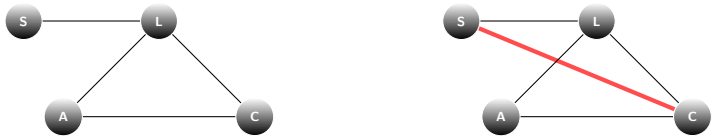
\begin{figure}[H]  
\centering
\begin{tikzpicture}[node distance=1.8cm,
                   main node/.style={top color =white , bottom color = black, shape=circle, draw,font=\sffamily\tiny\bfseries}]
       
  \node[main node, white] (1) {\textbf{S}{\tiny{}}};
  \node[main node, white] (4) [right of=1] {\textbf{L}{\tiny{}}};
  \node[main node, white] (3) [below right of=4] {\textbf{C}{\tiny{}}};
  \node[main node, white] (5) [below left of=4] {\textbf{A}{\tiny{}}};

  \path[every node/.style={font=\sffamily\small}]
    (4) edge node [] {} (3)
    (1) edge node [] {} (4)
    (5) edge node [] {} (4)
    (5) edge node [] {} (3);
\end{tikzpicture}
\hspace{2cm}
    \begin{tikzpicture}[node distance=1.8cm,
                   main node/.style={top color =white , bottom color = black, shape=circle, draw,font=\sffamily\tiny\bfseries}]
       
  \node[main node, white] (1) {\textbf{S}{\tiny{}}};

  \node[main node, white] (3) [below right of=4] {\textbf{C}{\tiny{}}};
  \node[main node, white] (4) [ right of=1] {\textbf{L}{\tiny{}}};
  \node[main node, white] (5) [below left of=4] {\textbf{A}{\tiny{}}};

  \path[every node/.style={font=\sffamily\small}]
    (1) edge [red!70!white, ultra thick] node [] {} (3)
    (4) edge node [] {} (3)
    (1) edge node [] {} (4)
    
    (5) edge node [] {} (4)
    (5) edge node [] {} (3);
    \end{tikzpicture}
\caption{{\small (Left) MRF structure $(\mathbf{Y},p)=(\{S,C,L,A\},p)$ over $\mathcal{G}$ with joint probability distribution $p$. $S\in\{s_0,s_1\}$, $L\in\{l_0,l_1\}$, $A\in\{a_0,a_1\}$ and $C\in\{c_0,c_1\}$. For example,  the values $s_0$ and $s_1$ can be thought as smoking and non-smoking respectively,  and so forth. The random variables $\mathbf{Y}=\{Y_1,Y_2,Y_3,Y_4\}=\{S,L,A, C\}$ are accordingly attached to the nodes in $\mathcal{V}=\{1,\dots, 4\}$ with edges in $\mathcal{E}=\{1-2,2-3,2-4,3-4\}$. The class of maximal cliques is $\mathcal{C}_{\mathcal{G}}=\Big\{\{1,2\}, \{2,3,4\}\Big\}$. (Right) A Type II model $(\tilde{\mathcal{G}},\tilde{\mathbf{w}},\tilde{p})$ over $\mathbf{Y}=\{S,C,L,A\}$ with joint probability distribution $\tilde{p}$. The associated graph is demonstrated by $\tilde{\mathcal{G}}=(\mathcal{V},\tilde{\mathcal{E}})$ with $\tilde{\mathcal{E}}=\mathcal{E}\cup\{1-4\}$. The new edge is shown in red color.}}\label{fig:M}  
\end{figure}

\noindent {\bf Baseline MRF.} Let $\mathcal{D}=\{\mathbf{d}[1],\dots,\mathbf{d}[N]\}$ be a large collection of patient records sampled from $p^*$. Using a  structure-learning algorithm  on the data  $\mathcal{D}$ (for instance, greedy score-based structure search algorithm for log-linear models \cite{KF, GBC}), a model  with the structure of $\mathcal{G}$ illustrated in Figure~\ref{fig:M}, (Left) is built, \cite{D1}. We assume that the graph is undirected as the directionality associated with the variable dependencies is not known (or is not expected). Subsequently, by parameter learning  (for instance, using maximum likelihood estimation \cite{KF}) the  weights $\mathbf{w}$ become specified from the available data.  From now on the resulting model  $(\mathcal{G},\mathbf{w},p)$ is called the baseline model.
 As in \cite{D1}, the  joint probability distribution could be a log-linear model (\cite{KF}, Section 4.4) and thanks to Hammersley-Clifford Theorem, is factorized over the maximal cliques with clique potentials $\Psi_{c}(\mathbf{y}_{c}\mid\mathbf{w}_{c})=e^{w_{c} f_{c}(\mathbf{y}_{c})},\;\;\mathbf{w}=\{\mathbf{w}_c\}_{c\in \mathcal{C}_{\mathcal{G}}}$, where $f_c$  is often called {\it feature}. 
\smallskip

\noindent{\bf Alternative models.} Both  learning steps   can induce uncertainties in structure and/or  parameters on the baseline. Next, we  model and  quantify  such uncertainties   by considering alternative models to the baseline of Type I and II:
we focus on graphical models that may have been obtained by learning structure and parameters from either a different data set $\tilde{\mathcal{D}}=\{\tilde{\mathbf{d}}[1],\dots,\tilde{\mathbf{d}}[\tilde{N}]\}$ or the same data set $\mathcal{D}$ but with different prior (expert)  knowledge. We denote the corresponding alternative  models  $(\tilde {\mathcal{G}},\tilde{\mathbf{w}},\tilde{p})$ and  assume they can be also represented by a MRF with $\tilde{p}>0$ in the class of log-linear models with clique potentials being given by $
\tilde{\Psi}_{c}(\mathbf{y}_{c})=e^{\tilde{w}_{c} \tilde{f}_{c}(\mathbf{y}_{c})}$
 be the clique potential. We consider the following QoIs defined as:
\begin{equation}\label{QoItoymodel}
g(\mathbf{Y})=\mathbf{1}_{A},\;\;\;{\mathrm{for\;any\;event\;of \; interest}}\;A\subset\Omega.
\end{equation}
For instance, $A=\!\{$patient is smoker with asthma$\}\!=\{\omega=(\omega_1,\omega_2, \omega_3, \omega_4):\omega_1=s_0, \omega_3=a_0\}$. 
\smallskip

\begin{figure}[H]
\vspace{-2.6cm}
\centering
  \includegraphics[width=0.6\linewidth]{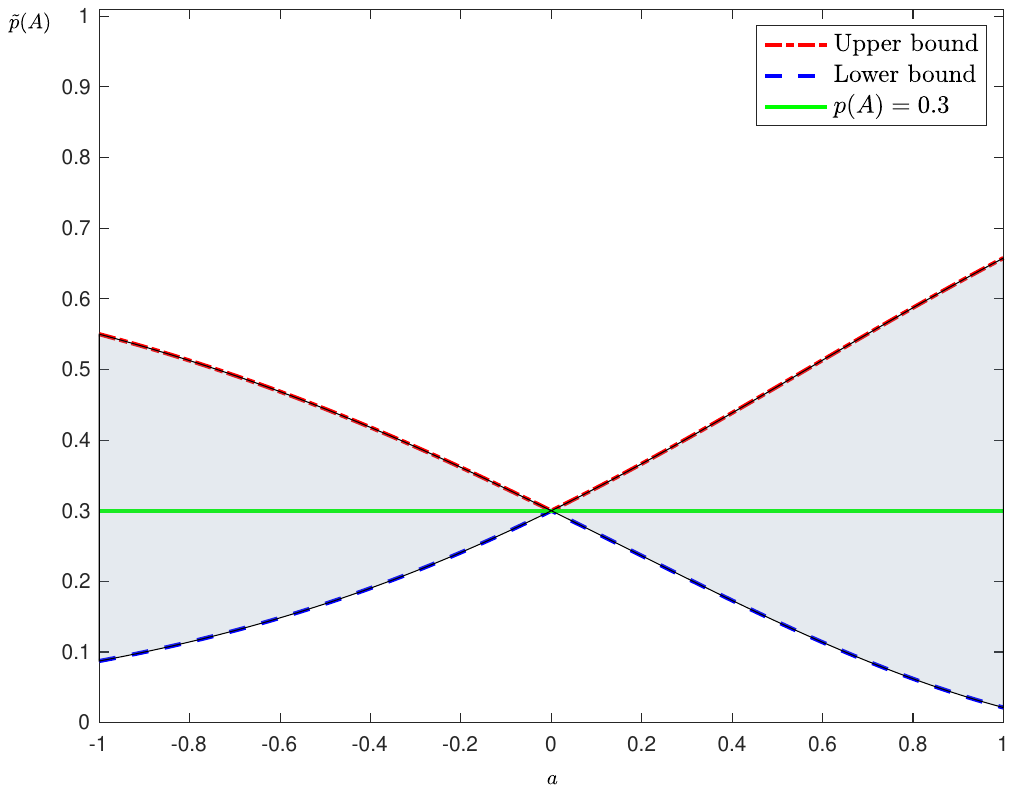}
  \vspace{-2.6cm}
  \caption{{\small For any event of interest, $A$ with $p(A)=0.3$, the red dashed-dot and the blue dashed curve are the upper bound  and lower bound for $\tilde{p}(A)$  provided in \eqref{ex_UQboundsI}, computed as functions of the weight change $a$.}}
  \label{fig:medDiagn1}
\end{figure}

\noindent {\bf Type I.} We consider the class of log-linear models $\tilde{p}$ over $
\mathcal{G}$ with weight change in one maximal clique after  learning weights from $\tilde{\mathcal{D}}$. Let $c$ be the maximal clique that a weight change occurred.  If $p_{\mathrm{I}}\equiv p(B_{c})$ and $a\in[-1,1]$ (depends on $\tilde{p}$), then for any  event of interest $A$, the following holds:
\begin{equation}\label{ex_UQboundsI}
\pm\tilde{p}(A)\leq\inf_{\lambda>0}\frac{1}{\lambda}\left\{\log\left(\frac{p(A)e^{\pm\lambda}+1-p(A)}{ e^{aw_c}p_{\mathrm{I}}+1-p_{\mathrm{I}}}\right)-\frac{aw_c e^{aw_c}p_{\mathrm{I}}}{e^{aw_c}p_{\mathrm{I}}+1-p_{\mathrm{I}}}\right\}
\end{equation}
where $a\in[-1,1]$ stands for the model uncertainty of alternative models of Type I and $w_c$ is the weight on $c$ of  $p$. The derivation of the UQ bounds in \eqref{ex_UQboundsI} is given in Appendix~\ref{App:MedicalDiagnostics}, while their demonstration as functions of the uncertainity parameter $a$ for any event of interest $A$ with $p(A)=0.3$ and when $p_{\mathrm{I}}=0.2$ is given in Figure~\ref{fig:medDiagn1}.
%\begin{figure}[tbhp]
%\hspace{-0.5cm}
%\subfloat[$\epsilon_{\max}=5$]{\label{fig:a}\includegraphics[width=0.5\linewidth]{MedDiag12}}
%\hspace{-0.5cm}
%\subfloat[$\epsilon_{\max}=0.5$]{\label{fig:b}\includegraphics[width=0.5\linewidth]{MedDiag22N}}
%\subfloat[$\epsilon_{\max}=0.5$]{\label{fig:b}\includegraphics[width=0.5\linewidth]{MedDiagn2NLAST}}
%\caption{Example figure using external image files.}
%\label{fig:testfig}
%\end{figure}
%

\begin{figure}[H]
\vspace{-2.3cm}
\centering
\hspace{-0.5cm} \includegraphics[width=0.5\linewidth]{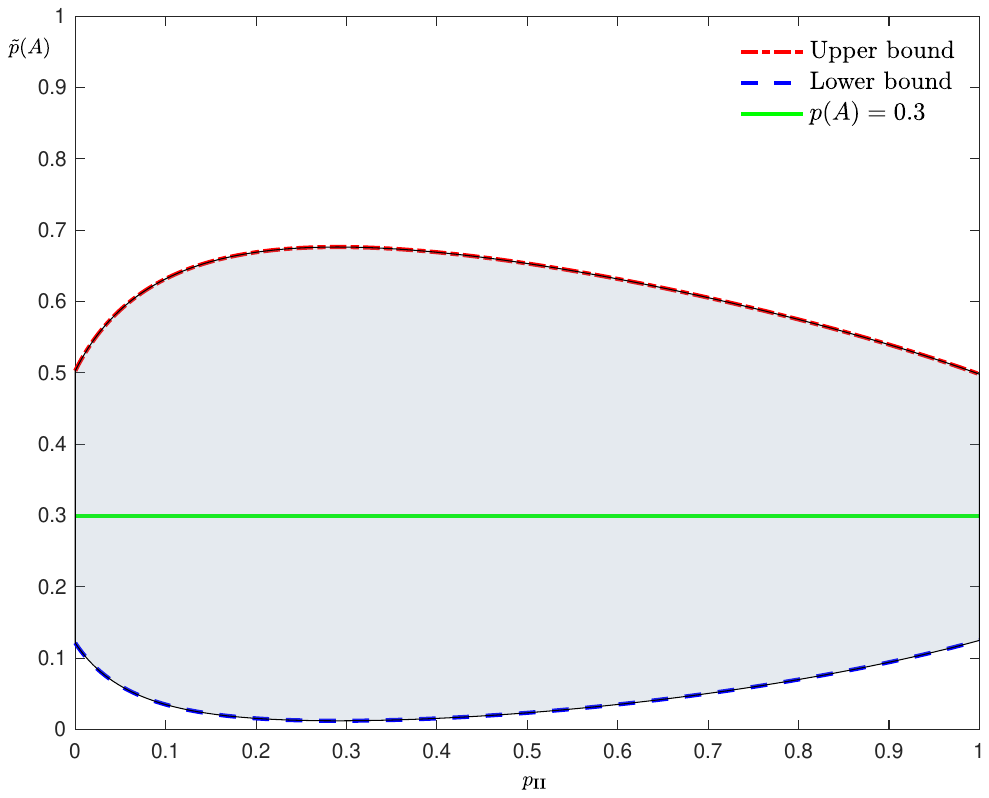}
 \hspace{0.1cm}
  \includegraphics[width=0.5\linewidth]{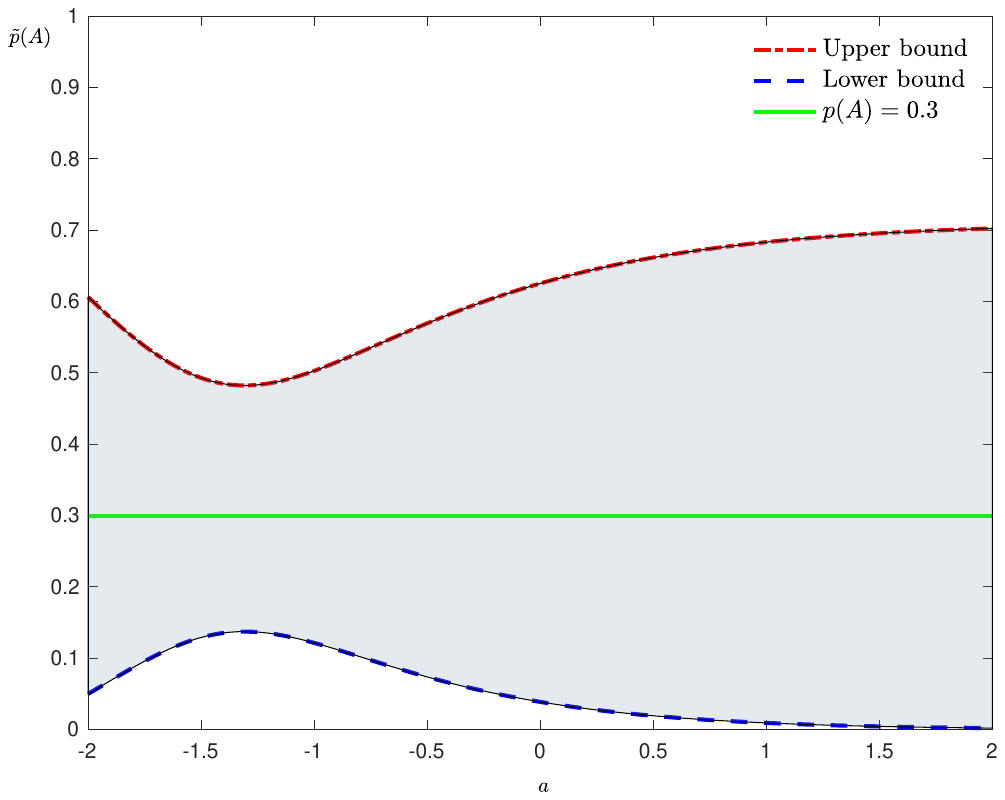}
  \vspace{-2.6cm}
\caption{{\small    $A$ is an event of interest with $p(A)=0.3$. (Left) For $p_{\mathrm{I}}=0.2$, $w_c=1.5$ and  $a=-0.2$,  the red dash-dot and the blue dashed curves are the upper bound  and lower bound  for $\tilde{p}(A)$  provided in \eqref{EBd}, computed as  functions of $p_{\mathrm{II}}$. (Right) For $p_{\mathrm{I}}=0.2$, $w_c=1.5$, $p_{\mathrm{II}}=0.7$, the red curve and the blue are the upper bound  and   lower bound for $\tilde{p}(A)$, computed as  functions of the weight change $a\in[-2,2]$.}}
  \label{fig:medDiagn2}
\end{figure}

\noindent {\bf Type II.} We consider the class of log-linear models $\tilde{p}$ over $
\tilde{\mathcal{G}}$ with $\tilde{\mathcal{V}}=\mathcal{V}$, $\tilde{\mathcal{E}}=\mathcal{E}\cup e$, where the new edge $e$ (e.g see Figure~\ref{fig:M}, (Right)) enlarges an already existing maximal clique $\tilde{c}$ in the sense of the analysis in subsection~\ref{subsec:altm} after structure-learning from $\tilde{\mathcal{D}}$. The model uncertainties lie in the binary function $\tilde{f}_{\tilde{c}}$ defined on $\tilde{c}$ and the new weight $\tilde{\mathbf{w}}_{\tilde{c}}$. The binary function $f_{\tilde{c}}$ induces a set $B_{\tilde{c}}=\{(\omega_1,\omega_2, \omega_3, \omega_4):\tilde{f}_{\tilde{c}}(\omega_{\tilde{c}})=1\}$. The set $B_{\tilde{c}}$ satisfies one of the following: $B_{\tilde{c}}\cap B_c=\emptyset$ or $B_{\tilde{c}}\cap B_c\neq\emptyset$. For $B_{\tilde{c}}\cap B_c=\emptyset$, if $p_{\mathrm{I}}\equiv  p(B_{c})$, $p_{\mathrm{II}}\equiv p(B_{\tilde{c}})$ and $a\in\mathbb{R}$, then for any  event of interest $A$, the following holds:
\begin{eqnarray}\label{EBd}
\pm\tilde{p}(A)&\leq&\inf_{\lambda>0}\frac{1}{\lambda}\Bigg\{\log\left(\frac{p(A)e^{\pm\lambda}+1-p(A)}{1-(1-e^{(1+a)w_c})p_{\mathrm{II}}-(1-e^{-w_{c}})p_{\mathrm{I}}}\right)\nonumber\\
&&\qquad-\frac{w_{c}e^{-w_c}p_{\mathrm{I}}-(1+a)w_ce^{(1+a)w_c}p_{\mathrm{II}}}{1-(1-e^{(1+a)w_c})p_{\mathrm{II}}-(1-e^{-w_{c}})p_{\mathrm{I}}}\Bigg\}
\end{eqnarray}
 The derivation of the UQ bounds in \eqref{EBd} is given in Appendix~\ref{App:MedicalDiagnostics} while their demonstration  for any event $A$ with $p(A)=0.3$ as functions of the uncertainty parameters $a$ (when $p_{\mathrm{I}}=0.2$, $w_c=1.5$ and $p_{\mathrm{II}}$=0.7) and $p_{\mathrm{II}}$ (when $p_{\mathrm{I}}=0.2$, $w_c=1.5$, $a=-0.2$) is given in Figure~\ref{fig:medDiagn2}. Note that the case where $B_{\tilde{c}}\cap B_c\neq\emptyset$ is more complicated.  However, the KL divergence is still explicitly computable (see Remark~\ref{impremark}). 
\subsection{Tightness}\label{subsec:tightnessMedDiagn}
Let $g$ be the QoI given by \eqref{QoItoymodel}. By applying Theorem~\ref{thm:Tightness}, there exist probability measures $p^{\pm}=p^{\pm}(\eta)\in\mathcal{Q}^{\eta}$, where $\mathcal{Q}^{\eta}$ is  given in \eqref{eq:ambiguity:*}, such that \eqref{UQmainine*} becomes an equality and  $p^{\pm}=p^{\lambda_{\pm}}$ are given by  $dp^{\lambda_{\pm}}=\frac{e^{\lambda_{\pm}\mathbf{1}_{A}}}{ p(A)e^{\lambda_{\pm}}+1-p(A)}\, dp$ and $\lambda_{\pm}$ being the unique solutions of $R(p^{\lambda_{\pm}}\|p)=\eta$. Depending on the event of interest $A$, we can determine the graph associated with $p^{\lambda_{\pm}}$.
Specifically, if $A=\cap_i A_i$ where all $A_i$ are defined on the same maximal clique of $\mathcal{G}$ given in Figure~\ref{fig:M}, then the graph associated with $p^{\lambda_{\pm}}$ is $\mathcal{G}$ and hence both models are Type I. If at least two $A_i, A_j$ are defined on different maximal cliques, the associated graphs are different than $\mathcal{G}$, e.g. let $A=\!\{$patient is smoker with asthma$\}\!=\{\omega=(\omega_1,\omega_2, \omega_3, \omega_4):\omega_1=s_0, \omega_3=a_0\}=\{\omega:\omega_1=s_0\}\cap\{\omega: \omega_3=a_0\}$. Since the total $p^{\pm}$-excess factor relative to $p$ $\Phi^{\pm}=e^{\lambda_{\pm}\mathbf{1}_{A}}$ cannot be further factorized, the new graph has the same set of nodes with an extra  edge $1-3$, that is $\tilde{\mathcal{E}}=\mathcal{E}\cup\{1-3\}$. In that case, both models are Type II.

\section{UQ for Statistical Mechanics} \label{sec:Ex2} Large-scale physical systems of interacting particles such as gases, liquids, and solids, are at the core of statistical mechanics and in particular of equilibrium statistical mechanics. The macroscopic properties of a system can be understood through its underlying microscopic description which fundamentally requires the microscopic states and an interaction between microscopic constituents. 
%Examples of well-known models with sufficient complexity are the one-dimensional and the square-lattice Ising model, the Curie-Weiss and the Mean Field model, \cite{Baxter}. The focus on statistical mechanics models from a machine learning perspective is driven by the study of Boltzmann Machines (BM) \cite{KF} and in particular, when one needs to compare BM with different parameters (Type I) and different nodes (Type III).
Statistical mechanics models such as the Ising model are fundamental in ML,  especially energy-based probabilistic models (generally defined as \eqref{eq:gibbs}) such as Boltzmann machines \cite{GBC}. Furthermore, methods from equilibrium statistical mechanics combined with information theory  can provide first insights into profound cornerstones of deep learning.  For example, although we use the KL divergence defined in Lemma~\ref{lemma:RE} for UQ, KL
between an energy-based model and  available data equals to the difference between Gibbs and Helmholtz free energy and is a natural ``distance" to use for statistical learning. Note that both UQ and statistical learning can be considered as dual concepts, \cite{BMP}.
 A more extensive analysis on these ideas, and generally on the intersection between statistical mechanics--also including non-equilibrium statistical mechanics--and deep learning  have been reviewed in \cite{BahriGanguli}.

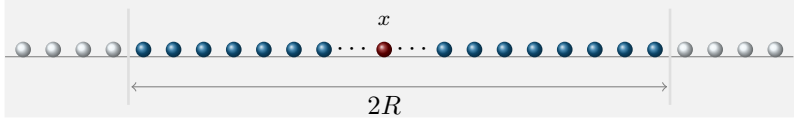
\begin{figure}[H]
\centering
\vspace{.01cm}
	\begin{tikzpicture}[scale=.8]
\draw[fill=lightgray!10!white, lightgray!20!white ] (-2.3,-1) rectangle (10.8,1);
\draw[-,color=gray](-2.3,0)--(10.8,0);
\draw[<->,color=gray](-0.2,-0.5)--(8.7,-0.5);
\draw[-,color=lightgray!50!white,line width=1pt](-.25,-.8)--(-.25,.8);
\draw[-,color=lightgray!50!white,line width=1pt ](8.75,-.8)--(8.75,.8);
\node at (4, 0.6) {\footnotesize $x$};

\shade[ball color=sky!10!white](-2,0.12)circle(1.3mm);
\shade[ball color=sky!10!white](-1.5,0.12)circle(1.3mm);
\shade[ball color=sky!10!white](-1,0.12)circle(1.3mm);
\shade[ball color=sky!10!white](-.5,0.12)circle(1.3mm);
\shade[ball color=sky](0,0.12) circle(1.3mm);
\shade[ball color=sky](.5,0.12)circle(1.3mm);
\shade[ball color=sky](1,0.12)circle(1.3mm);
\shade[ball color=sky](1.5,0.12)circle(1.3mm);
\shade[ball color=sky](2,0.12)circle(1.3mm);
\shade[ball color=sky](2.5,0.12)circle(1.3mm);
\shade[ball color=sky](3,0.12) circle(1.3mm);
\node at (3.5, 0.12) { $\cdots$};
\node at (4, -0.8) { $2R$};
\shade[ball color=red!50!black](4,0.12)circle(1.3mm);
\node at (4.5, 0.12) { $\cdots$};
\shade[ball color=sky](5,0.12) circle(1.3mm);
\shade[ball color=sky](5.5,0.12)circle(1.3mm);
\shade[ball color=sky](6,0.12)circle(1.3mm);
\shade[ball color=sky](6.5,0.12)circle(1.3mm);
\shade[ball color=sky](7,0.12)circle(1.3mm);
\shade[ball color=sky](7.5,0.12)circle(1.3mm);
\shade[ball color=sky](8,0.12) circle(1.3mm);
\shade[ball color=sky](8.5,0.12)circle(1.3mm);
\shade[ball color=sky!10!white](9,0.12)circle(1.3mm);
\shade[ball color=sky!10!white](9.5,0.12)circle(1.3mm);
\shade[ball color=sky!10!white](10,0.12)circle(1.3mm);
\shade[ball color=sky!10!white](10.5,0.12)circle(1.3mm);

\end{tikzpicture}
\caption{{\small One-dimensional Ising spin lattice on $\Delta$ (light gray area with blue, red, and white particles). The spin located at $x\in\Delta$ (red particle) interacts only with spins located at $y$ in $B_x(R)$ (blue particles) with strength of interaction $J(x,y)$. The red spin does not interact with the white ones as they are located at distance greater than $R$ from $x$.}}
  \label{fig:Range}
\end{figure}

\subsection{\bf Ising Model}\label{subsubsec:ising} An illustrative example is the Ising model, where the space of all  microstates is the collection of all spin configurations on a bounded region $\Delta \subseteq\mathbb{Z}^d$:
\[\Omega:=\{\pm1\}^{\Delta}=\Big\{\sigma_{\Delta}=\{\sigma_{\Delta}(x)\}_{x\in\Delta}: \sigma_{\Delta}(x)\in\{+1,-1\}\Big\}\]
 as in Figure~\ref{fig:Range}, \cite{errico,KF}. An interaction between spins can be short, long range or a combination (such as Lennard-Jones potential, \cite{Presutti}), positive (ferromagnetism), etc, \cite{errico,FV,G}. Here we consider a $d$-dimensional Ising spin system on $\Delta$ with a generic interaction $\mathbf{J}=\{J(x,y):x,y\in\Delta\}$ 
satisfying three properties: for all $x,y\in\Delta$ and $z\in\mathbb{R}^d$
\begin{equation}\label{proper:transinva}
  J(x+z,y+z)=J(x,y) \;\; \qquad \mathrm{(translational\; invariance)}
\end{equation}
\begin{equation}\label{proper:symmetry}
  J(x,y)=J(y,x) \;\; \qquad \mathrm{(symmetry)}
\end{equation} 
\begin{equation}\label{proper:summ}
\sum_{x\neq 0}|J(0,x)|<\infty \qquad \mathrm{(summability)}
\end{equation}
and an external field, $h\in\mathbb{R}$.
Let $R>0$ be the length of the range of interaction. For $x\in\mathbb{Z}^d$, $B_{x}(R)=\{y\in\mathbb{Z}^d:\|x-y\|_{d}\leq R\}$ is the set of all spins that the spin located at $x$ interacts with and $\|x-y\|_d := \sqrt
{\sum_{i=1}^d |x_i-y_i|^2}$. For  convenience, we denote $B_{x,R}^{\neq}:=B_{x}(R)\setminus x$. 
\subsubsection{\bf Boundary conditions}\label{subsubsec:BC}  Boundary conditions are a fundamental concept in statistical mechanics, \cite{S}. For simplicity, let us assume that $\Delta$ is a hypercube. We consider a system where particles not only interact with particles in $\Delta$, but  also with particles ``outside" of $\Delta$. Let $\bar{\sigma}_{\Delta^{c}}$ be a given fixed configuration of spins on the complement of $\Delta$ denoted by $\Delta^{c}$, see Figure~\ref{fig:BC}. The Hamiltonian energy of the system is given by:
\begin{equation}
H^{\mathbf{J},h}(\sigma_{\Delta}|\bar{\sigma}_{\Delta^{c}})=H^{\mathbf{J},h}(\sigma_{\Delta})-\sum_{x\in\Delta}\sum_{y\in\Delta^c}
J(x,y)\sigma_{\Delta}(x)\sigma_{\Delta}(y)
\end{equation}
where \begin{equation}\label{Hamiltoniangen}
H^{\mathbf{J},h}(\sigma_{\Delta})=-\frac{1}{2}\sum_{x\in\Delta}\sum_{y\in\Delta}
J(x,y)\sigma_{\Delta}(x)\sigma_{\Delta}(y)-h\sum_{x\in\Delta}\sigma_{\Delta}(x)
\end{equation}
The Gibbs measure with boundary condition $\bar\sigma_{\Delta^{\rm{c}}}$ is defined as
\begin{equation}\label{eq:gibbs}
\mu_{\mathbf{J},\beta, h}^{\Delta}(\sigma_{\Delta}\mid \bar\sigma_{\Delta^{\rm{c}}})=\frac{1}{Z_{\bar\sigma_{\Delta^{\rm{c}}}}(\mathbf{J},\beta, h)}e^{-\beta H^{\mathbf{J},h}(\sigma_{\Delta}|\bar\sigma_{\Delta^{c}})}.
\end{equation}
where 
$Z_{\bar\sigma_{\Delta^{\rm{c}}}}(\mathbf{J},\beta, h)=\sum_{\sigma_{\Delta}}e^{-\beta H^{\mathbf{J},h}(\sigma_{\Delta}|\bar\sigma_{\Delta^{c}})}$ is the partition function.
\subsubsection{\bf rMRF formulation} \label{subsubsec:rmrf} A system with configuration as boundary conditions does not admit a MRF description. So, we describe the system using rMRFs. The set of nodes is $\mathbb{Z}^d$, the set of edges can be constructed by looking at all $(x,y)$ such that $\|x-y\|_{d}\leq R$ and the context is $\mathbf{u}=\bar\sigma_{\Delta^c}$, which corresponds to a fixed boundary condition. Then $(\sigma_{\Delta}, \mu_{\mathbf{J},\beta, h}^{\Delta}(\cdot\mid \bar\sigma_{\Delta^{\rm{c}}}))$ is a rMRF with maximal cliques $c_x=\{y\in\Delta:y\in B_{x,R}^{\neq}\}$ (spins in $c_x$ interact with all spins in $c_x$). Let $\mathbf{w}=\{\mathbf{w}_{c_x}\}_{x\in\Delta}$ with $\mathbf{w}_{c_x}=(\mathbf{J}_{c_x},\beta,h)$ and $\mathbf{J}_{c_x}=\{J(x,y):y\in c_x\}$. We express each clique potential as 
\begin{equation}\label{cliquepotentialexample}
\Psi_{c_x}=\exp\left\{\beta\sigma_{\Delta}(x)\left(h+\frac{1}{2}\sum_{\substack{y\in\Delta\\ y\in B_{x,R}^{\neq}}}J(x,y)\sigma_{\Delta}(y)+\sum_{\substack{y\in\Delta^c\\ y\in B_{x,R}^{\neq}}}
J(x,y)\bar\sigma_{\Delta^c}(y)\right)\right\}
\end{equation}
Note that we may resume the full notation when we needed, that is  $\Psi_{c_x}\equiv\Psi_{c_x}[\bar\sigma_{\Delta^c}](\sigma_{c_x}\mid\mathbf{w}_{c_{x}})$ where $\sigma_{c_x}$ is the Ising spin configuration defined on all $y\in c_{x}$.
\subsection{UQ Formulation}

\subsubsection{Alternative models}\label{subsubsec:Altmodel} We consider models on a lattice with perturbed interaction in the strength (Type I) and/or range (Type II) such as truncated  or long range interaction. Given  $\mathbf{J}$ as in subsection~\ref{subsubsec:ising}, an interaction $F(x,y)$ satisfying \eqref{proper:transinva}-\eqref{proper:summ} with length of range $R_F$, we say that  $\tilde{\mathbf{J}}^{\mathbf{F}}=\{\tilde{J}^F(x,y):x,y\in\mathbb{Z}^d\}$  is a perturbed interaction if 
\begin{equation}\label{form:genpertinter}
\tilde{J}^F(x,y)= J(x,y){\bf 1}_{\|x-y\|_d\leq R}+F(x,y){\bf 1}_{\|x-y\|_d\leq R_F}+F(x,y){\bf 1}_{\|x-y\|_d> R_F}
\end{equation}
We say that a perturbed interaction is Type I iff
\begin{equation}\label{TTTTT1}
 R=R_F \;\mathrm{and}\; \mathrm{supp}(F)=\{(x,y):\|x-y\|_d\leq R_F\}.
 \end{equation}
 We say that a perturbed interaction is Type II iff
\begin{equation}\label{TTTTT2}
 R=R_F \;\mathrm{and} \;\mathrm{supp}(F)=\{(x,y):\|x-y\|_d> R_F\}.
 \end{equation}
The rMRF formulation of the system with $\tilde{\mathbf{J}}^{\mathbf{F}}$ goes similarly as in subsection~\ref{subsubsec:rmrf}. Note that the graph representation simplifies a possible complexity of $J$, $F$ and $\tilde{J}^F$ as we connect nodes $x,y$ according to the range of $J$, $F$ and $\tilde{J}^F$ and assign the corresponding strengths $J(x,y)$, $F(x,y)$ and $\tilde{J}^F(x,y)$.

\subsubsection{Total $\tilde{q}_{\Delta}$-excess factor relative to $q_{\Delta}$}\label{lemma:explicitphi}
\begin{lemma}
Let $\tilde{\mathbf{J}}^{\mathbf{F}}$ be defined in subsection~\ref{subsubsec:Altmodel} with support given by \eqref{TTTTT1} or \eqref{TTTTT2}, and $
q_{\Delta}(\cdot):=\mu_{\mathbf{J},\beta, h}^{\Delta}(\cdot\mid \bar\sigma_{\Delta^{c}})
$,  
 $
 \tilde{q}_\Delta(\cdot):=\mu_{\tilde{\mathbf{J}}^{\mathbf{F}},\beta, \tilde h}^{\Delta}(\cdot\mid \bar\sigma_{\Delta^{c}})$
be the corresponding Gibbs measures defined in \eqref{eq:gibbs}. The total $q_{\Delta}$-excess factor for $i=\mathrm{I},\mathrm{II}$ is given by
\begin{eqnarray}\label{tEF}
\Phi_{\bar\sigma_{\Delta^c}}^{\mathrm{i}}(\sigma_{\Delta})&=&\exp\Big\{\beta\sum_{x\in\Delta}\sigma_{\Delta}(x)\Big((\tilde{h}-h)+\frac{1}{2}\sum_{y\in A^{\mathrm{i}}_x\cap\Delta}F(x,y)\sigma_{\Delta}(y)\nonumber\\
&&\qquad+\sum_{y\in A^{\mathrm{i}}_x\cap\Delta^c}F(x,y)\bar{\sigma}_{\Delta^{\rm c}}(y)\Big)\Big\}
\end{eqnarray}
where for each $x\in\Delta$, $A^{\mathrm{I}}_x=B_{x}(R)$ and $A^{\mathrm{II}}_x=B_{x}(R)^c$, with $B_{x}(R)^c$ being the complement of $B_{x}(R)$. 
\end{lemma}
The proof is straightforward (see Appendix~\ref{prooflemma:explicitphi}).  Both $(\tilde{h}-h)$ and $F(x,y)$ in the total $\tilde{q}_{\Delta}$-excess factor relative to $q_{\Delta}$  point out  how different the external fields and interactions are respectively, as the latter satisfies $F(x,y)=\tilde{J}^F(x,y)-J(x,y)$.

\subsubsection{Quantities of Interest}\label{subsec:qoi}  The use of phase diagrams is  central in physics and material science. A phase diagram is defined as a graphical representation of equilibrium states under different thermodynamic parameters such as external field $h$, temperature $T$ and pressure $P$.  It is typically computed in the thermodynamic limit  (i.e a limiting process with $\Delta\nearrow\mathbb{Z}^d$ such that the ratio between inter-atomic distances
and macroscopic lengths vanishes), \cite{errico}. Equilibrium states are characterized by order parameters such as magnetization. For that, we consider the following observable
\begin{equation}\label{AverExample}
m(\sigma_{{\Delta}}):=\frac{1}{|{\Delta}|}\sum_{x\in {\Delta}}\sigma_{{\Delta}}(x)
\end{equation}
where $|\Delta|$ stands for the volume of a hypercube $\Delta\subset\mathbb{Z}^d$. As $\Delta$ invades the whole $\mathbb{Z}^d$, the expectation of $m(\sigma_{{\Delta}})$ yields the magnetization. Other QoIs could also be considered e.g. correlation functions $v(\sigma_{{\Delta}})=\frac{1}{|\Delta|^2}\sum_{x\in\Delta}\sum_{y\in\Delta}\sigma_{\Delta}(x)\sigma_{\Delta}(y)$.
 
%{\color{red}[UQ for finite size effects moved after KL]}

\subsubsection{\bf Cumulant Generating Function} \label{CGF} Let  $\Delta$ be a hypercube in $\mathbb{Z}^d$. Given a  configuration $\bar\sigma_{\Delta^c}$, the baseline model is an Ising model with interaction $\mathbf{J}$ defined in subsection~\ref{subsubsec:ising}. We compute the cumulant generating function  defined by \eqref{eq:mgf,RL} w.r.t the baseline model $q_{\Delta}$ (the computation is given in \eqref{specificmgf}):
\begin{equation}\label{scalMGF}
\Lambda_{q_{\Delta};|\Delta| m(\sigma_{\Delta})}(\pm\lambda)=\beta|\Delta|\left(P_{h\pm\frac{\lambda}{\beta},\beta,\mathbf{J}}^{\,\Delta}(\bar\sigma_{\Delta^c})-  P_{h,\beta,\mathbf{J}}^{\,\Delta}(\bar\sigma_{\Delta^c})\right)\end{equation}
where $P_{h,\beta,\mathbf{J}}^{\,\Delta}$ stands for the thermodynamic pressure, \cite{errico}, defined as \[
P_{h,\beta,\mathbf{J}}^{\,\Delta}(\bar\sigma_{\Delta^c}):=\frac{Z(\mathbf{J},\beta, h,\bar\sigma_{\Delta^c})}{\beta |\Delta|}.
\]

\subsubsection{\bf KL Divergence}\label{subsec:REwnorm} Here we utilize Lemma~\ref{lemma:RE} and specify the KL divergence in terms of $\kappa_{\mathrm{i}}$ and $\Phi_{\mathbf{u}}$ as involved in \eqref{eq:RE}  when the alternative models are Ising models with a perturbed interaction $\tilde{\mathbf{J}}^{\mathbf{F}}$ defined in subsection~\ref{subsubsec:Altmodel}. Then we bound it by using Lemma~\ref{lemma:important}. Before that, we use a well-established tool in statistical mechanics referred to as norm-$\|\cdot\|_{1}$,  \cite{S} to alternatively bound the KL divergence.  After all, we conclude that  our UQ approach gives a narrower area (i.e the area between the upper and lower UQ bound) provided by Theorem~\ref{Mainthm1} and thus smaller uncertainty, see Figure~\ref{fig:STATMECH1}. 

 {\it Norm-$\|\cdot\|_{1}$}: Let $\Phi^{h,\beta,\mathbf{J}}_{\Delta,\bar\sigma_{\Delta^{c}}}(\sigma_X)$ be the following quantity:
\begin{equation}\label{Phi_X}
\Phi^{h,\beta,\mathbf{J}}_{\Delta,\bar\sigma_{\Delta^{c}}}(\sigma_{X})=
\left\{
	\begin{array}{ll}
		-\frac{1}{2}\beta J(x,y)\sigma_{\Delta}(x)\sigma_{\Delta}(y)  & ,X=\{x,y\},  \;x\neq y,  \\
		-\beta \sigma_{\Delta}(x)\big(h+\sum_{y\in B_{x,R}^{\neq}\cap\Delta^c}J(x,y)\bar\sigma_{\Delta^{c}}(y)\big) & , X=\{x\}\\
		0 & ,\mbox{otherwise}
	\end{array}
\right.
\end{equation}
and similarly we define $\Phi^{\tilde h,\beta,\tilde{\mathbf{J}}^{\mathbf{F}}}_{\Delta,\bar\sigma_{\Delta^{c}}}(\sigma_{X})$. Then,
\begin{equation}
\beta H^{\mathbf{J},h}(\sigma_{\Delta}|\bar{\sigma}_{\Delta^{c}})=\sum_{X:X\cap\Delta\neq\emptyset} \Phi^{h,\beta,\mathbf{J}}_{\Delta,\bar\sigma_{\Delta^{c}}}(\sigma_{X})
\end{equation}
Also, $\beta H^{\tilde{\mathbf{J}}^{\mathbf{F}},\tilde h}(\sigma_{\Delta}|\bar{\sigma}_{\Delta^{c}})$ is defined similarly. Then the norm-$||\cdot||_1$ of $\Phi^{h,\beta,\mathbf{J}}_{\Delta,\bar\sigma_{\Delta^{c}}} -\Phi^{\tilde h,\beta,\tilde{\mathbf{J}}^{\mathbf{F}}}_{\Delta,\bar\sigma_{\Delta^{c}}} $ is defined as
\begin{equation}\label{norm1}
\| \Phi^{h,\beta,\mathbf{J}}_{\Delta,\bar\sigma_{\Delta^{c}}} -\Phi^{\tilde h,\beta,\tilde{\mathbf{J}}^{\mathbf{F}}}_{\Delta,\bar\sigma_{\Delta^{c}}} \|_1=\sum_{0\ni X}\|\Phi^{h,\beta,\mathbf{J}}_{\Delta,\bar\sigma_{\Delta^{c}}} -\Phi^{\tilde h,\beta,\tilde{\mathbf{J}}^{\mathbf{F}}}_{\Delta,\bar\sigma_{\Delta^{c}}} \|_{\infty}
\end{equation}
where $\|\Phi^{h,\beta,\mathbf{J}}_{\Delta,\bar\sigma_{\Delta^{c}}} -\Phi^{\tilde h,\beta,\tilde{\mathbf{J}}^{\mathbf{F}}}_{\Delta,\bar\sigma_{\Delta^{c}}} \|_{\infty}=\sup_{\sigma_X}|\Phi^{h,\beta,\mathbf{J}}_{\Delta,\bar\sigma_{\Delta^{c}}}(\sigma_{X}) -\Phi^{\tilde h,\beta,\tilde{\mathbf{J}}^{\mathbf{F}}}_{\Delta,\bar\sigma_{\Delta^{c}}} (\sigma_{X})|$ for $X\subset\mathbb{Z}^d$. 
\begin{lemma}\label{lemma:fornorm}
Let $F$ be an interaction satisfying \eqref{proper:transinva}-\eqref{proper:summ} with support given by \eqref{TTTTT1} or \eqref{TTTTT2}, then 
\[
R(\tilde{q}_{\Delta}\|q_{\Delta})\leq 2|\Delta|\| \Phi^{h,\beta,\mathbf{J}}_{\Delta,\bar\sigma_{\Delta^{c}}} -\Phi^{\tilde h,\beta,\tilde{\mathbf{J}}^{\mathbf{F}}}_{\Delta,\bar\sigma_{\Delta^{c}}} \|_1\leq 2\beta|\Delta|\left(|\tilde h-h|+\sum_{x\neq0}|F(0,x)|\right).
\]
\end{lemma}
\begin{proof}See Appendix~\ref{App:prooffornorm}.
\end{proof}
 Let us turn to our approach developed in Section~\ref{FMR}. We recall the total $\tilde{q}_{\Delta}$-excess factor relative to $q_{\Delta}$ from subsection~\ref{lemma:explicitphi} as well as the quantities from Section~\ref{qoigeneral}, and we express $
\log\Phi_{\bar\sigma_{\Delta^c}}^{\mathrm{i}}(\sigma_{\Delta})=C_{\mathrm{i}}|\Delta| m(\sigma_{\Delta})+\kappa_{\mathrm{i}}(\sigma_{\Delta})$ with
 \begin{equation}\label{eq:c1}
C_{\mathrm{I}}=\beta (\tilde{h}-h)<1, \;\;\;\kappa_{\mathrm{I}}(\sigma_{\Delta})=\beta\sum_{x\in\Delta}\sigma_{\Delta}(x)\Bigg(\frac{1}{2}\sum_{y\in A^{\mathrm{i}}_x\cap\Delta}F(x,y)\sigma_{\Delta}(y)\Bigg)+\beta F(\Delta|\bar\sigma_{\Delta^c})
\end{equation}
where $F(\Delta|\bar\sigma_{\Delta^c})=\sum_{x\in\Delta}\sum_{y\in A^{\mathrm{i}}_x\cap\Delta^c}F(x,y)\bar\sigma_{\Delta^c}(y)$.  We bound $\kappa_{\mathrm{I}}(\sigma_{\Delta})$as 
\begin{equation}\label{eq:c2}
0\leq \kappa_{\mathrm{I}}(\sigma_{\Delta})\leq\beta |\Delta| \left( \frac{1}{2}+2R  \frac{|\partial\Delta|}{|\Delta|}\right)\sum_{x\neq 0}|F(x,y)|
\end{equation}
 where we use the next lemma.
\begin{lemma}\label{lemma:important}
Let $L$ and $\partial\Delta$ be the side and the boundary of the hypercube $\Delta$ respectively with $L>>R_{F}$. Then, for any interaction $\mathbf{F}=\{F(x,y):x,y\in\mathbb{Z}^d\}$ satisfying \eqref{proper:transinva}-\eqref{proper:summ} and range $R_F$, the following holds: 
\begin{itemize}
\item[$(i)$] If the support of $F$ is given by \eqref{TTTTT1}, then \[
\sum_{x\in\Delta}\sum_{\substack{y\in\Delta^c\\ y\in B_{x,R_{F}}^{\neq}}}  F(x,y)\leq  R_F |\partial\Delta| \sum_{x\neq0}|F(0,x)|.
\]
\item[$(ii)$] If the support of $F$ is given by \eqref{TTTTT2}, then 
\[
\sum_{x\in\Delta}\sum_{y\in\Delta^c}F(x,y)\leq R_F |\Delta|\sum_{x\neq0}|F(0,x)|
\]

\end{itemize}
\end{lemma}
\begin{proof}The bounds are straightforward once we split the sum as follows:
\[
\sum_{x\in\Delta}\sum_{\substack{y\in\Delta^c\\ y\in B_{x,R_{F}}^{\neq}}}F(x,y)=\sum_{\substack{x\in\Delta\\dist(x,\Delta^c)\leq R_F}}\sum_{\substack{y\in\Delta^c\\ y\in B_{x,R_{F}}^{\neq}}}F(x,y)+\sum_{\substack{x\in\Delta\\dist(x,\Delta^c)> R_F}}\sum_{y\in\Delta^c}F(x,y)\leq R_F |\Delta|.
\]
where $dist(x,\Delta^c)=\inf\{\|x-y\|:y\in\Delta^c\}$. Note that when $L<<R_{F}$,  both (i) and (ii) are bounded by $R_F |\Delta|\sum_{x\neq0}|F(0,x)|$.
\end{proof}

\subsection{ UQ for finite-size effects and boundary conditions}\label{scalability}
Having computed all the ingredients needed for the analysis of subsections~\ref{subsec:mrKL}, \ref{qoigeneral} and \ref{subsec:mrUQ} under the above statistical mechanics formulation through rMRFs,  we capture the behavior of $m(\sigma_{\Delta})$ given in \eqref{AverExample} with respect to the perturbed model $\tilde{q}_\Delta$. The analysis from now on refers to models of Type I. Although Type II models can be worked on similarly, one example of Type II is discussed in Appendix~\ref{sec:long}. To get the UQ bounds for $E_{\tilde{q}_{\Delta}}[m(\sigma_{\Delta})]$, for $f(\mathbf{Z})=|\Delta| m(\sigma_{\Delta})$ we can either apply \eqref{probUQ2} using the crude bound in Lemma~\ref{lemma:fornorm}:
\begin{eqnarray}\label{bound:GenIs'}
\pm E_{\tilde{q}_{\Delta}}[m(\sigma_{\Delta})]\leq\inf_{\lambda>0}\Bigg\{ \frac{P_{h\pm\frac{\lambda}{\beta},\beta,\mathbf{J}}^{\,\Delta}-  P_{h,\beta,\mathbf{J}}^{\,\Delta}}{\lambda/\beta}+2\frac{\beta}{\lambda}(|\tilde h-h|+\mathcal{F})\Bigg\}
\end{eqnarray}
or  Theorem~\ref{Mainthm1} :
\begin{eqnarray}\label{bound:GenIs}
\;\;\;\;\;\;\;\;\;\;\pm E_{\tilde{q}_{\Delta}}[m(\sigma_{\Delta})]\leq\frac{1}{1- \beta (\tilde{h}-h)}\inf_{\lambda>0}\Bigg\{ \frac{P_{h\pm\frac{\lambda}{\beta},\beta,\mathbf{J}}^{\,\Delta}-  P_{h,\beta,\mathbf{J}}^{\,\Delta}}{\lambda/\beta}+\frac{\beta}{\lambda}\mathcal{F}\left(1+R_F \frac{|\partial\Delta|}{|\Delta|}\right)\Bigg\}
\end{eqnarray}
with  $\partial\Delta$ being the boundary of the hypercube $\Delta$ and $\mathcal{F}:=\sum_{x\neq 0}|F(0,x)|$ which is bounded due to the property \eqref{proper:summ} and $R_F=R$. 

 Furthermore, inequality \eqref{bound:GenIs} implies a new UQ formula for systems with a fixed configuration outside of the domain that here is considered as a Dirichlet-type boundary condition. In particular it allows us to quantify the effect of the boundary conditions on $\partial\Delta$ on the QoIs, as  can be seen more clearly when $\tilde{h}=h$. Note, the term $\frac{|\partial\Delta|}{|\Delta|}$ in \eqref{bound:GenIs} comes from a more careful bound  on the KL divergence using  Lemma~\ref{lemma:important} while this term  has been eliminated  in \eqref{bound:GenIs'} due to the relative crudeness of the bound of KL in Lemma~\ref{lemma:fornorm}, see also Fig.~\ref{fig:STATMECH1}. 
\subsection{ UQ for Phase Diagrams} Here we capture the phase diagram of the  perturbed model $\tilde{q}_{\Delta} $ looking at the magnetization defined in subsection~\ref{subsec:qoi}. We study the limit of  the bounds obtained in subsection~\ref{scalability}. The high-dimensionality of statistical mechanics models requires {\bf scalable bounds} at the thermodynamic limit. In fact, the MGF and the KL divergence scale correctly with the size of the system $|\Delta|$ (all are multiplied by $|\Delta|$ see \eqref{scalMGF}, Lemma~\ref{lemma:fornorm} and \eqref{eq:c2}). Let $M(\tilde{\mathbf{J}}^{\mathbf{F}},\tilde\beta, \tilde h)$ be the limit as $\Delta\nearrow\mathbb{Z}^d$ of $E_{\tilde{q}_{\Delta}}[m(\sigma_{\Delta})]$. Then the limit  $\Delta\nearrow\mathbb{Z}^d$ of \eqref{bound:GenIs}:
\begin{eqnarray}\label{limbound:GenIs}
\pm M(\tilde{\mathbf{J}}^{F},\beta, \tilde{h})\leq \frac{1}{1- \beta (\tilde{h}-h)}\inf_{\lambda>0}\Bigg\{ \frac{\left(P_{h\pm\frac{\lambda}{\beta},\beta,\mathbf{J}}-  P_{h,\beta,\mathbf{J}}\right)}{\lambda/\beta}+\frac{\beta}{\lambda}\mathcal{F}\Bigg\}
\end{eqnarray}
with  $
\lim_{\Delta\nearrow\mathbb{Z}^d}P_{h,\beta,\mathbf{J}}^{\,\Delta}=P_{h,\beta,\mathbf{J}}
$ by Theorem 2.3.3.1 in \cite{errico} and $\lim_{\Delta\nearrow\mathbb{Z}^d}\frac{|\partial\Delta|}{|\Delta|}=0
$, while in the limit of  \eqref{bound:GenIs'} the thermodynamic pressure is only replaced by its limit $P_{h,\beta,\mathbf{J}}
$. The bounds for the $\tilde{\beta}\neq\beta$ can be adjusted similarly. 

 \subsection{Ising-Kac Model} \label{subsec:Kac} Here we consider an Ising-spin model with a Kac-type interaction as a baseline model. Such a model combines sufficient
complexity–since it is not a mean field model –but it is still analytically fairly tractable
to serve as a good benchmark problem for high-dimensional rMRF. We illustrate the uncertainty area of the phase diagram for both \eqref{limbound:GenIs} and the limit of \eqref{bound:GenIs'}. when the alternative models are a Kac perturbation and a truncated Kac interaction.

An Ising-spin model with a Kac-type interaction behaves like a mean field (or Van der Waals model in gas lattice) in the limit with the convexity of free energy emerging naturally in the limit, contrary to mean field or Curie-Weiss models where Maxwell's equal area law is required to refine the non-convex free energy (double well shape), \cite{errico}. Such a discrepancy comes from the fact that each spin interacts with all particles in the same way  and independently. The idea of Kac was to keep such a picture on large regions but relatively small compared to the range of interaction. Then, the thermodynamically incorrect of the free energy  (i.e. the non-convex free energy) on these large regions looks refined at the scale of interaction. Therefore, the system contains a two-scale behavior that was carried out by introducing a small parameter $\gamma>0$ known as {\it Kac scaling}. As we suppose that an Ising spin model is endowed by such an interaction, the model  has overall three scales: the lattice spacing is 1, the range of interaction is $\gamma^{-1}$ while the size of the system is much larger than $\gamma^{-1}$ and all are well-separated, contrary to the mean field model where the range of interaction is the same as the size of the system.  Next, we formally introduce the model.
\medskip

\subsection{Mathematical Background of Ising-Kac Model} A Kac-type interaction  is defined as
 \begin{equation}\label{kac}
J_{\gamma}(x,y)=\gamma^d J(\gamma x,\gamma y), \;\;\;x,y\in \mathbb{Z}^d
\end{equation}
where $\gamma$ is a positive parameter sufficiently small and $J$ is a non-negative (ferromagnetic interaction), even, symmetric function (i.e $J(r,r')=J(r',r)$ for every $r,r'\in \mathbb{R}^d$), translational invariant (i.e $J(r,r')=J(r'+a,r+a)$ for every $r,r'\in \mathbb{R}^d$ and $a\in \mathbb{R}^d$) function such that
$J(r)=0$ for all $|r|>1$, $\int_{\mathbb{R}^{d}}J(r)dr=\mathcal{J}$ and $J\in C^2(\mathbb{R}^d)$. The use of $\mathbf{J}_{\gamma}$ stands for the collection of $J_{\gamma}(x,y)$, that is $\mathbf{J}_{\gamma}=\{J_{\gamma}(x,y)\}_{\mathbb{Z}^d\times\mathbb{Z}^d}$. As $\gamma$ becomes smaller, more particles are included in a spin neighborhood with $\gamma^{-1}$ diameter and while the strength of the interactions becomes weaker.
\medskip

\noindent Let $\Delta$ be a bounded, $\mathcal{P}^{(l)}_{\mathbb{R}^d}$-measurable region, with $L>>\gamma^{-1}$ (see Appendix~\ref{subsec:CG}), $\beta>0$ be the inverse temperature,  $h\in\mathbb{R}$ be the external magnetic field and $ \bar\sigma_{\Delta^{c}}$ be a given configuration on its complement (see Figure~\ref{fig:Range} with $R=\gamma^{-1}$).
\medskip

\noindent{\bf Hamiltonian energy.}
The Hamiltonian energy of a spin configuration $\sigma_{\Delta}$ given $\bar\sigma_{\Delta^{c}}$:
\begin{eqnarray}\label{Hamiltonian}
H_{\gamma}^{\mathbf{J},h}(\sigma_{\Delta}\mid\bar\sigma_{\Delta^c})&=&-\frac{1}{2}\sum_{x\neq y\in\Delta}
J_{\gamma}(x,y)\sigma_{\Delta}(x)\sigma_{\Delta}(y)-\sum_{\substack{x\in\Delta, \\ y\in\Delta^{c}}}J_{\gamma}(x,y)\sigma_{\Delta}(x)\bar\sigma_{\Delta^c}(y)\nonumber\\
&&-h\sum_{x\in\Delta}\sigma_{\Delta}(x),\qquad\textrm{Hamiltonian enery}.
\end{eqnarray}

\noindent{\bf Finite volume Gibbs measure.} The Gibbs measure given a fixed boundary condition $\bar\sigma_{\Delta^c}$ is defined as follows:
\begin{equation}\label{gibbs}
\mu_{\mathbf{J},\beta, h}^{\mathbf{}\Delta,\gamma}(\cdot\mid \bar\sigma_{\Delta^{\rm{c}}})=\frac{1}{Z_{\bar\sigma_{\Delta^{\rm{c}}}}(\mathbf{J},\beta, h)}e^{-\beta H_{\gamma}^{\mathbf{J},h}(\sigma_{\Delta};\bar\sigma_{\Delta^{c}})},\qquad\textrm{finite volume Gibbs measure}
\end{equation}
where  $Z_{\bar\sigma_{\Delta^{\rm{c}}}}(\mathbf{J},\beta, h)$ is the normalization (partition function). To simplify the notation, we shall often drop $\gamma$ and the given configuration in the complement of $\Delta$ from the Gibbs measure, resuming the full notation when needed, and therefore we  write $\mu^{J,h}_{\beta,\Delta}\equiv \mu^{\bar\sigma_{\Delta^{c}},J,h}_{\beta,\Delta,\gamma}$. 
\medskip

\noindent {\bf Thermodynamic pressure.} The thermodynamic pressure for the Ising-Kac model denoted by $P_{\mathbf{J},\beta,h}^{\,\Delta,\gamma}$  is defined as
\begin{equation}\label{pressure}
P_{\mathbf{J},\beta,h}^{\,\Delta,\gamma}(\bar\sigma_{\Delta^c}):=\frac{\log Z_{\bar\sigma_{\mathbf{I}^{\rm{c}}}}(\mathbf{J},\beta, h)}{\beta|\Delta|}
\end{equation}
Its Lebowitz-Penrose (LP) limit (i.e $\lim_{\gamma\to0} \lim_{\Delta\nearrow\mathbb{Z}^d}$)  $p_{\mathbf{J},\beta,h}$ is  given by
\begin{equation}\label{pre}
p_{\mathbf{J},\beta,h}:=-\inf_{m\in[-1,1]}\{-hm+\phi_{\mathbf{J},\beta,0}(m)\},\qquad \phi_{\mathbf{J},\beta,h}(m):=\left\{-\frac{\mathcal{J}}{2}m^2-hm\right\}-\frac{1}{\beta}I(m)
\end{equation}
\noindent (see also Appendix~\ref{THERPRES} for further discussion). The rMRF formulation of such a model and its perturbations considered next is structured analogously to the ones in subsection~\ref{subsubsec:rmrf} and for that we omit  it. 
\begin{figure}[H]
\centering
\vspace{-2.6cm}
 \includegraphics[width=0.6\linewidth]{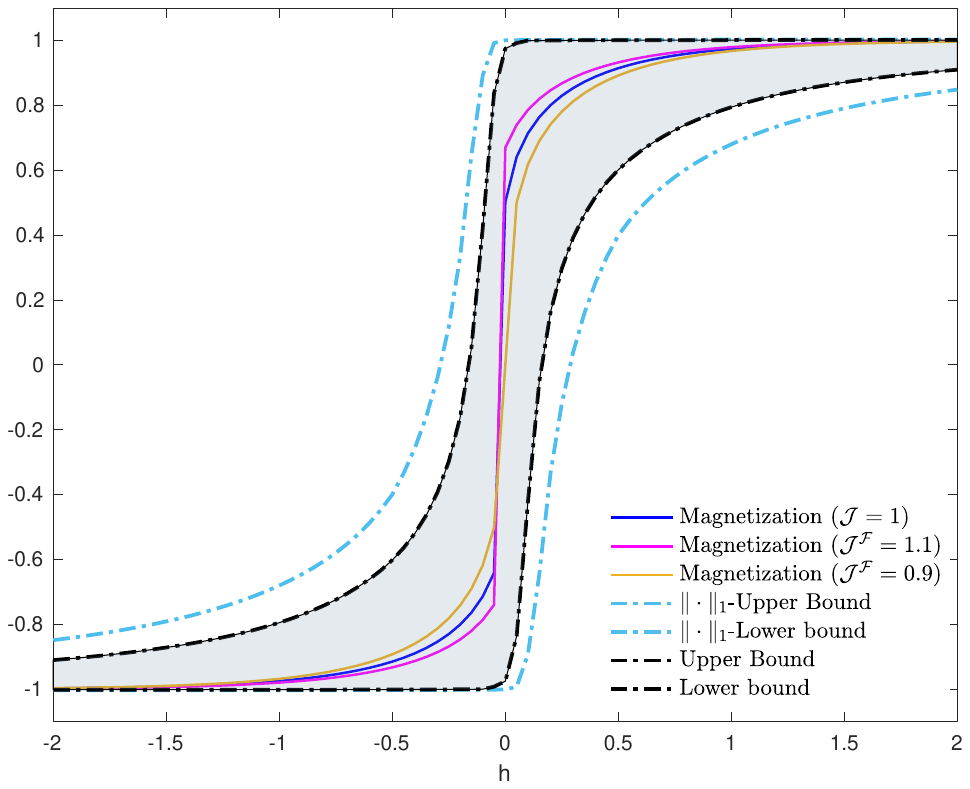}
    \vspace{-2.6cm}
  \caption{{\small The curves in blue, magenta and dark yellow color are the magnetizations of the Ising model with Kac interaction at inverse temperature $\beta=\tilde{\beta}=1.1$, $\tilde{h}=h$, and total strength $\mathcal{J}=1$, $\tilde {\mathcal{J}}^{F}=1.1\;(a=0.1)$ and $0.9\; (a=-0.1)$ (validation) respectively. The black dashed-dot curves are the UQ upper and lower bounds provided by Corollary~\ref{Corotothm1new'} and viewed as functions of $h\in[-2,2]$. The gray area depicts the size of the uncertainty region. The light blue dashed-dot curves are the UQ upper and lower bounds obtained using norm-$\|\cdot\|_{1}$.  The uncertainty area of the phase diagram in grey color is significantly better than the uncertainty area between the light blue dashed-dot curves.  This comes from the fact that the difference between the limit of  \eqref{bound:GenIs'} and \eqref{limbound:GenIs} lies on the term $\frac{\beta}{\lambda}\mathcal{F}$ which is multiplied by 2. }}
  \label{fig:STATMECH1}
\end{figure}

\subsubsection{\bf Phase Diagram of Perturbed Kac model} Let define a perturbation of a Kac potential.
\begin{definition}\label{def:type1} Let $F_{\gamma}$ be an even function satisfying \eqref{proper:transinva}-\eqref{proper:summ} and \eqref{kac} with length of range $\gamma^{-1}$ and $\mathcal{F}:=\int_{\mathbb{R}^d}F(r)dr$. We define
\begin{equation}\label{Kac2}
\tilde J_{\gamma}^{F}(x,y)=J_{\gamma}(x,y)+F_{\gamma}(x,y),\; \text{such that} \;\mathcal{F}=a\mathcal{J},\quad a\in[-1,1]
\end{equation}
\end{definition}
The parameter $a$ represents the percentage of increase or decrease of the total strength of interaction $\tilde {\mathcal{J}}^{F}:=\int_{\mathbb{R}^d}\tilde{J}^{F}(r)dr=(1+a)\mathcal{J}$.
\begin{corollary}\label{Corotothm1new'}
Let $\tilde{J}^{F}$ be the interaction given in Definition~\ref{def:type1}. Then, for $\gamma>0$ small enough, the UQ bounds \eqref{bound:GenIs'} and \eqref{bound:GenIs} hold for $R_F=R=\gamma^{-1}$ and $\mathcal{F}=|a|\mathcal{J}$.  The thermodynamic pressure $P_{\mathbf{J},\beta,h}^{\,\Delta,\gamma}$  is given in \eqref{pressure}. Let $M(\tilde{\mathbf{J}}^{F},\beta, \tilde h)$ be the LP-limit of $E_{\tilde{q}_{\Delta}}[m(\sigma_{\Delta})]$. Then, the UQ bounds  \eqref{limbound:GenIs}  and  LP-limit of \eqref{bound:GenIs'} hold with the LP-limit of $P_{\mathbf{J},\beta,h}^{\,\Delta,\gamma}$ being $p_{\mathbf{J},\beta,h}$ given in \eqref{pre}. 
\end{corollary}
 \begin{remark}
 \eqref{bound:GenIs'} represents crude bounds as norm-$\|\cdot\|_{1}$ (subsection~\ref{subsec:REwnorm}) has been used, while \eqref{bound:GenIs} obtained by Theorem~\ref{Mainthm1}, includes more detail. The difference is illustrated in Figure~\ref{fig:STATMECH1}. Furthermore, even if there is a $\gamma^{-1}$ in the term $2 \gamma^{-1} \frac{|\partial\Delta|}{|\Delta|}$ in \eqref{bound:GenIs'}, the order of the LP-limit makes it vanish as $L\to\infty$.
 \end{remark}

{\bf Validation.} Given $\beta$,  $h$, $\mathcal{J}$ and a tolerance $\eta>0$, we can construct  with the use of norm-$\|\cdot\|_{1}$ and Lemma~\ref{lemma:fornorm} a class of models such that 
$\mathcal{Q}^{\mathrm{I}}_{\eta}:=\{\tilde{q}_{\Delta}:2\beta a\mathcal{J}\leq\eta\}$. This is subclass of $\mathcal{Q}^{\eta}$ defined in \eqref{eq:ambiguity:0} with the KL divergence in place of $d$. In Figure~\ref{fig:STATMECH1}, $\beta=1.1$ and $\mathcal{J}=1$ while the external field $h$ varies from $-2$ and $2$. The positive parameter $\eta=0.1$ and the perturbed model with $10\%$ decrease ($a=-0.1$) of the total strength (magnetization in magenta color) is in $\mathcal{Q}^{\mathrm{I}}_{0.1}$ as demonstrated in dark yellow color.

\begin{figure}[H] 
\centering
 \vspace{-2.1cm}
  \includegraphics[width=0.45\linewidth]{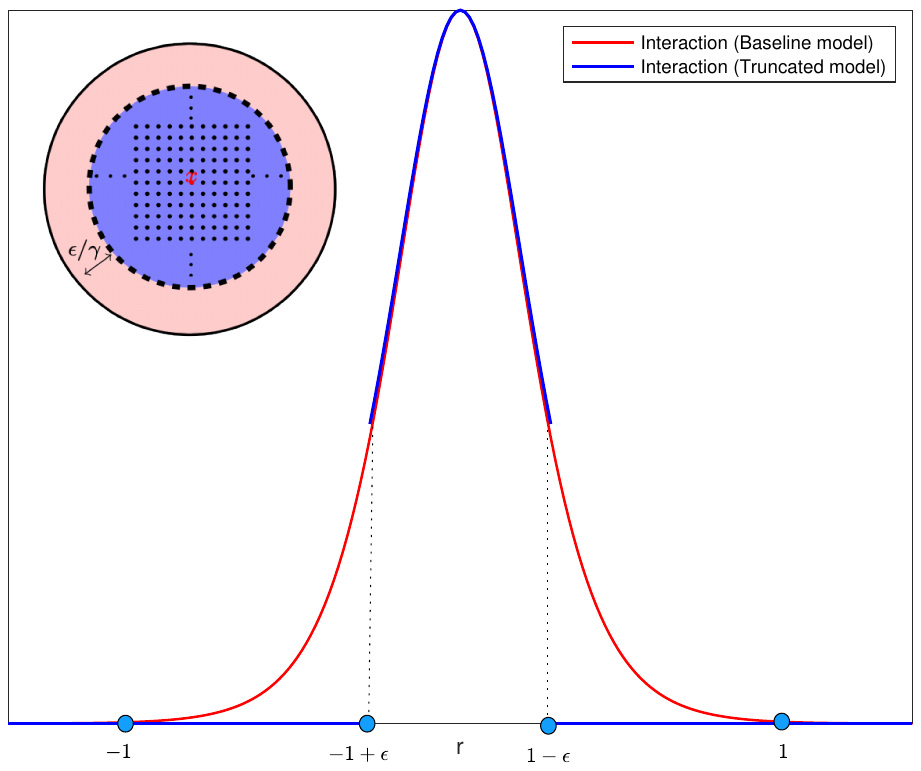}
    \includegraphics[width=0.45\linewidth]{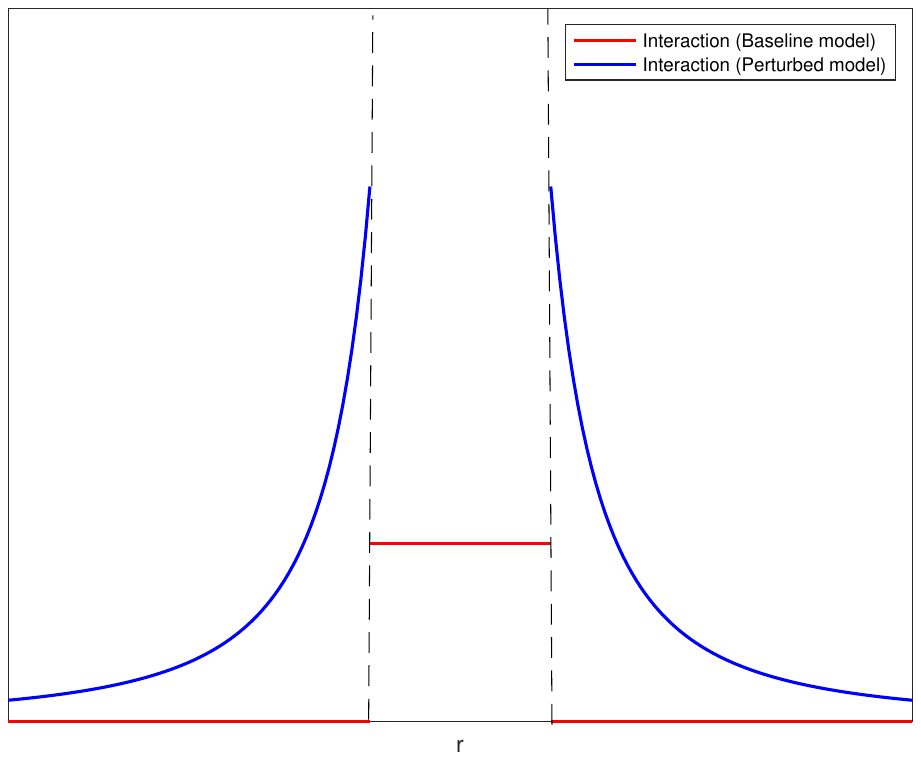}
  \vspace{-2.1cm}
    \caption{{\small (Left) The red curve is a Kac interaction and the blue curve is a truncation of it. The two curves coincide at all $r$ with $|r|\leq 1-\epsilon$. The embedded picture demonstrates the two interactions at the microscopic level. The red particle located at the site $x\in\Delta\subset\mathbb{Z}^2$ interacts with the particles in the blue and the light red through $J_{\gamma}$.The particle interacts only with the particles in the blue area through $\tilde{J}^{-J}_{\gamma}$ with range $\gamma^{-1}(1-\epsilon)$. (Right) The red curve is an example of Kac interaction (piecewise constant) with $J(r)={\bf 1}_{r\leq\frac{1}{2}}(r)$ and the blue curve is a perturbation given by $G(r)=\frac{a}{r^2}{\bf 1}_{r>\frac{1}{2}}(r)$,for some $a>0$.}}
  \label{fig:kacpic1}
\end{figure}
\subsubsection{\bf  Phase diagram of Truncated Potential} 
\begin{figure}[H] 
\vspace{-1.4cm}
  \includegraphics[width=0.32\linewidth]{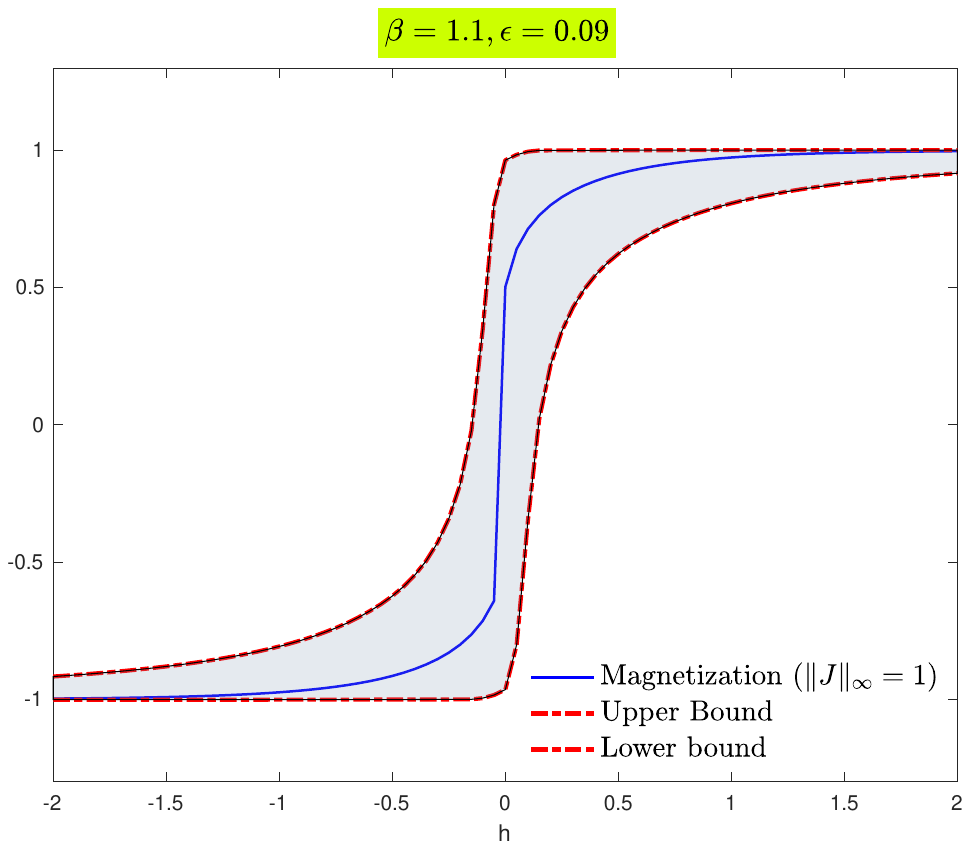}
  \includegraphics[width=0.32\linewidth]{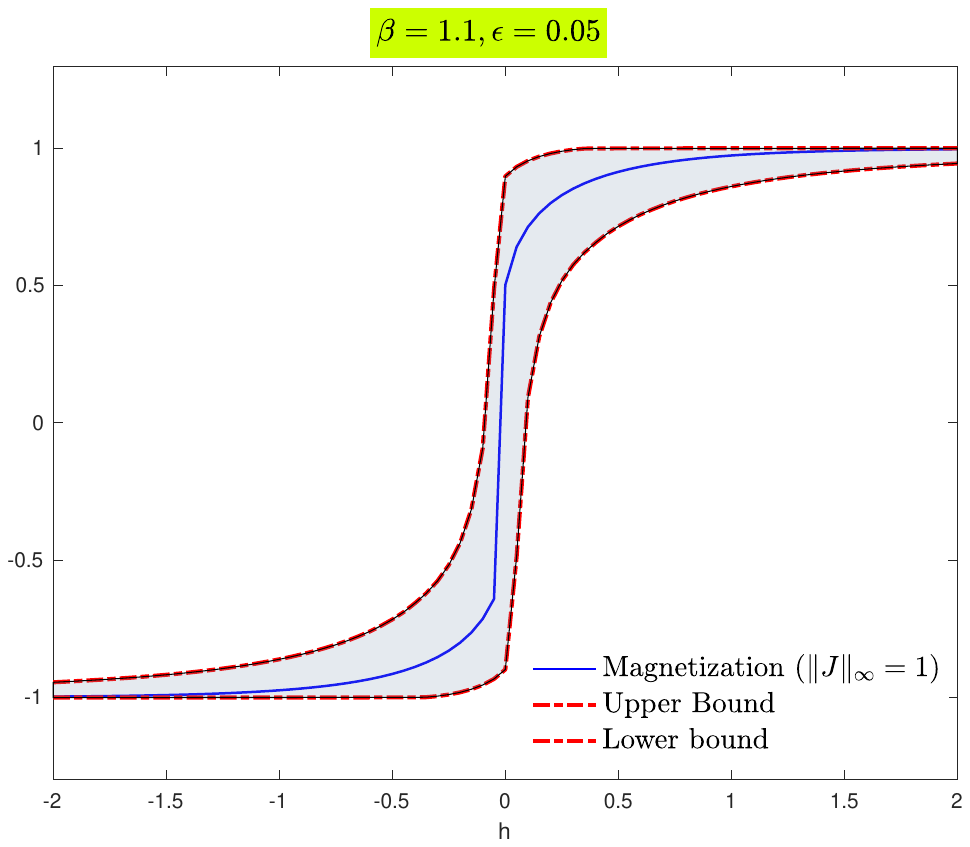}
  \includegraphics[width=0.32\linewidth]{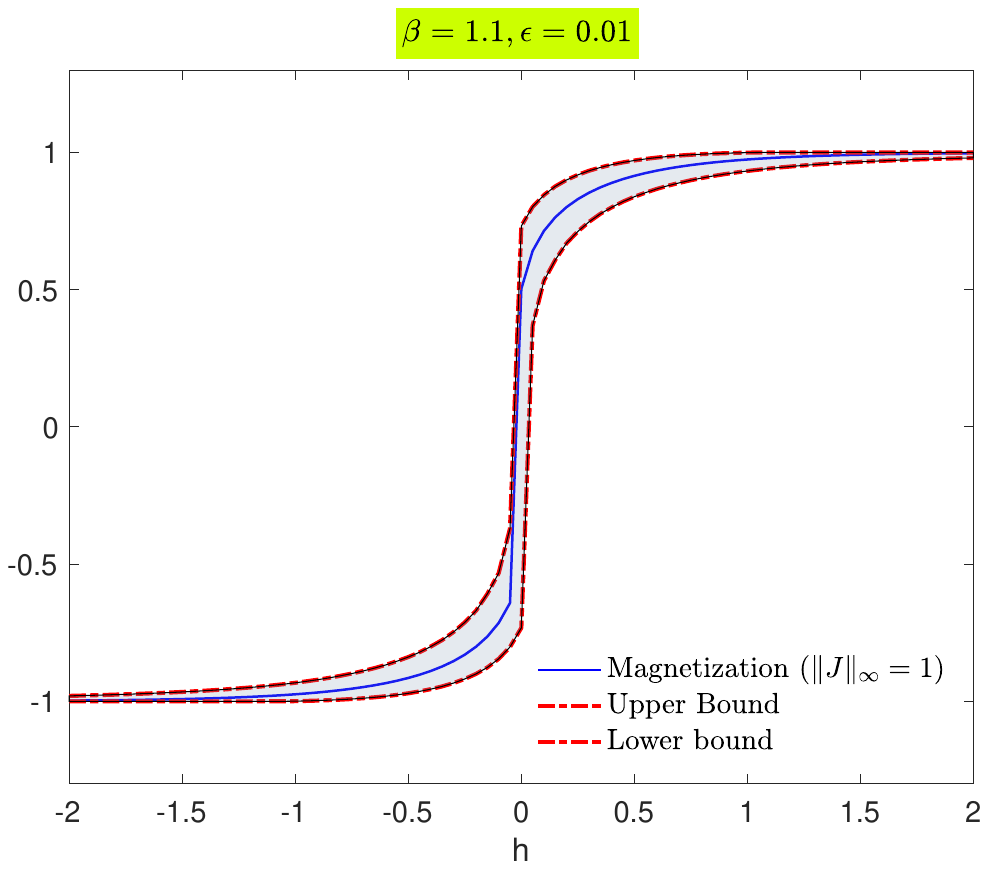}
\vspace{-1.3cm}
    \caption{{\small The three graphs demonstrate the uncertainty area in gray color for different values of $\epsilon$. In all graphs, the blue solid line is the magnetization of $d$-sing model with Kac interaction at inverse temperature $\beta=1.1$, $\|J\|_{\infty}=1$ and $\tilde{h}=h$.  The black dashed-dot curves are the upper and lower bound of magnetization of the truncated interaction ${\tilde{J}}^{-J}$, viewed as functions of $h$. (Left) $\epsilon=0.09$. (Center) $\epsilon=0.05$. (Right) $\epsilon=0.01$.}}
  \label{fig:epsilon}
\end{figure}

From a computational point of view, macroscopic properties of high dimensional systems can be studied through simulation models where one can consider an appropriate truncated interaction which can reduce the computational overhead associated with the interaction \cite[Chapter~3]{T}. In our context, a truncated interaction can be thought of as: The support of the interaction $J$ is $[-1,1]$ as in Fig. ~\ref{fig:kacpic1}.  $J$ is cut off at $1-\epsilon$ and $-1+\epsilon$ for some  parameter $\epsilon\in[-1,1]$. Then the resulting interaction is called truncated interaction of $J$ and its support is $[-1+\epsilon, 1-\epsilon]$ of length $2\epsilon$. The introduced parameter $\epsilon$ quantifies the impact of the truncation of the interaction $J$. Moreover, Fig~\ref{fig:epsilon}  quantifies how the uncertainty area becomes smaller as $\epsilon$ becomes smaller (and hence the truncated interaction tends to be the original $J$). We mathematically define such an interaction as follows:
\begin{definition}\label{def:trunc} Let $0<\epsilon<1$.  We define the truncated interaction as
\begin{equation}\label{inter_type1}
\tilde J^{-J}(0,r)=
\left\{
	\begin{array}{ll}
		J(0,r)  &,|r|\leq1-\epsilon\\
		 0 &,\,otherwise
		\end{array}
\right.
\end{equation}
\end{definition}
The truncated model can be viewed as Type II.  However, to be consistent with the assumption $\mathcal{E}\subset\tilde{\mathcal{E}}$ in Definition~\ref{def:types}, we view it as perturbed interaction of Type I arising from the subtraction of $J$ (also explains the notation $\tilde{J}^{-J}$ in \eqref{inter_type1}) on regions of radius greater than $1-\epsilon$ as illustrated in Figure~\ref{fig:kacpic1}.
\begin{corollary}\label{Corotothm1new''}
Let $\tilde{J}^{-J}$ be the interaction given in Definition~\ref{def:trunc}. Then, for $0<\epsilon<1$ and $\gamma>0$ small enough, the UQ bounds \eqref{bound:GenIs'} and \eqref{bound:GenIs} hold for $R_{-J}=\gamma^{-1}$ and $\mathcal{F}\leq \epsilon\|J\|_{\infty}$. The thermodynamic pressure $P_{\mathbf{J},\beta,h}^{\,\Delta,\gamma}$ is given in \eqref{pressure}. Let $M(\tilde{\mathbf{J}}^{F},\beta, \tilde h)$ be the LP-limit of $E_{\tilde{q}_{\Delta}}[m(\sigma_{\Delta})]$. Then, the UQ bounds  \eqref{limbound:GenIs}  and LP-limit of \eqref{bound:GenIs'} hold with the limit of $P_{\mathbf{J},\beta,h}^{\,\Delta,\gamma}$ being $p_{\mathbf{J},\beta,h}$ given by \eqref{pre}. 
\end{corollary}
\begin{remark}
Given $\beta,\|J\|_{\infty}$, we can choose $\epsilon\equiv\epsilon(\beta,\|J\|_{\infty})$ sufficiently small.  Consequently, the phase diagram of the two models are close to each other as the uncertainty area is very small (Figure~\ref{fig:epsilon}). The parameter $\epsilon$ quantifies the length of the area that one cuts off the initial interaction. 
\end{remark}
The same methods are applicable to other perturbations, e.g. the very long range in Appendix~\ref{sec:long} and perturbations in "contexts"/configuration as boundary conditions.
\smallskip

{\bf Conclusion and future work.}  In this article, we developed an information-based uncertainty quantification method for Markov Random Fields/rMRFs. We considered
a surrogate (baseline) MRF/rMRF constructed by physical modeling or by learning structure
and parameters from data, and we quantify uncertainties inherited from data, modeling
choices, or numerical approximations, that are also propagated in predictions for QoIs. Our UQ method quantifies uncertainties not only in parameters but also in structure as well as is capable in handling of the inherent high-dimensionality of systems that admit a MRF/rMRF formulation. This was achieved by obtaining tight and scalable, information-based bounds on the predictions for QoIs. 

We demonstrated our UQ method in an example from medical diagnostics as well as several high dimensional equilibrium statistical mechanics models defined on bounded domains with suitable boundary conditions. We aim to extend the developed approach to non-equilibrium statistical mechanics systems \cite{Presutti} also arising in ML \cite{BahriGanguli}. Furthermore, motivated by \cite{FLKV} we plan to develop robust uncertainty quantification for Bayesian networks defined on Directed Acyclical Graphs.
\smallskip

{\bf Acknowledgments:} The research of M. K. was partially supported by the NSF HDR  TRIPODS CISE-1934846. The research of P. B. and M. K., was partially supported  by the Air Force Office of Scientific Research (AFOSR) under the grant FA-9550-18-1-0214.

%------------------------------------------------------------
%APPENDIX----------------------------------------------------
\appendix
\section{Reduced Markov Random Fields (rMRFs)}\label{sec:PII}
Let $\mathbf{Y}=\{Y_i\}_{i\in \mathcal{V}}$ be a MRF indexed by a set of nodes $\mathcal{V}$ (finite or infinite) of a graph $\mathcal{G}$. Let us consider $\mathcal{M}\subset\mathcal{V}$. Let also $\mathbf{U}=\{Y_i\}_{i\in\mathcal{M}}$ and $\mathbf{u}$ be an assignment to them, namely $\mathbf{U}=\mathbf{u}$. If $\mathbf{Z}:=\{Y_i\}_{i\in\mathcal{V}\setminus\mathcal{M}}$, how does the underlying graph corresponding to  $\mathbf{Z}\mid\mathbf{U}=\mathbf{u}$ look like? Can the conditional probability $p(\mathbf{z}\mid\mathbf{U}=\mathbf{u})$ still keep a product structure/factorization as the joint distribution given in \eqref{distributioncliques}? To answer the questions, we need a special class of MRF which is called {\it reduced Markov Random Fields} (rMRFs). 
\begin{figure}[h]  
\centering
\begin{tikzpicture}[node distance=1cm,
                   main node/.style={circle,top color =white , bottom color = black, draw,font=\sffamily\tiny\bfseries}]
       
  \node[main node, white] (1) {1};
  \node[main node, white] (2) [above right of=1] {2};
  \node[main node, white] (3) [below right of=1] {3};
  \node[main node, white] (4) [right of=3] {4};
  \node[main node, white] (5) [above right of=4] {5};
  \node[main node, white] (6) [below right of =4] {6};
  \node[main node, white] (7) [below left of=6] {7};
  \node[main node, white] (8) [above right of=5] {8};
  \node[main node, white] (9) [above right of=8] {9};
  \node[main node, white] (10) [below right of=8] {10};

  \path[every node/.style={font=\sffamily\small}]
    (1) edge[ultra thick] node [] {} (2)
    (2) edge[ultra thick] node [] {} (3)
    (3) edge[ultra thick] node [] {} (4)
    (3) edge[ultra thick] node [] {} (7)
    (3) edge[ultra thick] node [] {} (6)
    (4) edge[ultra thick] node [] {} (6)
         edge[ultra thick]  node [] {} (5)
    (5) edge[ultra thick]  node [] {} (8)
    (5) edge[ultra thick]  node [] {} (6)
    (9) edge[ultra thick]  node [] {} (10)
    (8) edge[ultra thick]  node [] {} (9)
    (6) edge[ultra thick] node [] {} (7)
    (8) edge[ultra thick]  node [] {} (10)
    (5) edge[ultra thick]  node [] {} (8);    
\end{tikzpicture}
\hspace{1cm}
\begin{tikzpicture}[node distance=1cm,
                   main node/.style={circle,draw,font=\sffamily\tiny\bfseries}]
  \node[main node, white] (1) [top color =white , bottom color = black]{1};
  \node[main node, white] (2) [top color =white , bottom color = black,above right of=1] {2};
  \node[main node, white] (3) [top color =white , bottom color = black,below right of=1] {3};
  \node[main node,gray!30!white] (4) [top color =white , bottom color = gray!30!white,right of=3] {4};
  \node[main node, white] (5) [top color =white , bottom color = black,above right of=4] {5};
  \node[main node, white] (6) [top color =white , bottom color = black,below right of =4] {6};
  \node[main node, white] (7) [top color =white , bottom color = black,below left of=6] {7};
  \node[main node, white] (8) [top color =white , bottom color = black, above right of=5] {8};
  \node[main node,gray!30!white] (9) [top color =white , bottom color = gray!30!white,above right of=8] {9};
  \node[main node, white] (10) [top color =white , bottom color = black,below right of=8] {10};

  \path[every node/.style={font=\sffamily\small}]
    (1) edge[ultra thick] node [yellow,] {} (2)
    (2) edge[ultra thick] node [] {} (3)
    (3) edge[gray!30!white] node [] {} (4)
    (3) edge[ultra thick] node [] {} (7)
    (3) edge[ultra thick] node [] {} (6)
    (9) edge[gray!30!white] node [] {} (8)
    (5) edge[ultra thick]  node [] {} (8)
    (5) edge[ultra thick]  node [] {} (6)
    (9) edge[gray!30!white] node [] {} (10)
    (4) edge[gray!30!white] node [] {} (6)
    (4) edge[gray!30!white] node [] {} (5)
    (6) edge[ultra thick] node [] {} (7)
    (8) edge[ultra thick]  node [] {} (10)
    (5) edge[ultra thick]  node [] {} (8);    
\end{tikzpicture}
\caption{\small {The set of nodes is $\mathcal{V}=\{1,\cdots, 10\}$ and $\mathcal{M}=\{4,9\}$}. Left: $\mathbf{Y}=\{Y_i\}_{i=1}^{10}$ with joint distribution $p$ is a MRF over $\mathcal{G}$. The set of maximal cliques is given by $\mathcal{C}_{\mathcal{G}}=\left\{\{1,2\}, \{2,3\}, \{3,4,6,\}, \{3,6,7\},\{4,5,6\},\{5,8\},\{8,9,10\}\right\}$. Right: $\mathbf{Z}=\{Y_i\}_{i\in\mathcal{V}\setminus\mathcal{M}}$ with joint distribution $q$ is the corresponding rMRF over $\mathcal{G}'$ with $\mathbf{U}=\{Y_4, Y_9\}$ and $\mathbf{u}=\{u_4, u_9\}$. The rMRF is demonstrated by  removing the node 4 and 9 (faded nodes) from the graph $\mathcal{G}$. $\mathcal{C}_{\mathbf{U}}=\left\{\{3,4,6,7\},\{4,5,6\},\{8,9,10\}\right\}$ while $\mathcal{C}_{\emptyset}=\left\{\{1,2\},\{2,3\},\{5,8\}\right\}$.}\label{fig:K}  
\end{figure}
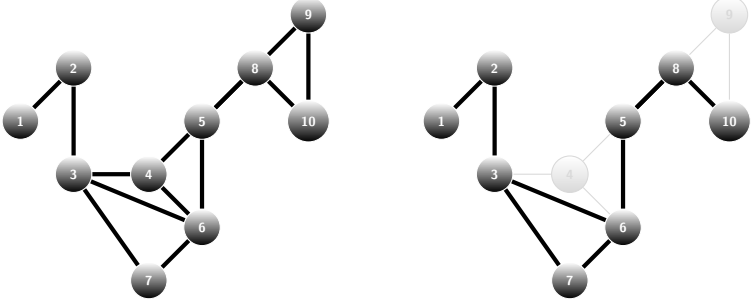

\begin{definition}\label{rMRF}
Let $\mathbf{Y}=\{Y_i\}_{i\in \mathcal{V}}$  be a collection of random variables indexed by a set of nodes $\mathcal{V}$ (finite or infinite) of a graph $\mathcal{G}$. If  $(\mathbf{Y}, p)$ is a MRF, $\mathbf{u}$ a context,  $\mathcal{M}\subset\mathcal{V}$ and $\mathbf{U}=\{Y_i\}_{i\in\mathcal{M}}$, we define as  {\rm{\bf reduced Markov Random Field}},  a MRF $\mathbf{Z}=\{Y_i\}_{i\in\mathcal{V}\setminus\mathcal{M}}$ indexed by the set of nodes $\mathcal{V}\setminus\mathcal{M}$ of the subgraph $\mathcal{G}[\mathcal{V}\setminus\mathcal{M}]$ with joint distribution $\mathbb{Q}$ such that 
\begin{equation}
q(\mathbf{z})\equiv \mathbb{Q}(\mathbf{Z}=\mathbf{z}):=p(\mathbf{z}\mid\mathbf{U}=\mathbf{u}).
\end{equation}
\end{definition}
Therefore, $\mathbf{Z}\mid\mathbf{U}=\mathbf{u}$ could be thought as a induced subgraph of $\mathcal{G}$ with set of nodes $\mathcal{V}\setminus\mathcal{M}$, that is  eliminating any node corresponding to random variables $\mathbf{U}$ and any edge adjacent to them. Furthermore, according to Definition~\ref{rMRF}, $\mathbf{Z}$ is clearly MRF and therefore the conditional probability $p(\mathbf{z}\mid\mathbf{U}=\mathbf{u})$ is expected to have a product structure. All the above are summarized in the following proposition:

\begin{proposition}\label{thmrMRF}
Let $\mathbf{Y}$ be a MRF with probability distribution $p>0$ parametrized by some parameters $\mathbf{w}=\{\mathbf{w}_c\}_{c\in\mathcal{C}_{\mathcal{G}}}$ given in \eqref{distributioncliques} and let $\mathbf{U}, \mathbf{Z}$ be defined as in the beginning of the subsection. Then, $q$ parametrized by $\mathbf{w}$ is expressed as
\begin{equation}\label{distributioncliquesrMRF}
q^{\mathbf{w}}(\mathbf{z})\equiv p(\mathbf{z}\mid\mathbf{U}=\mathbf{u},\mathbf{w})=\frac{1}{Z_{\mathbf{u}}(\mathbf{w})}\prod_{c\in\mathcal{C}_{\mathcal{G}}}\Psi_{c}[\mathbf{u}](\mathbf{z}_{c}\mid\mathbf{w}_c)
\end{equation}
where for every $c\in \mathcal{C}_{\mathcal{G}}$
\begin{equation}\label{rclique}
\Psi_{c}[\mathbf{u}](\mathbf{z}_{c}\mid\mathbf{w}_c):=\Psi_{c}(\mathbf{z}_{c},\mathbf{u}_{c}\mid\mathbf{w}_c)
\end{equation}
Moreover, $Z_{\mathbf{u}}(\mathbf{w})$ is given by 
\begin{equation}\label{rpartitioncliques}
Z_{\mathbf{u}}(\mathbf{w})=\sum_{\mathbf{Y}}\prod_{c\in\mathcal{C}_{\mathcal{G}}}\Psi_{c}[\mathbf{u}](\mathbf{z}_{c}\mid\mathbf{w}_c)
\end{equation}
\end{proposition}

 We refer to \cite{KF} and \cite{Murphy} for further discussion about MRFs, rMRFs and the proof of the Hammersley-Clifford Theorem and Propostion~\ref{thmrMRF}.

 \subsection{\bf Partition of the class of maximal cliques}\label{subsubsec:partition} We further investigate the structure of the class of all maximal cliques. Precisely, we collect  $c\in\mathcal{C}_{\mathcal{G}}$ such that $\mathbf{U}\cap \mathbf{Y}_c\neq\emptyset$. This leads to partition the set of maximal cliques $
 \mathcal{C}_{\mathcal{G}}=\mathcal{C}_{\mathbf{U}}\sqcup \mathcal{C}_{\emptyset} $
 with
 \begin{equation}\label{classofcliques}
 \mathcal{C}_{\mathbf{U}}=\{c:\mathbf{U}\cap \mathbf{Y}_c\neq\emptyset\} \;\;\mathrm{and} \;\;\;\mathcal{C}_{\emptyset}=\{c:\mathbf{U}\cap \mathbf{Y}_c=\emptyset\}.
 \end{equation}
 (see example shown in Figure~\ref{fig:K}). On top of that, the partition of $\mathcal{C}_{\mathcal{G}}$ makes the joint distributions $q$ take the form
 \begin{equation}\label{classwithpart}
q(\mathbf{z})=P_{\Psi}^{\mathbf{w}}[\mathbf{u}](\mathbf{z})=\frac{1}{Z_{\mathbf{u}}(\mathbf{w})}\prod_{c\in\mathcal{C}_{\emptyset}}\Psi_{c}(\mathbf{y}_{c}\mid\mathbf{w}_c)\prod_{c\in\mathcal{C}_{\mathbf{U}}}\Psi_{c}[\mathbf{u}](\mathbf{z}_{c}\mid\mathbf{w}_c)\end{equation}

\section{Proofs of the main results}\label{App:proofs}
\subsection{Proof of Lemma~\ref{lem:rationPart}}\label{prooflem:rationPart} In the following computation we use either \eqref{AsA} for type I or \eqref{AsB} for type II:
\begin{eqnarray*}
\tilde{Z}_{\mathrm{u}}(\tilde{\mathbf{w}})&=&\sum_{\mathbf{z}}\prod_{\tilde{c}}\tilde{\Psi}_{\tilde{c}}[\mathbf{u}](\mathbf{z}_{\tilde{c}}\mid\tilde{\mathbf{w}}_{\tilde{c}})\nonumber\\
&=&\sum_{\mathbf{z}}\prod_{c}\Psi_{c}[\mathbf{u}](\mathbf{z}_{c}\mid\mathbf{w}_{c})\Phi_{\mathbf{u}}^{\mathrm{i}}(\mathbf{z})\nonumber\\
&=&\sum_{\mathbf{z}}\Phi_{\mathbf{u}}^{\mathrm{i}}(\mathbf{z})\prod_{c}\Psi_{c}[\mathbf{u}](\mathbf{z}_{c}\mid\mathbf{w}_{c})\nonumber\\
&=&Z_{\mathrm{u}}(\mathbf{w})\sum_{\mathbf{z}}\Phi_{\mathbf{u}}^{\mathrm{i}}(\mathbf{z})\prod_{c}\Psi_{c}[\mathbf{u}](\mathbf{z}_{c}\mid\mathbf{w}_{c})\frac{1}{Z_{\mathrm{u}}(\mathbf{w})}\nonumber\\
&=&Z_{\mathrm{u}}(\mathbf{w})E_{q}[\Phi_{\mathbf{u}}^{\mathrm{i}}(\mathbf{z})]
\end{eqnarray*}

\subsection{Proof of Theorem~\ref{Mainthm1}}\label{app:Proofmain}
We are mostly based on the proof of the characterization of the exponential integrals (see, e.g. \cite{DE}). Let the probability measure $R$ be defined by 
\[
dR/dq=e^{f(\mathbf{Z})}/E_{q}[f(\mathbf{Z})].
\]
Note that $\mathcal{R}(\tilde{q}\|q)<\infty,$ since $q,\tilde{q}>0$. Thus, 
\begin{eqnarray}\label{MGF}
-\mathcal{R}(\tilde{q}\|q) + E_{\tilde{q}}[f(\mathbf{Z})]&=&-\mathcal{R}(\tilde{q}\|R)+\log E_{q}[e^{f(\mathbf{Z})}]\leq \log E_{q}[e^{f(\mathbf{Z})}].
\end{eqnarray}
where for the last inequality  we use that $\mathcal{R}(\tilde{q}\|R)\geq 0$ and $\mathcal{R}(\tilde{q}\|R)=0$ iff $\tilde{q}=R$ \cite[Lemma 1.4.1]{DE}.  For part a., we combine \eqref{eq:KLnew} of Lemma~\ref{lemma:RE} and \eqref{MGF} and we get
\[
E_{\tilde{q}}[f(\mathbf{Z})]\leq \log E_{q}[e^{f(\mathbf{Z})}]+\frac{1}{E_{q}[\Phi_{\mathbf{u}}^{\mathrm{i}}]}E_{q}\left[\Phi_{\mathbf{u}}^{\mathrm{i}}\log\Phi_{\mathbf{u}}^{\mathrm{i}}\right]-\log E_{q}[\Phi_{\mathbf{u}}^{\mathrm{i}}]
\]
By replacing $f(\mathbf{Z})$ to $\pm\lambda f(\mathbf{Z})$, we obtain 
\[
\pm E_{\tilde{q}}[f(\mathbf{Z})]\leq \frac{1}{\lambda}\Big\{\log E_{q}[e^{\pm\lambda f(\mathbf{Z})}]+\frac{1}{E_{q}[\Phi_{\mathbf{u}}^{\mathrm{i}}]}E_{q}\left[\Phi_{\mathbf{u}}^{\mathrm{i}}\log\Phi_{\mathbf{u}}^{\mathrm{i}}\right]-\log E_{q}[\Phi_{\mathbf{u}}^{\mathrm{i}}]\Big\}
\]
By optimizing over $\lambda>0$  (see \cite{CD} and \cite{LX}), the following tight estimates are obtained:
\[
\pm E_{\tilde{q}}[f(\mathbf{Z})]\leq \inf_{\lambda>0}\frac{1}{\lambda}\Big\{\log E_{q}[e^{\pm\lambda f(\mathbf{Z})}]+\frac{1}{E_{q}[\Phi_{\mathbf{u}}^{\mathrm{i}}]}E_{q}\left[\Phi_{\mathbf{u}}^{\mathrm{i}}\log\Phi_{\mathbf{u}}^{\mathrm{i}}\right]-\log E_{q}[\Phi_{\mathbf{u}}^{\mathrm{i}}]\Big\}
\]
Part b. is proved similarly, utilizing \eqref{eq:RE} instead of \eqref{eq:KLnew}.

\begin{expl} (Single-parameter exponential families) This is a straightforward example and a simple illustration of the ideas in the proof of  part b., Theorem~\ref{Mainthm1}, giving us insights on how well the ideas work together with a {\it rearranging argument}. The simplicity of this example arises from the fact that the exponential family is single parametric and therefore the structural part is not present. The probability density function of a random variable $X$ with range $R(X)$, is given by 
\[
p^{\theta}(x)=P^{\theta}(X=x)=e^{ \theta \phi(x)-F(\theta)}
\]
taken with respect to some measure $d\nu$ where $F(\theta)=\log\int_{x}e^{ \theta \phi(x)}\nu(dx)$ and $\phi(x)$ is a real-valued function also known as sufficient statistic. Suppose a second probability density function of the same single-parameter exponential family associated with $\phi$  
\[
p^{\theta+\zeta}(x)=P^{\theta+\zeta}(X=x)=e^{ (\theta +\zeta)\phi(x)-F(\theta+\zeta)}
\]
for some $\zeta<1$. One may want to investigate how sensitive the model is in such a change in $\theta$ by $\zeta$ with respect to $\phi(X)$ as means to bound $E_{P^{\theta+\zeta}}[\phi(X)]$ or to find the error in replacing the first distribution by the "perturbed" one and phrased as bound $E_{P^{\theta+\zeta}}[\phi(X)]-E_{P^{\theta}}[\phi(X)]$. The second exponential family is apparently a perturbation on parameters by $\zeta$, so we can think of the model as Type I. In addition, after employing UQ bounds, the cumulant generating function and KL divergence are the two main ingredients to  compute: for any $\lambda>0$,
\begin{align*}
\Lambda_{P^{\theta}}^{\phi}(\lambda)&=\log E_{P^{\theta}}[e^{\lambda\phi(X)}]=F(\theta+\lambda)-F(\theta)\\
R(P^{\theta+\zeta}\|P^{\theta})&=\zeta E_{P^{\theta+\zeta}}[\phi(X)]-\log E_{P^{\theta}}[e^{\zeta\phi(X)}]
\end{align*}
The above expression for KL divergence comes from the calculation of expressing $F(\theta+\lambda)$ in terms of $F(\theta)$ and for that  every  term is computed with respect to $P^{\theta}$. By substituting the quantities to the UQ bounds and by doing a delicate rearrangement of terms that is feasible because the QoI is a sufficient statistic for the model, we get
\[
\pm E_{P^{\theta+\zeta}}[\phi(X)]\leq \frac{1}{1-\zeta} \inf_{\lambda>0}\left\{\frac{F(\theta+\lambda)-F(\theta)}{\lambda}+\frac{1}{\lambda}\log E_{P^{\theta}}[e^{\zeta\phi(X)}]\right\}
\]
\end{expl}
\subsection{Proof of Theorem~\ref{thm:Tightness}}\label{proofofthm:Tightness}
The existence and the explicit form of the distribution $q^{\pm}$ relies on \cite{GKBW}, Theorem 2. Consequently, given a QoI $f$, we identify the total $\tilde{q}$-excess factor relative to $q$ explicitly, that is $\Phi_{\mathbf{u}}^{\pm}=e^{\lambda_{\pm}f}$. However, the new element is that by utilizing the Hammersley-Clifford Theorem, $q^{\pm}$ defined on $\mathbf{Z}$ are rMRFs, lie in the class $\mathcal{Q}_{\mathcal{P}}^{\eta}$ and the total $\tilde{q}$-excess factor relative to $q$ is explicitly determined. 

\section{Coarse-Graining, Kac and Hamiltonian Estimates}\label{CGKHam}
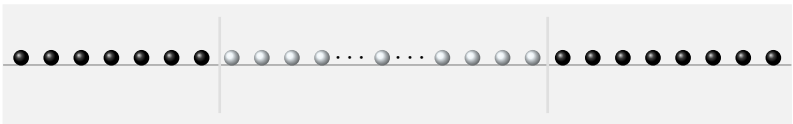
\begin{figure}[H]
\centering
\vspace{.01cm}
	\begin{tikzpicture}[scale=.8]
\draw[fill=lightgray!10!white, lightgray!20!white ] (-2.3,-1) rectangle (10.8,1);
\draw[-,color=gray](-2.3,0)--(10.8,0);
\draw[-,color=lightgray!50!white,line width=1pt](1.3,-.8)--(1.3,.8);
\draw[-,color=lightgray!50!white,line width=1pt ](6.75,-.8)--(6.75,.8);

\shade[ball color=black](-2,0.12)circle(1.3mm);
\shade[ball color=black](-1.5,0.12)circle(1.3mm);
\shade[ball color=black](-1,0.12)circle(1.3mm);
\shade[ball color=black](-.5,0.12)circle(1.3mm);
\shade[ball color=black](0,0.12) circle(1.3mm);
\shade[ball color=black](.5,0.12)circle(1.3mm);
\shade[ball color=black](1,0.12)circle(1.3mm);
\shade[ball color=sky!10!white](1.5,0.12)circle(1.3mm);
\shade[ball color=sky!10!white](2,0.12)circle(1.3mm);
\shade[ball color=sky!10!white](2.5,0.12)circle(1.3mm);
\shade[ball color=sky!10!white](3,0.12) circle(1.3mm);
\node at (3.5, 0.12) { $\cdots$};
\shade[ball color=sky!10!white](4,0.12)circle(1.3mm);
\node at (4.5, 0.12) { $\cdots$};
\shade[ball color=sky!10!white](5,0.12) circle(1.3mm);
\shade[ball color=sky!10!white](5.5,0.12)circle(1.3mm);
\shade[ball color=sky!10!white](6,0.12)circle(1.3mm);
\shade[ball color=sky!10!white](6.5,0.12)circle(1.3mm);
\shade[ball color=black](7,0.12)circle(1.3mm);
\shade[ball color=black](7.5,0.12)circle(1.3mm);
\shade[ball color=black](8,0.12) circle(1.3mm);
\shade[ball color=black](8.5,0.12)circle(1.3mm);
\shade[ball color=black](9,0.12)circle(1.3mm);
\shade[ball color=black](9.5,0.12)circle(1.3mm);
\shade[ball color=black](10,0.12)circle(1.3mm);
\shade[ball color=black](10.5,0.12)circle(1.3mm);

\end{tikzpicture}
\caption{{\small One-dimensional Ising spin lattice on $\Delta$ (white spins)  with configuration boundary conditions on the complement of $\Delta$ denoted as $\bar\sigma_{\Delta^c}$ (black spins). }}
  \label{fig:BC}
\end{figure}

\subsection{Coarse-graining} \label{subsec:CG}We divide $\mathbb{R}^{d}$ into cubes of side $l=\gamma^{-1/2}$. We denote by $\mathcal{P}^{(l)}_{\mathbb{R}^d}$ the partition of $\mathbb{R}^{d}$. Namely, for every $i\in l\mathbb{Z}^{d}$ we set
\begin{equation}
I_{\gamma,i}=\{r\in \mathbb{R}^d :i_{k}\leq r_{k}\leq i_{k}+l, k=1,\dots, d\}
\end{equation}
($r_k$ and $i_k$ being the $k$-th coordinate of $r$ and $i$). Then we call
\begin{equation}
\mathcal{P}^{(l)}_{\mathbb{R}^d}=\{I_{\gamma,i}: i\in l\mathbb{Z}^{d}\},
\end{equation}
the collection of all the above cubes. 

\begin{definition} [\cite{errico}]\label{def:measu}
(1) A function $f(r)$ is $\mathcal{P}^{(l)}_{\mathbb{R}^d}$-measurable, if it is constant in each cube $I_{\gamma,i}$, $i\in l\mathbb{Z}^d$.

\smallskip

(2) A region $\Delta\subset\mathbb{R}^{d}$ is $\mathcal{P}^{(l)}_{\mathbb{R}^d}$-measurable, if it can be written as a union of cubes of $\mathcal{P}^{(l)}_{\mathbb{R}^d}$ (or its characteristic is $\mathcal{P}^{(l)}_{\mathbb{R}^d}$-measurable). 

\medskip

(3) Any $\Delta\subset\mathbb{Z}^d$ can be identified as a union of cubes with length 1.

\medskip
(4) The size of each cube is given by
\begin{equation}\label{sizeofcube}
|I_{\gamma,i}|=|I|=l^d=\gamma^{-d/2}
\end{equation}
 for every $i\in l\mathbb{Z}^d$. For notational simplicity, we drop $\gamma$ from $I_{\gamma,i}$.

\medskip

For any bounded region $\Delta$ $\mathcal{P}^{(l)}_{\mathbb{R}^d}$-measurable, we denote $\mathbf{\Delta}:=\Delta\cap\mathbb{Z}^{d}$. Hence, $\mathbf{I}_i=I_i\cap\mathbb{Z}^d$.
\end{definition}

\subsection{Coarse-grained Interaction} We introduce a new interaction $\bar{J}_{\gamma}$ which describes the interaction between cubes. More precisely, for every $i,j\in l\mathbb{Z}^{d}$  with $i\neq j$, we consider 

\begin{equation}\label{Jbardiff}
\bar{J}_{\gamma}(i,j)=\frac{1}{|I|^2}\sum_{x\in \mathbf{I}_i}\sum_{y\in \mathbf{I}_j}J_{\gamma}(x,y),
\end{equation}
and for $i=j$, we define
\begin{equation}\label{Jbarequal}
\bar{J}_{\gamma}(i,i)=\frac{1}{|I|(|I|-1)}\sum_{x\in \mathbf{I}_i}\sum_{\substack{x\in \mathbf{I}_i, \\ y\neq x}}J_{\gamma}(x,y)
\end{equation}
\begin{lemma}\label{lemmakac}For fixed and small $\gamma>0$, for any $x\in \mathbf{I}_i$ and any $y\in \mathbf{I}_j$, $i,j\in l\mathbb{Z}^{d}$  with $i\neq j$, we have 
\begin{equation}\label{J_tildeJ1}
|J_{\gamma}(x,y)-\bar{J}_{\gamma}(i,j)|\leq \gamma^{d+\frac{1}{2}}\|DJ\|_{\infty} \mathbf 1_{|x-y|\leq 2\gamma^{-1}}.
\end{equation}
Also, for any $i\in l\mathbb{Z}^{d}$ and any $x,y\in \mathbf{I}_i$, we have
\begin{equation}\label{J_tildeJ2}
|J_{\gamma}(x,y)-\bar{J}_{\gamma}(i,i)|\leq \gamma^{d}\|J\|_{\infty}
\end{equation}
\end{lemma}
\begin{proof} Let $x\in \mathbf{I}_i$ and any $y\in \mathbf{I}_j$, $i,j\in l\mathbb{Z}^{d}$  with $i\neq j$, we have 
\begin{eqnarray*}
|J_{\gamma}(x,y)-\tilde J_{\gamma}(x,y)|&=&|J_{\gamma}(x,y)-\frac{1}{|I|^2}\sum_{z\in \mathbf{I}_i}\sum_{w\in \mathbf{I}_j}J_{\gamma}(z,w)|\nonumber\\
&\leq&\frac{1}{|I|^2}\sum_{z\in I_{i}}\sum_{w\in I_{j}}|J_{\gamma}(x,y)-J_{\gamma}(z,w)|\nonumber\\
&\leq&\frac{1}{|I|^2}\sum_{z\in I_{i}}\sum_{w\in I_{j}}\gamma^{d}\|DJ\|_{\infty}\gamma |x-y-z+w|  \mathbf 1_{|x-y|\leq \gamma^{-1}}\nonumber\\
&\leq&\frac{1}{|I|^2}|I|^2\gamma^{d}\|DJ\|_{\infty}\gamma \gamma^{-1/2} \mathbf 1_{|x-y|\leq \gamma^{-1}}\nonumber\\
&=&\gamma^{d+\frac{1}{2}}\|DJ\|_{\infty} \mathbf 1_{|x-y|\leq \gamma^{-1}}
\end{eqnarray*}
We prove \eqref{J_tildeJ2}  similarly.
\end{proof}
\subsection{Coarse-grained Hamiltonian Energy}\label{CGHam}
In this section we analyze the Hamiltonian energy by using the new interaction defined in \eqref{Jbardiff} and  the estimates in Lemma~\ref{lemmakac}. We start by introducing some notation: for any $r\in\mathbb{R}^{d}$, we define the following quantity as  {\it block spin configuration}:
\begin{equation}\label{sigmal}
\sigma^{(\gamma^{-1/2})}(r):=\frac{1}{|I|}\sum_{x\in \mathbf{I}_{r}}\sigma_{I_i}(x)
\end{equation}
so that 
\begin{equation*}
\sigma^{(\gamma^{-1/2})}(r)=\frac{1}{|I|}\int_{I_{r}}\sigma^{(1)}(r')dr'
\end{equation*}
Let  $\Delta\subset\mathbb{R}^{d}$ be $\mathcal{P}^{(l)}_{\mathbb{R}^d}$-measurable region. We denote by $\mathcal{M}_{\Delta}^{(\gamma^{-1/2})}$ all $\mathcal{P}^{(l)}_{\mathbb{R}^d}$-measurable functions on $\Delta$ with values in 
\begin{equation}\label{M}
M^{(\gamma^{-1/2})}:=\{-1, -1+\frac{1}{\gamma^{-d/2}},\dots, 1-\frac{1}{\gamma^{-d/2}}, 1\}
\end{equation}
For any bounded $\mathcal{P}^{(l)}_{\mathbb{R}^d}$-measurable region $\Delta$ and $m_{\Delta}\in\mathcal{M}_{\Delta}^{(\gamma^{-1/2})}$,  we define as  {\it coarse-grained Hamiltonian energy}\begin{eqnarray}\label{CGHE}
\bar H_{\gamma,h}^{\bar{\mathbf{J}}}( m_{\Delta}; m_{\Delta^c})&:=&\int_{\Delta}\phi_{\beta,h}(m_{\Delta}(r))dr+\frac{1}{4}\int_{\Delta}\int_{\Delta}J_{\gamma}(r,r')[m_{\Delta}(r)-m_{\Delta}(r')]^2 dr dr'\nonumber\\
&&+\frac{1}{2}\int_{\Delta}\int_{\Delta^c}J_{\gamma}(r,r')[m_{\Delta}(r)-m_{\Delta^c}(r')]^2 dr dr'\nonumber\\
&&-\frac{1}{2}\int_{\Delta}\int_{\Delta^c}J_{\gamma}(r,r')m_{\Delta^c}(r')^2 dr dr'\nonumber\\
&&+\frac{1}{\beta}\int_{\Delta} I(m_{\Delta}(r))dr
\end{eqnarray}
where 
\begin{equation}\label{entropy}
I(m):= -\frac{1-m}{2}\log\frac{1-m}{2}-\frac{1+m}{2}\log\frac{1+m}{2}
\end{equation}
with $\phi_{\mathbf{J},\beta,h}(m)$ being given in \eqref{pre}. We recall that $\mathcal{J}=\int_{\mathbb{R}^{d}}J(r)dr$. 
\begin{lemma}Let $\Delta$ be any bounded $\mathcal{P}^{(l)}_{\mathbb{R}^d}$-measurable region $\Delta$, then  there exists a constant $C>0$ such that the following estimate holds:
\begin{equation}\label{estimatehamiltonian1}
\!\!\!\left|H_{\gamma,h}^{\mathbf{J}}(\sigma_{\Delta};\bar\sigma_{\Delta^c})-\bar H_{\gamma,h}^{\bar{\mathbf{J}}}( \sigma^{(\gamma^{-1/2})}_{\Delta}; \bar\sigma^{(\gamma^{-1/2})}_{\Delta^c})\right|\leq C |\Delta|\gamma^{1/2},
\end{equation}
where $ \sigma^{(\gamma^{-1/2})}_{\Delta}$ and $ \bar\sigma^{(\gamma^{-1/2})}_{\Delta^c}$ are defined in \eqref{sigmal}.

\end{lemma}

%--------------------------------------------------------------
\subsection{Estimates for the thermodynamic pressure of an Ising-Kac model}\label{THERPRES}

We recall that 
\[
P_{\mathbf{J},\beta,h}^{\,\Delta,\gamma}(\bar\sigma_{\Delta^c}):=\frac{\log Z_{\bar\sigma_{\mathbf{I}^{\rm{c}}}}(\mathbf{J},\beta, h)}{\beta|\Delta|}
\]
and 
\[
p_{\mathbf{J},\beta,h}:=-\inf_{m\in[-1,1]}\{-hm+\phi_{\mathbf{J},\beta,0}(m)\}
\]
If $\epsilon(\gamma)=\gamma^{1/2}+\gamma^{d/2}\log\gamma^{-1}$, then the following bounds hold: there exist constants $c, c'>0$ such that 
\begin{equation}\label{UpperPres}
P_{\mathbf{J},\beta,h}^{\,\Delta,\gamma}(\bar\sigma_{\Delta^c})\leq p_{\mathbf{J},\beta,h} + \left(c\frac{\gamma^{-1}}{L}+c\epsilon(\gamma)\right),\;\;\;\;\;\;\;{\rm \bf{Upper\;Bound}}
\end{equation}
Let $m^*$ be the minimizer of $\phi_{\mathbf{J},\beta,h}$, then $p_{\mathbf{J},\beta,h}=-\phi_{\mathbf{J},\beta,h}(m^*)$, then 
\begin{equation}\label{lowerPres}
P_{\mathbf{J},\beta,h}^{\,\Delta,\gamma}(\bar\sigma_{\Delta^c})\geq p_{\mathbf{J},\beta,h} -|\phi_{\mathbf{J},\beta,h}([m^*]_{\gamma})-\phi_{\mathbf{J},\beta,h}(m^*)|-c\epsilon(\gamma)-c'\frac{\gamma^{-1}}{L}, \;\;\;\;\;\;\;{\rm \bf{Lower\;Bound}}
\end{equation}
where  $[m^*]_{\gamma}$ is the value in \eqref{M} closest to $m^*$.
\subsection{Limit as $\Delta\nearrow \mathbb{Z}^{d}$ and then $\gamma\to 0$}\label{app:lplimitpre} By using the estimates for the hamiltonian energy given in \eqref{estimatehamiltonian1}, \eqref{UpperPres} and \eqref{lowerPres} we can prove that 

\begin{equation}\label{thermpresupper}
\limsup_{\gamma\to0} \lim_{\Delta\nearrow\mathbb{Z}^d}P_{\mathbf{J},\beta,h}^{\,\Delta,\gamma}(\bar\sigma_{\Delta^c})\leq p_{\mathbf{J},\beta,h}
\end{equation}

\begin{equation}\label{thermpreslower}
\liminf_{\gamma\to0} \lim_{\Delta\nearrow\mathbb{Z}^d}P_{\mathbf{J},\beta,h}^{\,\Delta,\gamma}(\bar\sigma_{\Delta^c})\geq p_{\mathbf{J},\beta,h}
\end{equation}

and therefore if $P_{\mathbf{J},\beta,h}^{\,\gamma}:=\lim_{\Delta\to\mathbb{Z}^d}P_{\mathbf{J},\beta,h}^{\,\Delta,\gamma}$, then

\begin{equation}\label{thermpre}
\lim_{\gamma\to 0} P_{\mathbf{J},\beta,h}^{\,\gamma}= p_{\mathbf{J},\beta,h}=-\inf_{m\in[-1,1]}\{-hm+\phi_{\mathbf{J},\beta,0}(m)\}.
\end{equation}
Hence, the thermodynamic pressure converges to the mean field pressure at the LP-limit, namely
\[\lim_{\gamma\to0} \lim_{\Delta\nearrow\mathbb{Z}^d}P_{\mathbf{J},\beta,h}^{\,\Delta,\gamma}= p_{\mathbf{J},\beta,h}
\]
where $p_{\mathbf{J},\beta,h}$ is defined in \eqref{pre}.  The convexity properties are provided by the limit as $\Delta\nearrow\mathbb{Z}^d$ and then preserved by $\gamma\to 0$. 

\subsection{\bf Thermodynamics of an Ising-spin model with a Kac potential}\label{thmdynamic1} It is shown that when  $\gamma>0$ is sufficiently small, the phase diagram of an Ising-spin model with a Kac potential is close to  the phase diagram of a mean field model. Precisely, in \cite{CP, BZ} (see also \cite{errico}) it is proved that for $d\geq 2$, if $h\neq 0$ then there exists a unique DLR measure, \cite{Presutti}. If $h=0$ there is a critical value of inverse temperature $\beta_c(\gamma)>0$ such that for any $\beta<\beta_c(\gamma)$, there exists one DLR measure while for $\beta>\beta_c(\gamma)$ there are at least two distinct DLR measures $\mu_{\beta,\gamma}^{\pm}$.
 %which are obtained by taking the thermodynamic limit of finite volume Gibbs measures with boundary conditions $\bar\sigma_{\Delta^c}=\pm$ (all spins on $\Delta^c$ are plus/minus 1). 
Finally,  there is an absence of phase transition when $\gamma$ is kept small (for more details see \cite{errico, Presutti} and references therein).
\section{Detailed Analysis of Medical Diagnostics}\label{App:MedicalDiagnostics}
\subsection{Baseline model}\label{subsec:BLMEX}Let us consider the undirected graph in Figure~\ref{fig:M}, \cite{D1} denoted by $\mathcal{G}$.  The class of maximal cliques is  $\mathcal{C}_{\mathcal{G}}=\Big\{\{1,2\}, \{2,3,4\}\Big\}$. The distribution defined over the graph is a log-linear model with clique potentials given by $\Psi_{c}(\mathbf{y}_{c}\mid\mathbf{w}_{c})=e^{w_{c} f_{c}(\mathbf{y}_{c})}$,  where all the weights $\mathbf{w}_c$, and the binary functions $f_{c}$ are known. For example, for $c=\{1,2\}$, $\mathbf{w}_{\{1,2\}}=1.5$ and 
\[
f_{\{1,2\}}(\mathbf{y}_{\{1,2\}})=
\begin{cases}
1&,\mathbf{y}_{\{1,2\}}\in\{(s_1,l_1),(s_1,l_0),(s_0,l_0)\}\\
0&, \mathbf{y}_{\{1,2\}}\in\{(s_0,l_1)\}
\end{cases}
\]
Each binary function $f_c$ induces a set $B_c=\{(\omega_1,\omega_2, \omega_3, \omega_4):f_c(\omega_{c})=1\}$. For example,  $B_{\{1,2\}}:=\left\{\omega: \omega_{\{1,2\}}\in\{(s_1,l_1),(s_1,l_0),(s_0,l_0)\}\right\}$. We compare predictions between the baseline and alternatives of Type I and II (see Section~\ref{sec:AlternMed}) for the following QoIs:
\[g(\mathbf{Y})=\mathbf{1}_{A},\;\;\;{\mathrm{for\;any\;event\;of \; interest}}\;A\subset\Omega.
\]
For instance, $A=\!\{$patient is smoker with asthma$\}\!=\{\omega=(\omega_1,\omega_2, \omega_3, \omega_4):\omega_1=s_0, \omega_3=a_0\}$.

\subsection{Alternative models}\label{sec:AlternMed}
\subsubsection{Type I}\label{subsec:T1EX} First, we consider the class of log-linear models $\tilde{p}$ over $
\mathcal{G}$ with weight change in one maximal clique. Let $c$ be the maximal clique that a weight change occurred. Then the clique potential is given by 
\[
\tilde{\Psi}_{c}(\mathbf{y}_{c})=e^{\tilde{w}_{c} f_{c}(\mathbf{y}_{c})}
\]
The weight after increasing or decreasing by $100 a \%$ equals to $\tilde{w}_c=(1+a) w_c$, where $a\in[-1,1]$ stands for the model uncertainty of alternative models of Type I and $w_c$ is the weight on $c$ of the baseline model $p$.  For example,  for $\{1,2\}$, the corresponding clique potential is expressed as
\begin{eqnarray*}
\tilde{\Psi}_{\{1,2\}}(\mathbf{y}_{\{1,2\}}\mid\tilde{\mathbf{w}}_{\{1,2\}})&=&e^{\tilde{w}_{\{1,2\}} \tilde{f}_{\{1,2\}}(\mathbf{y}_{\{1,2\}})}\nonumber\\
&=&\Psi_{\{1,2\}}(\mathbf{y}_{\{1,2\}}\mid\mathbf{w}_{\{1,2\}})\Phi_{\{1,2\}} (\mathbf{y}_{\{1,2\}}\mid\tilde{\mathbf{w}}_{\{1,2\}})
\end{eqnarray*}
with
\[
\Phi_{\{1,2\}} (\mathbf{y}_{\{1,2\}}\mid\tilde{\mathbf{w}}_{\{1,2\}})=e^{-0.2w_{\{1,2\}} f_{\{1,2\}}(\mathbf{y}_{\{1,2\}})}
\]
since we consider the simplest case where $\tilde{f}_{\{1,2\}}(\mathbf{y}_{\{1,2\}})=f_{\{1,2\}}(\mathbf{y}_{\{1,2\}})$ as well as the fact that $\tilde{w}_{\{1,2\}}-w_{\{1,2\}}=-0.2w_{\{1,2\}}$. Note that $\mathcal{B}=\{c\}$, where $\mathcal{B}$ defined in subsection~\ref{subsec:altm}.
\medskip

\noindent{\bf Derivation of \eqref{ex_UQboundsI}}: We compute all the quantities involved in \eqref{UQmainine*} explicitly. Let us start with the cumulant generating function:
\begin{eqnarray*}
\Lambda_{p}^{f}(\lambda)&=&\log E_{p}[e^{\lambda g}]=\log \left(\sum_{\mathbf{y}\in A}e^{\lambda g} p(\mathbf{y})+\sum_{\mathbf{y}\notin A}e^{\lambda g} p(\mathbf{y})\right)\nonumber\\
&=&\log\left(e^{\lambda}p(A)+1-p(A)\right)
\end{eqnarray*} 
It is straightforward to see that 
\begin{equation}\label{eq:ratioEX}
\frac{d\tilde{p}}{dp}=\frac{\Phi^{\mathrm{I}}}{E_p[\Phi^{\mathrm{I}}]}=\frac{e^{aw_cf_c}}{e^{aw_c}p_{\mathrm{I}}+1-p_{\mathrm{I}}}.
\end{equation}
 and we now go through the computation of $E_{p}[\Phi^{\mathrm{I}}]$:
\begin{eqnarray*}
E_{p}[\Phi^{\mathrm{I}}]&=&\sum_{\mathbf{y}}\Phi^{\mathrm{I}}(\mathbf{y})p(\mathbf{y})=\sum_{\mathbf{y}}e^{aw_cf_c(\mathbf{y}_c)}p(\mathbf{y}_c)\nonumber\\
&=&\sum_{\mathbf{y}\in B_c}e^{aw_cf_c(\mathbf{y}_c)} p(\mathbf{y})+\sum_{\mathbf{y}\notin B_c}e^{aw_cf_c(\mathbf{y}_c)} p(\mathbf{y})\nonumber\\
&=&e^{aw_c} p_{\mathrm{I}}+1-p_{\mathrm{I}}.
\end{eqnarray*}
Similarly, we prove that 
\begin{equation}
E_{p}[\Phi^{\mathrm{i}}\log\Phi^{\mathrm{i}}]=a w_c e^{aw_c}p_{\mathrm{I}}
\end{equation}
Overall, by recalling \eqref{eq:KLnew} the KL divergence equals to 
\begin{eqnarray*}
R(\tilde{p}\| p)=\frac{a w_c e^{aw_c}p_{\mathrm{I}}}{e^{aw_c} p_{\mathrm{I}}+1-p_{\mathrm{I}}}-\log\left(e^{aw_c} p_{\mathrm{I}}+1-p_{\mathrm{I}}\right)
\end{eqnarray*}

\subsubsection{Type II}\label{subsec:T2EX} We consider the class of log-linear models $\tilde{p}$ over $
\tilde{\mathcal{G}}$ with $\tilde{\mathcal{V}}=\mathcal{V}$, $\tilde{\mathcal{E}}=\mathcal{E}\cup e$, where $e$ is a new edge (for example, see Figure~\ref{fig:M}, (Right)). We assume that the edge $e$ enlarges an already existing maximal clique in the sense of the analysis in subsection~\ref{subsec:altm}. The model uncertainties arising from structure-learning from either a new  data set  $\tilde{\mathcal{D}}$ and/or  different prior  knowledge;  see for example Figure~\ref{fig:M} (Right) lie in the binary function $\tilde{f}_{\tilde{c}}$ defined on $\tilde{c}$ and the new weight $\tilde{\mathbf{w}}_{\tilde{c}}$, where $\tilde c$ is the enlargement of an existing maximal clique $c$.  The weight $\tilde{w}_{\tilde{c}}$ can also be expressed with respect to $w_c$: $\tilde{w}_{\tilde{c}}=(1+a)w_c$. This time $a\in\mathbb{R}$, not necessarily in $[-1,1]$ as before (e.g $w_c=1.5$ and $\tilde{w}_{\tilde{c}}=5$).  Then the corresponding clique potential is given by
\[
\tilde{\Psi}_{\tilde{c}}(\mathbf{y}_{\tilde{c}})=e^{\tilde{w}_{\tilde{c}} \tilde{f}_{\tilde{c}}(\mathbf{y}_{\tilde{c}})}=e^{(1+a)w_c \tilde{f}_{\tilde{c}}(\mathbf{y}_{\tilde{c}})}
\] 
The binary function $f_{\tilde{c}}$ induces a set $B_{\tilde{c}}=\{(\omega_1,\omega_2, \omega_3, \omega_4):\tilde{f}_{\tilde{c}}(\omega_{\tilde{c}})=1\}$. For example, Let  $\tilde{\mathcal{G}}\neq\mathcal{G}$ (also $\mathcal{C}_{\mathcal{G}}\neq\mathcal{C}_{\tilde{\mathcal{G}}}$) and $\mathbf{w}\neq\tilde{\mathbf{w}}$. Intuitively, a change on the set of edges can be thought of as  structure-learning from either a new  data set  $\tilde{\mathcal{D}}$ and/or  different prior  knowledge;  see for example Figure~\ref{fig:M}, (Right) where only one new edge has been added. 
 
The set $B_{\tilde{c}}$ satisfies one of the following: $B_{\tilde{c}}\cap B_c=\emptyset$ or $B_{\tilde{c}}\cap B_c\neq\emptyset$. Note that $\mathcal{B}_{\subseteq}=\{\tilde{c}\}$ and $\mathcal{B}_{\cup}=\mathcal{B}_{new}=\emptyset$ with $\mathcal{B}_{\subseteq},\mathcal{B}_{\cup}$ and $\mathcal{B}_{new}$ are defined in subsection~\ref{subsec:altm}. 

\medskip

\noindent{\bf Derivation of \eqref{EBd}}: The cumulant generating function is the same as in the derivation of \eqref{ex_UQboundsI}. Let us compute the expected value of the total $\tilde{p}$-excess factor of type II relative to $p$ with respect to $p$:
\begin{eqnarray}\label{sum}
E_{p}[\Phi^{\mathrm{II}}]&=&\sum_{\mathbf{y}}\Phi^{\mathrm{II}}(\mathbf{y})p(\mathbf{y})=\sum_{\mathbf{y}}e^{(1+a)w_c\tilde{f}_{\tilde{c}}-w_c f_c}p(\mathbf{y})\nonumber\\
&=&\sum_{\mathbf{y}\in B_c}e^{aw_cf_c(\mathbf{y}_c)} p(\mathbf{y})+\sum_{\mathbf{y}\in B_{\tilde{c}}}e^{(1+a)w_c\tilde{f}_{\tilde{c}}-w_c f_c} p(\mathbf{y})+\sum_{\mathbf{y}\notin B_c\cup B_{\tilde{c}}}e^{(1+a)w_c\tilde{f}_{\tilde{c}}-w_c f_c} p(\mathbf{y})\nonumber\\
&=&e^{(1+a)w_c} p_{\mathrm{II}}+e^{-w_c} p_{\mathrm{I}}+1-p_{\mathrm{I}}-p_{\mathrm{II}}.
\end{eqnarray}
We split the sum into the three sums since $B_{c}\cap B_{\tilde{c}}=\emptyset$. Similarly, we prove that 
\begin{equation}
E_{p}[\Phi^{\mathrm{II}}\log\Phi^{\mathrm{II}}]=-w_{c}e^{-w_c}p_{\mathrm{I}}+(1+a)w_ce^{(1+a)w_c}p_{\mathrm{II}}
\end{equation}
Overall, by recalling \eqref{eq:KLnew} the KL divergence equals to 
\begin{eqnarray*}
R(\tilde{p}\| p)=\frac{-w_{c}e^{-w_c}p_{\mathrm{I}}+(1+a)w_ce^{(1+a)w_c}p_{\mathrm{II}}}{e^{(1+a)w_c} p_{\mathrm{II}}+e^{-w_c} p_{\mathrm{I}}+1-p_{\mathrm{I}}-p_{\mathrm{II}}}-\log\left(-w_{c}e^{-w_c}p_{\mathrm{I}}+(1+a)w_ce^{(1+a)w_c}p_{\mathrm{II}}\right)
\end{eqnarray*}
\begin{remark}\label{impremark}
If $B_{c}\cap B_{\tilde{c}}\neq \emptyset$, then we need to split the sum of \eqref{sum} as follows: Let $U\equiv B_{c}\cap B_{\tilde{c}}$, then 
\begin{eqnarray*}
E_{p}[\Phi^{\mathrm{II}}]&=&\sum_{\mathbf{y}}\Phi^{\mathrm{II}}(\mathbf{y})p(\mathbf{y})=\sum_{\mathbf{y}}e^{(1+a)w_c\tilde{f}_{\tilde{c}}-w_c f_c}p(\mathbf{y})\nonumber\\
&=&\sum_{\mathbf{y}\in B_c\setminus U}e^{aw_cf_c(\mathbf{y}_c)} p(\mathbf{y})+\sum_{\mathbf{y}\in B_{\tilde{c}}\setminus U}e^{(1+a)w_c\tilde{f}_{\tilde{c}}-w_c f_c} p(\mathbf{y})+\sum_{\mathbf{y}\in U}e^{(1+a)w_c\tilde{f}_{\tilde{c}}-w_c f_c} p(\mathbf{y})\nonumber\\
&&\quad+\sum_{\mathbf{y}\notin B_c\cup B_{\tilde{c}}}e^{(1+a)w_c\tilde{f}_{\tilde{c}}-w_c f_c} p(\mathbf{y})\nonumber\\
&=&e^{(1+a)w_c} (p_{\mathrm{II}}-p(U))+e^{-w_c} (p_{\mathrm{I}}-p(U))+e^{a w_c}p(U)+1-p_{\mathrm{I}}-p_{\mathrm{II}}+p(U).
\end{eqnarray*}
Note that $p_{\mathrm{I}}, p_{\mathrm{II}}$ and $p(U)$ are computable as $p$ is known.
\end{remark}

\section{Analysis of UQ for Statistical Mechanics}
\subsection{ Proof of Lemma~\ref{lemma:fornorm}}\label{App:prooffornorm}It is not difficult to show (see also Proposition II.1.2 and Lemma II.2.2C in \cite{S}) that 
\begin{eqnarray}
|\log Z_{\bar\sigma_{\Delta^{\rm{c}}}}(\mathbf{J},\beta, h)-\log Z_{\bar\sigma_{\Delta^{\rm{c}}}}(\tilde{\mathbf{J}}^{\mathbf{F}},\beta, h) |&\leq &\beta \|H^{\mathbf{J},h}(\sigma_{\Delta}|\bar{\sigma}_{\Delta^{c}})-H^{\tilde{\mathbf{J}}^{\mathbf{F}},h}(\sigma_{\Delta}|\bar{\sigma}_{\Delta^{c}})\|_{\infty}\nonumber\\
&\leq& |\Delta| \| \Phi^{h,\beta,\mathbf{J}}_{\Delta,\bar\sigma_{\Delta^{c}}} -\Phi^{h,\beta,\tilde{\mathbf{J}}^{\mathbf{F}}}_{\Delta,\bar\sigma_{\Delta^{c}}} \|_1
\end{eqnarray}
which in turn gives 
\begin{equation}
%\frac{1}{|\Delta|}
R(\tilde{q}_{\Delta}\|q_{\Delta})\leq 2|\Delta| \| \Phi^{h,\beta,\mathbf{J}}_{\Delta,\bar\sigma_{\Delta^{c}}} -\Phi^{h,\beta,\tilde{\mathbf{J}}^{\mathbf{F}}}_{\Delta,\bar\sigma_{\Delta^{c}}} \|_1\end{equation}
since 
\begin{eqnarray*}
R(\tilde{q}_{\Delta}\|q_{\Delta})&=&\beta\left( E_{\tilde{q}_{\Delta}}[H^{\mathbf{J},h}(\sigma_{\Delta}|\bar{\sigma}_{\Delta^{c}})]-E_{q_{\Delta}}[H^{\tilde{\mathbf{J}}^{\mathbf{F}},\tilde{h}}(\sigma_{\Delta}|\bar{\sigma}_{\Delta^{c}})]\right)\nonumber\\
&&\qquad+\log Z_{\bar\sigma_{\Delta^{\rm{c}}}}(\mathbf{J},\beta, h)-\log Z_{\bar\sigma_{\Delta^{\rm{c}}}}(\tilde{\mathbf{J}}^{\mathbf{F}},\beta, \tilde{h})
\end{eqnarray*}
A straightforward bound yields that
\[
\| \Phi^{h,\beta,\mathbf{J}}_{\Delta,\bar\sigma_{\Delta^{c}}} -\Phi^{h,\beta,\tilde{\mathbf{J}}^{\mathbf{F}}}_{\Delta,\bar\sigma_{\Delta^{c}}} \|_1\leq \beta\left(|\tilde{h}-h|+\sum_{x\neq0}|F(0,x)|\right).
\]
\subsection{Proof of Lemma~\ref{lemma:explicitphi}}\label{prooflemma:explicitphi} It is a straightforward computation after  subtracting the hamiltonian energies with interaction $J$ and 
\[
\tilde{J}^{F}(x,y)=J(x,y){\bf 1}_{\|x-y\|_{d}\leq R}+F(x,y){\bf 1}_{\|x-y\|_{d}\leq R},\;\; \mbox{Type I},
\]
and
\[
\tilde{J}^{F}(x,y)=J(x,y){\bf 1}_{\|x-y\|_{d}\leq R}+F(x,y){\bf 1}_{\|x-y\|_{d}> R},\;\; \mbox{Type II}
\]

 \subsubsection {Cumulant generating function for $f(\mathbf{Z})=|\Delta| m(\sigma_{\Delta})$} \label{App:cgf}

\begin{eqnarray}\label{specificmgf}
\Lambda_{q_{\Delta};|\Delta| m(\sigma_{\Delta})}(\pm\lambda)&=&\log E_{q_{\Delta}}[e^{\lambda|\Delta|\frac{1}{|\Delta|}\sum_{x\in\mathbf{\Delta}}\sigma_{\Delta}(x)}]\nonumber\\
&=&\log\left(\frac{1}{Z_{\bar\sigma_{\Delta^{\rm{c}}}}(\mathbf{J},\beta, h)}\sum_{\sigma_{\Delta}}e^{\lambda\sum_{x\in \mathbf{\Delta}}\sigma_{\Delta}(x)}e^{-\beta H^{\mathbf{J},h}(\sigma_{\Delta}\mid\sigma_{\Delta^c})}\right)\nonumber\\
&=&\log \left( e^{\lambda\sum_{x\in \mathbf{\Delta}}\sigma_{\Delta}(x)-\beta H^{\mathbf{J},h}(\sigma_{\Delta}\mid\sigma_{\Delta^c})}\right)-\log Z_{\bar\sigma_{\Delta^{\rm{c}}}}(\mathbf{J},\beta, h)\nonumber\\
&:=&\log Z_{\bar\sigma_{\mathbf{\Delta}^{\rm{c}}}}(\mathbf{J},\beta, h\pm\frac{\lambda}{\beta})-\log Z_{\bar\sigma_{\Delta^{\rm{c}}}}(\mathbf{J},\beta, h)
\end{eqnarray}
Then by using the definition of the thermodynamic pressure in \eqref{pressure}, we get:
\begin{equation}
\frac{1}{|\Delta|}\Lambda_{q_{\Delta};|\Delta| m(\sigma_{\Delta})}(\pm\lambda)= \beta\left(P_{h\pm\frac{\lambda}{\beta},\beta,\mathbf{J}}^{\,\Delta,\gamma}-  P_{h,\beta,\mathbf{J}}^{\,\Delta,\gamma}\right)
\end{equation}

\section{Phase diagram of a long range perturbation}\label{sec:long}

\subsection{\bf Thermodynamics of a long range perturbation of 1-dimensional Kac model} \label{thmdynamic2}There is a significant number of works in the literature studying the phase diagram of one-dimensional ferromagnetic Ising model with long range interactions of the form $1/r^{k}$ with  $k$ indicating the decay of interaction and $k\leq2$. For $k<2$, the occurrence of phase transition has been proved (see \cite{D1',D2,D3}). For $k=2$, the existence of a spontaneous magnetization at low temperature is proved in \cite{FS}. The establishment of the existence of phase transition, proving the discontinuity of the magnetization at a critical point, also known as {\it Thouless effect}, was proved by Aizenman et al in \cite{ACCN}. In \cite{CMV}, the authors study the phase diagram of the system with interaction defined in \eqref{inter_type2} with $F$ given in Definition~\ref{def:type2} as illustrated in the right graph of Figure~\ref{fig:kacpic1}. Precisely, they have shown that there is a critical value of the inverse temperature depending on $a$ and $\gamma$ sufficiently small such that the system exhibits phase transition.
\subsubsection{\bf  Phase diagram of a long range perturbation} We consider a one dimensional ferromagnetic Ising spin system with interactions that correspond to a $1/r^2$ long range perturbation of the
usual Kac model, see the right picture of Figure~\ref{fig:kacpic1}.
\begin{definition}\label{def:type2}Let $J^{{\rm pwc}}_{\gamma}(x,y)=\gamma^d{\bf 1}_{|x-y|\leq\frac{\gamma^{-1}}{2}}$(i.e. a special case of Kac-type interaction where in fact $J^{{\rm pwc}}_{\gamma}(x,y)$ is  piecewise constant interaction). Then we define
\begin{equation}\label{inter_type2} 
\tilde J_{\gamma}^{F}(x,y)=
\left\{
	\begin{array}{ll}
		J^{\mathrm{pwc}}_{\gamma} &, 0 \leq|x-y|\leq(2\gamma)^{-1}\\
		F(x,y)  &,\,|x-y|>(2\gamma)^{-1},
		\end{array}
\right.
\end{equation}
with $F(x,y)=\frac{a}{|x-y|^2}$ for some number $a\in(0,\infty)$, Figure~\ref{fig:kacpic1} (right). 
\end{definition}
The range of the perturbation $F$ is clearly Type II. We derive the UQ bounds as follows:
\begin{eqnarray}
\log\Phi_{\bar\sigma_{\Delta^c}}^{\mathrm{i}}(\sigma_{\Delta})&=&\beta\sum_{x\in\Delta}\sigma_{\Delta}(x)\Big(\tilde{h}-h+\frac{1}{2}\sum_{y\in A^{\mathrm{II}}_x\cap\Delta}F(x,y)\sigma_{\Delta}(y)\nonumber\\
&&+\sum_{y\in A^{\mathrm{II}}_x\cap\Delta^c}F(x,y)\bar{\sigma}_{\Delta^{\rm c}}(y)\Big)
\end{eqnarray}
then $C^{\mathrm{II}}:=\beta(\tilde{h}-h)$ and 
\[
\kappa_{\mathrm{II}}:=\beta\sum_{x\in\Delta}\sigma_{\Delta}(x)\Big(\frac{1}{2}\sum_{y\in A^{\mathrm{II}}_x\cap\Delta}F(x,y)\sigma_{\Delta}(y)+\sum_{y\in A^{\mathrm{II}}_x\cap\Delta^c}F(x,y)\bar{\sigma}_{\Delta^{\rm c}}(y)\Big)
\]
We bound $\kappa_{\mathrm{II}}$ based on the following:
\begin{eqnarray}
\sum_{x\in\Delta}\sum_{y\in A^{\mathrm{II}}_x\cap\Delta}F(x,y)&\leq&|\Delta|\sum_{y\in A^{\mathrm{II}}_x}F(0,y)=|\Delta|\sum_{y\in A^{\mathrm{II}}_0}\frac{a}{y^2}\nonumber\\
&=&\gamma |\Delta|\sum_{y\in A^{\mathrm{II}}_0}\frac{\gamma a}{(\gamma y)^2}\leq C\gamma |\Delta|
\end{eqnarray}
for some constant $C$ arises from $\sum_{y\in A^{\mathrm{II}}_0}\frac{a}{y^2}<\infty$.
Then $
\kappa_{\mathrm{II}}\leq 2 C\gamma |\Delta|$
and  the UQ bounds for long range perturbation with $\beta(\tilde{h}-h)<1$ are
\begin{eqnarray}\label{bound:GenIslong}
\pm E_{\tilde{q}_{\Delta}}[m(\sigma_{\Delta})]\leq\frac{1}{1- \beta(\tilde{h}-h)}\inf_{\lambda>0}\Bigg\{ \frac{P_{h\pm\frac{\lambda}{\beta},\beta,\mathbf{J}}^{\,\Delta,\gamma}-  P_{h,\beta,\mathbf{J}}^{\,\Delta,\gamma}}{\lambda/\beta}+\frac{\beta}{\lambda} 2 C\gamma\Bigg\}
\end{eqnarray}
In the LP-limit we get 

\begin{eqnarray}\label{limbound:GenIslong}
\pm M(\tilde{\mathbf{J}}^{F},\beta, \tilde{h})\leq\frac{1}{1- \beta(\tilde{h}-h)}\inf_{\lambda>0}\Bigg\{ \frac{p_{h\pm\frac{\lambda}{\beta},\beta,\mathbf{J}}-  p_{h,\beta,\mathbf{J}}}{\lambda/\beta}\Bigg\}
\end{eqnarray}

\end{document}